\documentclass[journal]{IEEEtran}

\usepackage{times}
\usepackage{epsfig}
\usepackage{graphicx}
\usepackage{amsmath}
\usepackage{amssymb}

\usepackage{multirow}
\usepackage{subfigure}
\usepackage{threeparttable}
\usepackage{stackengine}
\usepackage{footnote}
\usepackage{array}
\usepackage{balance}
\usepackage{tabu}
\usepackage{booktabs}
\usepackage{comment}
\usepackage{arydshln}
\usepackage{dcolumn}
\newcolumntype{d}[1]{D{.}{.}{#1}}
\usepackage{graphbox}

\makeatletter
\newcommand\figcaption{\def\@captype{figure}\caption}
\newcommand\tabcaption{\def\@captype{table}\caption}
\makeatother

\usepackage[ruled,vlined]{algorithm2e}
\SetKwInput{KwInput}{Input}
\SetKwInput{KwOutput}{Output}

\newcommand\IncG[2][]{\addstackgap{%
  \raisebox{-.5\height}{\includegraphics[#1]{#2}}}}

\def\etal{\emph{et al.}}
\newcolumntype{P}[1]{>{\centering\arraybackslash}p{#1}}
\newcolumntype{C}[1]{>{\centering}m{#1}}

\makeatother
\makeatletter
\newcommand{\thickhline}{%
    \noalign {\ifnum 0=`}\fi \hrule height 1pt
    \futurelet \reserved@a \@xhline
}
\newcolumntype{"}{@{\hskip\tabcolsep\vrule width 1pt\hskip\tabcolsep}}

\usepackage{arydshln}

\usepackage[dvipsnames]{xcolor}
\definecolor{subject}{RGB}{255, 147, 0}
\definecolor{pattern}{RGB}{0, 176, 81}
\definecolor{Equiv}{HTML}{FF5964}
\definecolor{Inv}{HTML}{09ACEE}

\usepackage{calc}
\newcommand*{\Equiv}{\mathrm{equiv}}
\newcommand*{\eqequivU}{\ensuremath{\mathop{\overset{\Equiv}{=}}}}%
\newcommand*{\eqequiv}{\mathop{\overset{\Equiv}{\resizebox{\widthof{\eqequivU}}{\heightof{=}}{=}}}}

\newcommand*{\Inv}{\mathrm{inv}}
\newcommand*{\eqinvU}{\ensuremath{\mathop{\overset{\Inv}{=}}}}%
\newcommand*{\eqinv}{\mathop{\overset{\Inv}{\resizebox{\widthof{\eqinvU}}{\heightof{=}}{=}}}}

\usepackage{amsmath}
\usepackage{amssymb}

\usepackage{pifont}% http://ctan.org/pkg/pifont
\usepackage[pagebackref=true,breaklinks=true,letterpaper=true,colorlinks,bookmarks=false]{hyperref}

\begin{document}

%%%%%%%%% TITLE
\title{Occlusion-Invariant Rotation-Equivariant Semi-Supervised Depth Based \\Cross-View Gait Pose Estimation}

\author{Xiao~Gu, Jianxin~Yang, Hanxiao~Zhang, Jianing~Qiu,
        Frank~Po~Wen~Lo, Yao~Guo, Guang-Zhong~Yang,~\IEEEmembership{Fellow,~IEEE}, and Benny~Lo
\thanks{This work was supported by a Newton Fund grant ID:527636592 funded by the UK Department of Business, Energy and Industrial Strategy (BEIS) and delivered by the British Council, and also supported by the Science and Technology Commission of Shanghai Municipality under Grant 20DZ2220400, and the Interdisciplinary Program of Shanghai Jiao Tong University under Grant YG2021QN117.}
\thanks{Xiao Gu, Jianing Qiu, Frank Po Wen Lo, and Benny Lo are with the Hamlyn Centre, Imperial College London, London SW7 2AZ, UK.}% <-this % stops a space
\thanks{Jianxin Yang, Hanxiao Zhang, Yao Guo, and Guang-Zhong Yang are with the Institute of Medical Robotics, Shanghai Jiao Tong University, Shanghai 200240, China.}}% <-this % stops a space}

% \author{Xiao Gu$^1$, Jianxin Yang$^2$, Hanxiao Zhang$^2$, Jianning Qiu$^1$, \\ Frank Po Wen Lo$^1$, Yao Guo$^2$, Guang-Zhong Yang$^2$, and Benny Lo$^1$\\
% $^1$ Imperial College London, UK\\
% $^2$ Shanghai Jiao Tong University, China\\
% {\tt\small xiao.gu17@imperial.ac.uk}

\maketitle

\begin{abstract}
Accurate estimation of three-dimensional human skeletons from depth images can provide important metrics for healthcare applications, especially for biomechanical gait analysis. However, there exist inherent problems associated with depth images captured from a single view. The collected data is greatly affected by occlusions where only partial surface data can be recorded. Furthermore, depth images of human body exhibit heterogeneous characteristics with viewpoint changes, and the estimated poses under local coordinate systems are expected to go through equivariant rotations. Most existing pose estimation models are sensitive to both issues. To address this, we propose a novel approach for cross-view generalization with an occlusion-invariant semi-supervised learning framework built upon a novel rotation-equivariant backbone. Our model was trained with real-world data from a single view and unlabelled synthetic data from multiple views. It can generalize well on the real-world data from all the other unseen views. Our approach has shown superior performance on gait analysis on our ICL-Gait dataset compared to other state-of-the-arts and it can produce more convincing keypoints on ITOP dataset, than its provided ``ground truth''.
\end{abstract}


\section{Introduction}
\label{intro}

% pose estimation to health monitoring
% robot vision and healthcare
\IEEEPARstart{A}{dvances} in computer vision technologies have enabled numerous healthcare-related applications covering from automated disease diagnosis, ambient health monitoring to assistive/therapeutic rehabilitation \cite{haque2020illuminating}. Especially, vision-based markerless human motion capture algorithms play an important role in biomechanical analysis and neurological disease diagnosis. The body skeletons derived from these algorithms can provide macro-level motion trajectories characterizing activities of daily living and movement intentions, as well as detailed kinematics associated with musculoskeletal and neurological conditions~\cite{gu2020cross,deligianni2019emotions,guo2021mcdcd}. Accurately interpreting such informatics, especially the latter health related ones, from human skeletons, requires high estimation accuracy in three-dimensional (3D) space.      

% Pose estimation development 
The area of vision-based 2D pose estimation has shown remarkable achievements recently \cite{cao2019openpose,li2019crowdpose}. However, precise 3D skeleton estimation remains a challenging task caused by the ambiguity when lifting monocular-derived 2D skeleton estimation to 3D. To reduce such ambiguity, multi-view constraints \cite{iqbal2020weakly} and temporal constraints \cite{pavllo20193d} have been utilized. Another line of research targets at 3D sensors such as depth cameras utilizing available 3D information. Several deep learning based methods have been proposed for depth based 3D pose estimation as discussed in Section \ref{sec:depthpose}. %state-of-the-art work either applied regression or detection methods for human pose estimation. 

\begin{figure}
    \centering
    \includegraphics[width=0.95\linewidth]{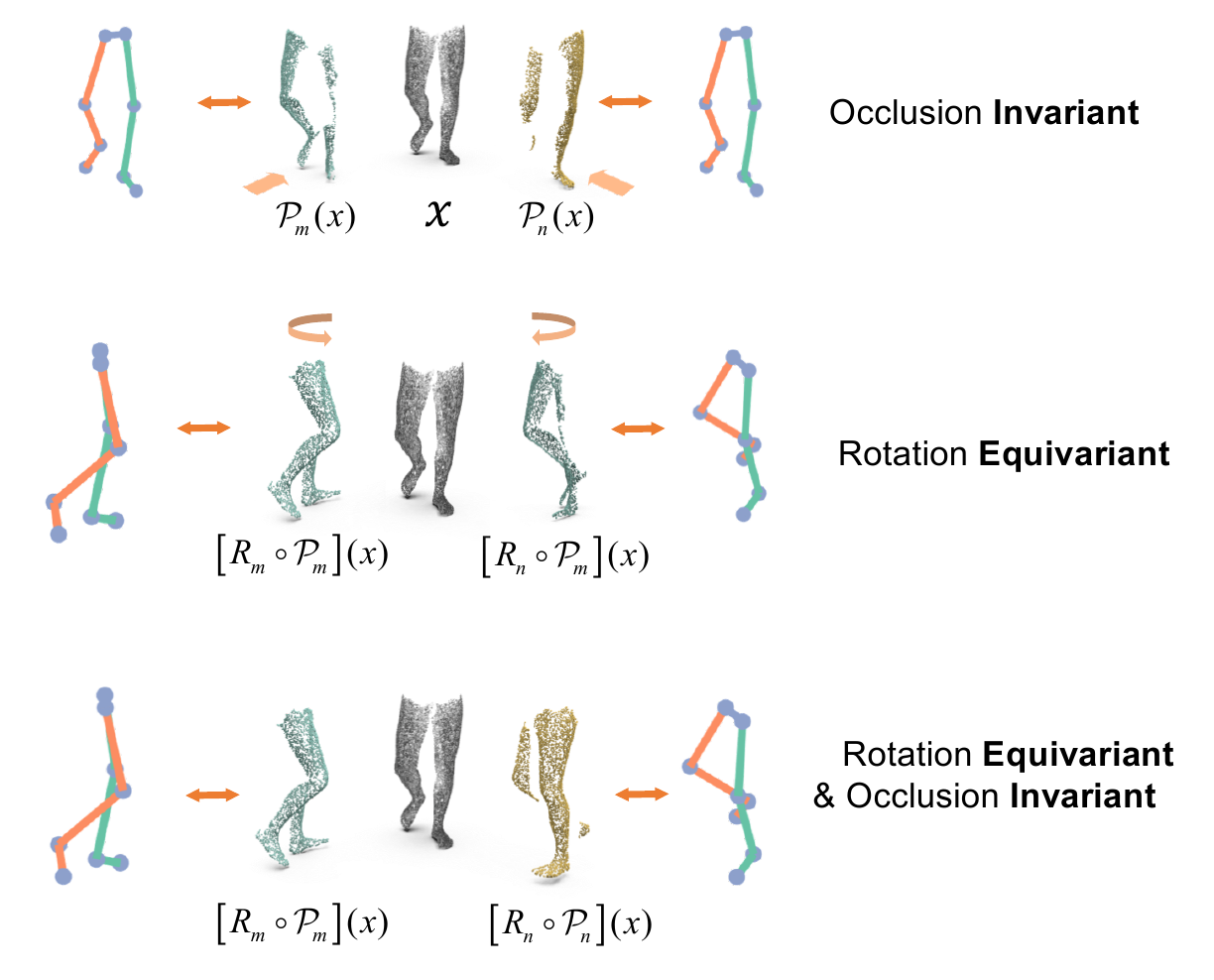}
    \vspace{-15pt}
    \caption{Illustration of occlusion-invariant and rotation-equivariant depth based pose estimation. We disentangle the operations leading to the observed depth image from viewpoint $m$ $[R_m\circ\mathcal{P}_m](x)$ into two parts $\mathcal{P}_m \& R_m$, which denote occlusions and rotations respectively.}
    \label{fig:explanation}
    \vspace{-20pt}
\end{figure}

%The estimated poses should be the same under the occlusions associated with different viewpoints, whereas should be get through the equivariant rotations associated with different viewpoints.

Albeit the 3D information provided by depth images, it is still challenging to perform 3D pose estimation in cross-view contexts \cite{haque2016towards}. As the observed images demonstrating distinct characteristics in terms of different viewpoints \cite{haque2016towards,armagan2020measuring}, the single viewpoint observation leads to two fundamental problems, i.e., occlusion and rotation variations, as shown in Fig.~\ref{fig:explanation}. On one hand, the 3D data encoded in the depth maps will be significantly affected by occlusions caused by self-occlusion and partial-observation. Therefore, from a single viewpoint, estimating anatomical joint positions (underneath the surface) from occluded 3D surface data is very difficult. If the observed 3D data is rotated to a canonical coordinate system, \textcolor{black}{as shown in the first row of Fig.~\ref{fig:explanation}}, the final estimated poses are expected to stay the same even with different occlusions, called occlusion \mbox{``\textbf{\textit{invariant}}''}.  

On the other hand, the body orientations of the depth images captured from a single camera are transformed along with the rotation of the viewpoints (cameras), as shown in the second row of Fig.~\ref{fig:explanation}. Under such changes, the predicted skeleton is expected to undergo the same rotation operation, which can be called rotation ``\textbf{\textit{equivariant}}''. Hence, the final 3D pose estimation model is expected to be both occlusion-invariant and rotation-equivariant, as depicted in the bottom of Fig.~\ref{fig:explanation}.

Recent research has proposed to utilize synthetic data generated by sophisticated virtual human model to enhance the generalization capacity \cite{varol2017learning}. However, there are two issues associated with utilizing synthetic data for healthcare applications with high-precision requirements, such as biomechanical analysis of human gait \cite{colyer2018review}. On one hand, the defined joint positions in these synthetic human models are not the same as those defined in clinical gait models \cite{kainz2017reliability}. On the other hand, the observed human surface data tends to exhibit slight differences compared to the naked synthetic human model due to cloth/shoe distortions. Consequently, direct supervised training with labelled synthetic data would not overcome the heterogeneity between real and synthetic data, and thus not be able to derive convincing clinically related biomechanical indexes for real-world data. %Instead, we proposed a semi-supervised framework by utilizing unlabelled synthetic data for training.

Therefore in this paper, to achieve robust depth based 3D lower-limb pose estimation, we propose a novel multi-view occlusion-invariant semi-supervised framework built upon a rotation-equivariant backbone. To facilitate rotation equivariance, the Cylindrical representation was utilized with 2D heatmap estimation. To achieve the invariance against multi-view occlusions, unlabelled synthetic data from random viewpoints was generated to achieve consistency across occlusions. Training labelled real data from a single view, together with unlabelled synthetic data from multiple views, form our semi-supervised cross-view generalization framework. In this work, we targeted at one healthcare application requiring precise pose estimation, i.e. biomechanical gait analysis. To achieve this, a multi-view dataset was collected, with lower-limb depth images and joint positions regressed from clinical gait models recorded simultaneously. %It is demonstrated that the trained model can generalize well to real-world data from other views with discriminative features preserved. 

Our contributions are three-fold:

\textbf{Rotation Equivariant Pose Estimation}
Our proposed network architecture enables rotation equivariant by performing 3D keypoint estimation under Cylindrical coordinates. Under such coordinate system, rotation group equivariance can be directly realized by conventional planar convolutions with the property of translation group equivariance. %Its efficiency is also comparable to the state-of-the-art methods.   

\textbf{Cross-View Pose Estimation Generalization}
Our proposed framework enables cross-view generalization of pose estimation. With the use of unlabelled multi-view synthetic data and the rotational equivariant backbone, a semi-supervised learning framework can enable both occlusion-invariance and rotation-equivariance.  

\textbf{Multi-View Depth Pose Dataset}
A multi-view dataset with depth data is collected from multiple viewpoints for cross-view human pose estimation benchmarking. Precise lower-limb joints derived from Vicon motion capture system using a clinical gait model were recorded as the ground truth for accurate validation, which is available upon request \url{https://xiaogu.site/ICL_gait}.

% challenges of depth based pose estimation 
% view generalization

%% novelty
% cross-view generalization
% introduce benchmark
% kinematics

\section{Related Works}
\subsection{Multi-View Pose Estimation}
\paragraph{Datasets} Existing multi-view human motion datasets include Human3.6M~\cite{ionescu2013human3}, Total Capture~\cite{trumble2017total}, MPII-INF-3DH~\cite{mehta2017monocular}, ITOP~\cite{haque2016towards}, etc. Most datasets target at 2D RGB images, whereas ITOP captures depth images. Their ground truth skeleton data comes from either manual annotations or tracked marker trajectories, which are not the actual anatomical positions. In fact, a commonly used metric for ITOP dataset is mean average precision (detected ratio of all joints below 10cm)~\cite{haque2016towards,moon2018v2v}. However, this threshold is too high to satisfy the precision required for biomechanical analysis of gait, where the subtle changes between different walking conditions are difficult to recognize. 

% These ground truths are used for activity recognition purposes; thus the error or offset caused by such annotations are fair.
\paragraph{Cross-View Consistency} Based on the above existing multi-view RGB datasets, recent work either made use of epipolar geometry to predict 3D poses from multiple images \cite{qiu2019cross}, or proposed semi-/weakly supervised learning strategies to enhance the estimation in 2D/3D images while reducing the dependency on ground truth annotations \cite{rhodin2018learning,iqbal2020weakly}. The key idea of these work is to ensure the estimation consistency across multiple views. 
Our multi-view framework is inspired by this line of research, as an extension to 3D data. Differently, to avoid the light projection interference across different depth sensors in the real world, we utilized synthetic data to achieve occlusion-invariance. 

\subsection{Depth Based Pose Estimation}
\label{sec:depthpose}
\paragraph{Computational models} For depth based pose estimation, deep learning based approaches have demonstrated superior performance compared to conventional ones. Based on the input representations (depth maps, volume, point sets, etc.), these approaches apply Conv2D \cite{xiong2019a2j,fang2020jgr}, Conv3D \cite{ge20173d,moon2018v2v}, MLPs \cite{ge2018hand,chen2019so}, Transformers \cite{huang2020hand}, etc. to automatically extract spatial features. Among these models, it has been validated that direct 3D convolutions \cite{moon2018v2v} can better preserve the spatial information, yet is computationally inefficient. 

\paragraph{Regression or Detection}
Depth based pose estimation can be considered as a regression or detection task in terms of the output categories. Directly performing 3D coordinate regression \cite{ge20173d, ge2018hand} was utilized by most relevant research. To better capture the spatial information, Xiong~\etal{} \cite{xiong2019a2j} proposed an anchor-to-joint framework by firstly generating 2D joint heatmaps for anchor points and subsequently estimating the joint depth offsets from the surrounding depth values. However, with regards to the accuracy, it is still inferior to the state-of-the-art V2V estimating per-voxel probabilities, via an hourglass-like model composed of a series of 3D convolutions~\cite{moon2018v2v}. 

%Till now, V2V-PoseNet with 3D volume as input and heatmap as output 
%stands for the state-of-the-art method in terms of pose estimation. However,  

%Approaches to depth based human pose estimation can be classified into discriminative (learning-based) \cite{moon2018v2v} or generative (model-based) \cite{}. Hand pose estimation methods follow similar principles, and recent work also combined generative and discriminative approaches together.  

%Based on the pose inference manner, they can also be divided into regression (coordinate regression) or detection (per-pixel probability) based methods. 

% Towards viewpoint invariant 3d human pose estimation
% \cite{haque2016towards}
% \cite{moon2018v2v}
% \cite{xiong2019a2j}
% % Hand Pose
% \cite{malik2020handvoxnet}
\paragraph{Training with Synthetic} It has been a popular method to incorporate simulators and 3D hand/human models, with the purpose of generating synthetic data for pose estimation training~\cite{lin2020cross,varol2017learning}. Different from previous work that used synthetic data for supervised training, we do not rely on the keypoints extracted from synthetic models. As discussed in Section~\ref{intro}, these keypoint positions may be biased from their counterparts defined in clinical gait models \cite{kainz2017reliability}. 

\subsection{Cross-View Generalization}
In fact, cross-view generalization problem arises in a variety of applications, such as face recognition~\cite{tran2017disentangled}, gait recognition~\cite{chao2021gaitset}, pedestrian re-identification~\cite{liang2019cross}, with series of solutions proposed. The main idea of these solutions is to abstract view-invariant features~\cite{tran2017disentangled,liang2019cross} or to realize robust adaptation to the novel view with transfer learning or cross-view transformation~\cite{chao2021gaitset,wang2019learning}. They all focused on cross-view invariant representation learning, limited in our pose estimation task, since the pose estimation model should be equivariant, not invariant, to the rotational transformations. 

Especially, for pose estimation task, Haque~\etal{}~\cite{haque2016towards} proposed a local-patch view invariant model to iteratively refine the results. Although it highlighted the cross-view generalization issue in depth based pose estimation, its model did not perform well when performing cross-view validation.

\subsection{Equivariance Network}
Extracting equivariant feature representations under different data transformations is an expected behaviour for several computer vision tasks such as pose estimation \cite{worrall2017harmonic} and semantic segmentation \cite{lafarge2021roto}. With such equivariance, the input transformations and feature-space transformations can be explicitly associated. %Especially for 3D pose estimation tasks, the feature representations should be correspond to the associated rotations.  

For deep neural networks, the translation-equivariance is ``hard-baked'' into the planar translation convolutions. In pursuit of the equivariance of other transformations like rotations, most widely used method is to train with aggressive data augmentation in a data-driven manner. This might force the network into learning equivariant output, but cannot ensure equivariant representations of the intermediate layers. Existing work further explored the convolution operations, either setting harmonics constraints to filter structures~\cite{worrall2017harmonic} or performing group convolutions on an equivariant filter orbit~\cite{cohen2016group,esteves2019equivariant}. %One interesting work~\cite{joung2020cylindrical} applied cylindrical convolution to learn rotational 

Esteves~\etal{}~\cite{esteves2018polar} proposed a simple yet effective Polar transformer network for rotation equivariant 2D/3D recognition problem. It performs convolution in canonical Polar coordinate systems, with the success of converting translation-equivariance in Polar coordinates to rotation-equivariance in Cartesian coordinates. Our work is inspired by it. Differently, we extended it for pose estimation with 3D data, with heatmap prediction in $\theta\mbox{-}\rho$ and $\theta\mbox{-}z$ planes under Cylindrical representations with the goal of realizing rotation equivariance. 

%\cite{esteves2018polar}
% \cite{worrall2018cubenet}
% \cite{cohen2016group}
% \cite{liu2018deep}
\vspace{-10pt}
\section{Multi-View Lower-Limb Pose Dataset}
% As mentioned above, there are few online available multi-view datasets for depth based pose estimation and their provided ground truth skeletons are of low quality, either coming from manual annotations or recorded marker trajectories

\subsection{Real-World Datasets}    
To benchmark the multi-view depth based pose estimation algorithms, we built a multi-view gait pose dataset\footnote{Experiments were approved by the College Ethics Committee of Imperial College London with the reference No. 18IC4915.} with high-quality anatomical joint positions annotated. 

\begin{figure}[]
    \centering
    \includegraphics[width=\linewidth]{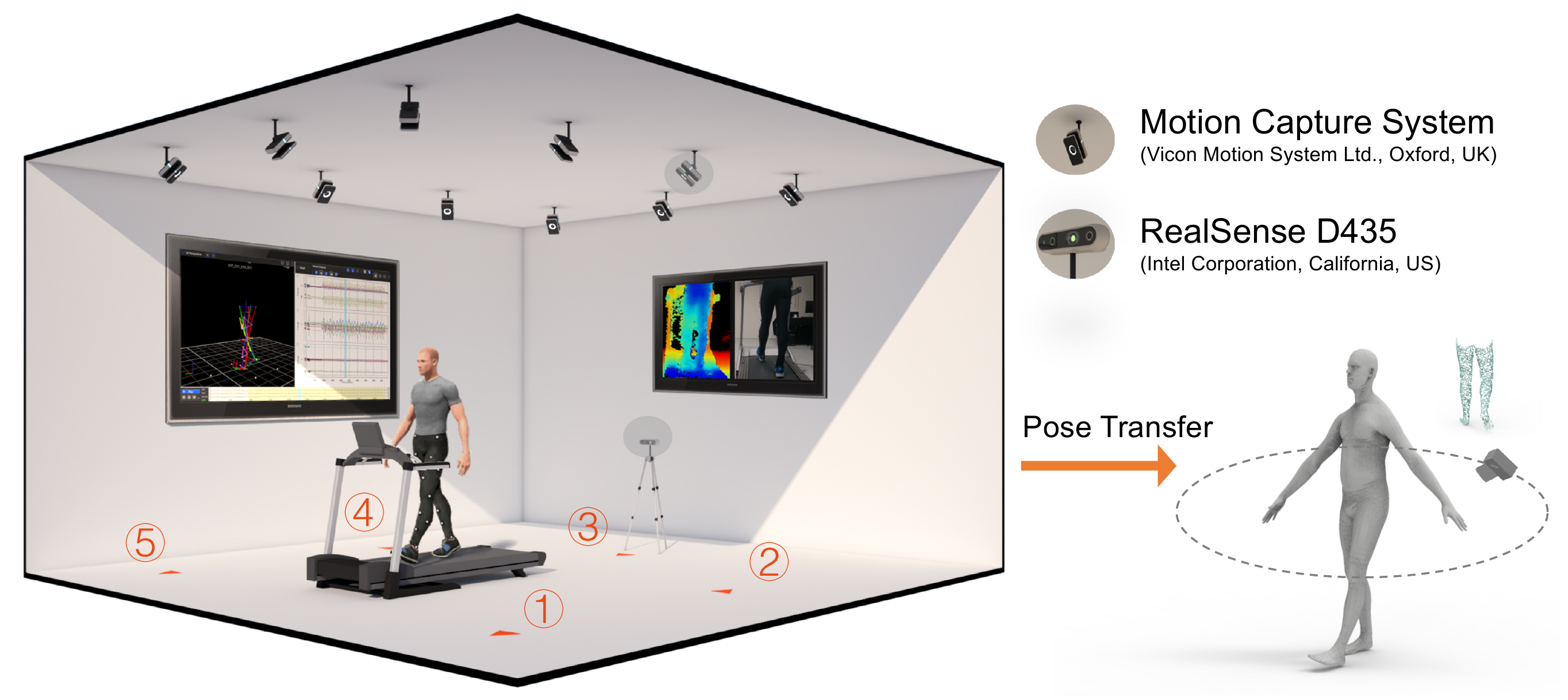}
    \caption{Experiment settings of our multi-view gait pose dataset. The depth camera and Mocap system were synchronized to collect images and ground truth skeletons simultaneously. The derived kinematics were transferred to synthetic human model for synthetic data generation.}
    \label{fig:dataset}
    \vspace{-10pt}
\end{figure}

As shown in Fig.~\ref{fig:dataset}, the \textbf{S}ubjects (6 males, 2 females) were instructed to present different walking styles on a treadmill by supplementary wedged insoles or imitation. It follows the experimental settings similar to existing gait research \cite{gu2020cross, gu2021cross, cui2019effects,mahmood2020evaluation}. In total, five \textbf{C}onditions were considered (normal, supination, pronation, toe-in, and toe-out). The subjects walked on a treadmill, with a calibrated RGBD camera (around 20Hz) and a multi-camera motion capture system (Mocap; 100Hz) collecting data simultaneously. The RGBD camera was positioned in one of the annotated \textbf{V}iewpoints in Fig.~\ref{fig:dataset}, towards the subject without strict requirements on positions and orientations. Reflective markers were placed on the lower-limb surfaces based on the conventional gait model\footnote{\url{https://pycgm2.netlify.app/}} and we followed the standard biomechanic analysis pipeline to derive regressed anatomical joint positions. The RGBD camera and Mocap were synchronized by internet triggered timestamps. The recorded RGB images were fed into the human parsing algorithm~\cite{lin2020cross} to generate lower-limb masks for segmenting valid areas of interest. 100 frames of each \textbf{S}-\textbf{V}-\textbf{C} were provided for training if recorded. 

In our real-world dataset, data from different viewpoints was not collected simultaneously to minimize light projection interference between cameras (depth-depth \& depth-Mocap) and to avoid the intractable synchronization issue between multiple depth cameras \cite{Joo_2017_TPAMI}. 

\subsection{Synthetic Datasets}
As per our experiment settings, the real-world data is only available from single viewpoint A. To learn occlusion-invariant feature representations, we generated multi-view synthetic data based on a synthetic human mesh model SMPL (skinned multi-person linear model \cite{loper2015smpl}). The kinematics derived from our real-world dataset together with varied realistic body shape parameters were transferred to the model. Hidden point removal was applied to simulate occlusions. For each subject, we generated 1000 samples with simultaneously five viewpoints recorded. They are similar to, but not exactly the same as the five views in our real-world settings.  

%To simulate real-world surface point clouds with self-occlusions caused by viewpoint rotations, we applied

The real-world dataset, together with the synthetic data generation program based on our derived kinematics, are available upon request \url{https://xiaogu.site/ICL_gait}. Please refer to the Supplementary Material for more details.

\section{Methods}
\subsection{Method Overview}
For real-scanned depth maps, the occlusion problem $\mathcal{P}$ and rotation problem $R$ are entangled, associated with the given viewpoint. In this work, we divide these two problems apart, and introduce our occlusion-invariant and rotation-equivariant approaches to handle them respectively. 

Without loss of generality, in this section, the original lower limb representation is denoted as $x$ under canonical coordinates, which can be of different representations such as Euclidean volume $\mathbb{R}^{D\times W\times H}$, non-Euclidean point sets $\mathbb{R}^{n\times3}$, etc. The effect of the rotation and occlusion under viewpoint $m$ is denoted as $[R_m\circ \mathcal{P}_m](x)$, which can be viewed as the raw data we actually get from the camera. The corresponding pose is denoted as $y\in\mathbb{R}^{j\times3}$. The objective of our model $\mathcal{F}$ is as below, 

% As we divide these two problems apart, they can viewed as commutative operators, $P_m \circ R_m = R_m \circ P_m $.
\begin{equation}
\small
   \mathcal{F}([R_m\circ\mathcal{P}_m](x)) = R_m\,y 
\end{equation}

We denote the viewpoint in the real-world training set as A, and the set composed of all other testing views as X. The set of all the viewpoints as V, where V = \{A\} $\cup$ X. The $i-$th element in Set V or Set X is denoted as V$_i$, X$_i$, separately. Our final objective is to make a model trained on A work on X as well. The superscript $r$ or $s$ of $x$ (as appears in Eq.~\ref{eq:LS}\&\ref{eq:LM}\&\ref{eq:Lreg}) represents real or synthetic, respectively.

\paragraph{Occlusion-Invariant}
The first issue to address is the occlusion problem, which is formulated as below,  
\begin{equation}
\small
 \mathcal{F}([R_A\circ\textcolor{Inv}{\mathcal{P}_A} ](x)){\tiny =}  {R_A}\,y{\tiny =}\mathcal{F}([R_A \circ \textcolor{Inv}{\mathcal{P}_{X_i}}](x))
\end{equation}

\paragraph{Rotation-Equivariant}
The rotation-equivariance property is formulated as below, 

\begin{equation}
\small
   \textcolor{Equiv}{R'}R_{V_i}\,y{\tiny =} \textcolor{Equiv}{R'}\mathcal{F}([R_{V_i}\circ\mathcal{P}_{V_i}](x)){\tiny =}\mathcal{F}([\textcolor{Equiv}{R'}R_{V_i}\circ\mathcal{P}_{V_i}](x))
\end{equation}

% \\ & = \mathcal{F}(R_{X_i}(x_{\mathcal{P}_A}))
It should be noted that in our overall method, occlusion-invariance is data-driven, whereas rotation-equivariance is model-inherent. With these two properties, it is easy to sort out the following equations,  
\begin{equation}
\small
\begin{aligned}
    \mathcal{F}([\textcolor{Equiv}{R_{X_i}}{\tiny\circ} \mathcal{P}_{X_i}](x)) &\textcolor{Equiv}{\eqequiv}\textcolor{Equiv}{R_{X_i}R_{A}^{-1}}\mathcal{F}([\textcolor{Equiv}{R_{A}}{\tiny\circ}\textcolor{Inv}{\mathcal{P}_{X_i}}](x)) \\
    &\textcolor{Inv}{\eqinv}(R_{X_i} R_{A}^{-1}\mathcal{F}([R_{A}{\circ} \textcolor{Inv}{\mathcal{P}_{A}}](x)) \\
    & =R_{X_i}R_{A}^{-1}R_{A}y = R_{X_i}y
\end{aligned}
\end{equation}

% \textcolor{Inv}{\eqinv}\) \(R_{X_i} R_{A}^{-1}\mathcal{F}([R_{A}{\circ} \textcolor{Inv}{\mathcal{P}_{A}}](x)) {\tiny =}R_{X_i}R_{A}^{-1}R_{A}y {\tiny =} R_{X_i}y\)

\subsection{Semi-Supervised Occlusion-Invariant Framework}
To achieve cross-view occlusion-invariance, we propose a semi-supervised framework to train labelled real data from one viewpoint and unlabelled synthetic data with different occlusions. The reason for such semi-supervised training manner mainly comes from two aspects. On one hand, in real-world settings, it is practically difficult to synchronize multiple depth cameras. This makes sense for our cross-view generalization purposes, as the training procedure is not supposed to see any real data from the other views. On the other hand, synthetic data remains inevitable heterogeneity compared to real data, in terms of both the input data and keypoint positions. Directly mixing these two would lead to poor model performance due to the data divergence.     

\begin{figure}[]
    \centering
    \includegraphics[width=\linewidth]{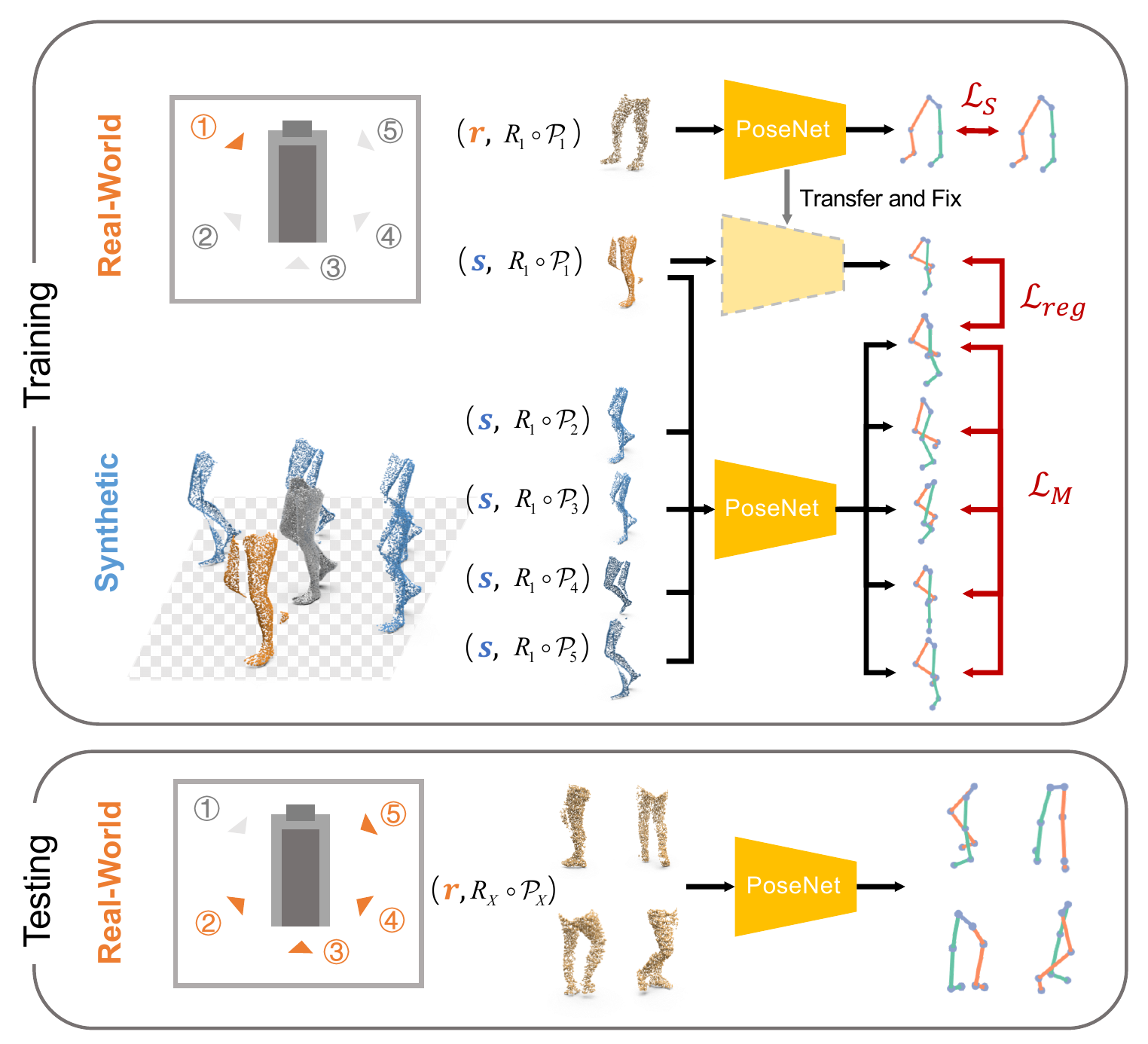}
    \caption{Illustration of our multi-view occlusion-invariant semi-supervised training framework based on labelled single-view real-world data and unlabelled multi-view synthetic data. }
    \label{fig:occ_framework}
    \vspace{-10pt}
\end{figure}

Inspired by existing work~\cite{iqbal2020weakly,rhodin2018learning} utilizing multi-view consistency for RGB based pose estimation, our framework applies multi-view occluded point clouds in the simulators to reach cross-view occlusion-invariance. As shown in Fig.~\ref{fig:occ_framework}, our framework mainly consists of three components as introduced in the following. 

% Our framework was introduced in 

% The occlusion operation $\mathcal{P}$ is not symmetric, thus the line of group convolution theories \cite{cohen2016group} cannot be applied to achieve invariance. To simplify this problem, we   

% Firstly, we aim to address the occlusion-invariant problem in this section. 

% \begin{equation}
% \begin{aligned}
%     \mathcal{F}(\mathcal{P}_m(R_m(x))) &= \mathcal{F}(\mathcal{P}_n(R_m(x))) \\
%     \mathcal{F}(\mathcal{P}_m(x)) &= \mathcal{F}(\mathcal{P}_n(x)) \\
%     \mathcal{F}(\mathcal{P}(x)) &= \mathcal{F}(\mathcal{P}(x))
% \end{aligned}
% \end{equation}

% Based on this, 

\subsubsection{Supervised Single-View Pose Estimation Loss}
Based on the available single-view data from the real world, we apply the supervised loss based on available ground truth data, formulated as below,

\begin{equation}
\small    
     \mathcal{L}_S = \mathcal{L}_{dist}(\mathcal{F}([\mathcal{P}_A\circ R_{A}](x^r)), R_A\,y)
     \label{eq:LS}
\end{equation}

\noindent where $\mathcal{L}_{dist}$ refers to the explicit loss functions depending on the output formats. It can be $\mathcal{L}_2$ loss between ground truth and the predicted heatmaps, the Euclidean distance between poses, etc. This is to make our semi-supervised framework compatible with other pose estimation methods. Specifically, it is $\mathcal{L}_2$ loss in our case for multi-planar heatmap estimation as depicted in Section \ref{sec:backbone}. 

\subsubsection{Unsupervised Multi-View Consistency Loss}
On the basis of the straightforward principle that the estimated poses should be the same under different occlusions, we leverage synthetic data from multiple viewpoints and apply the following loss to learn such property.  

\begin{equation}
\small
\begin{aligned}
    \mathcal{L}_M ={\sum_i} \mathcal{L}_{dist} ({\small \mathcal{F}([ R_A \circ \mathcal{P}_{X_i}](x^s)),
    \mathcal{F}([ R_A  \circ \mathcal{P}_A](x^s))})
\end{aligned}
\label{eq:LM}
\end{equation}

\subsubsection{Regularization Loss}
There exists heterogeneity between real and synthetic data, in terms of data quality and shape distortions. Under such situations, directly applying $\mathcal{L}_M$ plus $\mathcal{L}_S$ would cause overfitting, where $\mathcal{L}_M$ might force the network into a degenerate trivial solution, generating multi-view consistent but drifted poses. Similar issues were also mentioned in previous RGB based pose estimation work \cite{rhodin2018learning, iqbal2020weakly, kocabas2019self}. They applied either regularization loss or front-layer freezing as solutions. In our paper, following the regularization loss proposed in \cite{rhodin2018learning}, we employed a similar loss to enforce the constraint. 

As observed from the training with only real-world data $x^r$, testing results on the synthetic data are reasonably acceptable for those with similar views to the real-world viewpoint A. Therefore, we save the model fully trained with only $x^r$ as $\gamma$, and the regularization for synthetic data pose estimation is formulated as below, 

\begin{equation}
\small
    \mathcal{L}_{reg} = \mathcal{L}_{dist} {(\footnotesize{\mathcal{F}([R_A \circ  \mathcal{P}_A](x^s)),
    \mathcal{F}_\gamma([ R_A \circ \mathcal{P}_A](x^s))}}
    \label{eq:Lreg}
\end{equation}

In practice, we did a progressive training procedure, performing training with only $\mathcal{L}_S$ on $x^r$ during the first n epochs, and subsequently applying semi-supervised training with $\mathcal{L}_S\&\mathcal{L}_M\&\mathcal{L}_{reg}$ on $x^r \& x^s$. 

\begin{figure*}[]
\begin{minipage}[b]{0.75\linewidth}
\includegraphics[width=\linewidth]{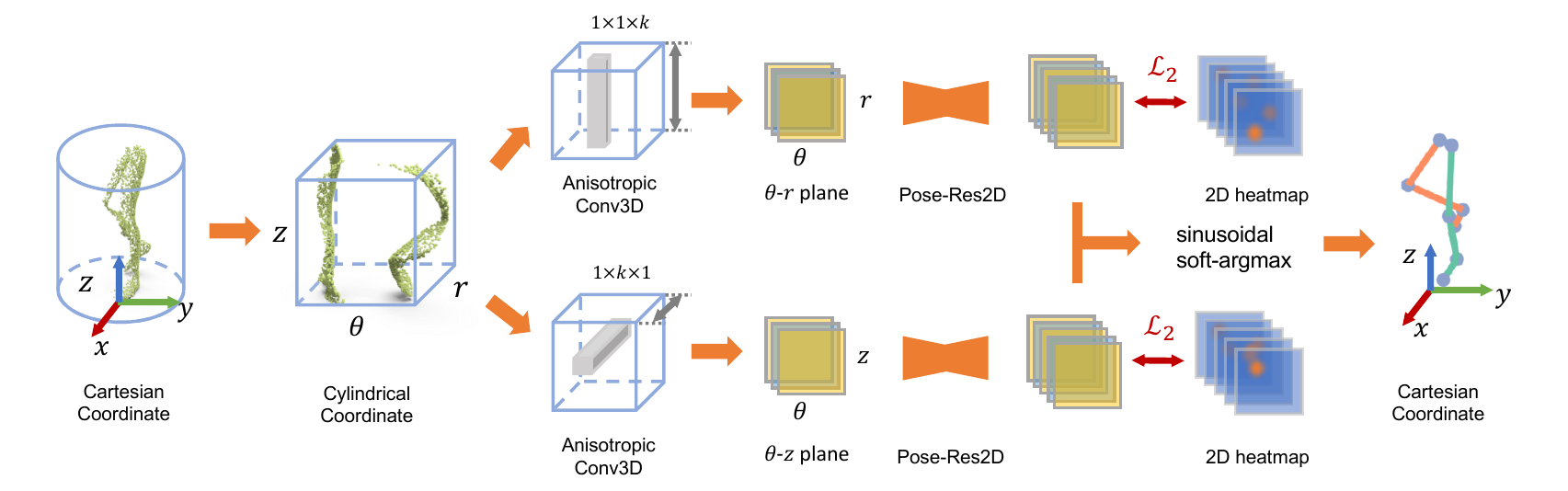}
\caption{Illustration of our proposed cross-view rotation-equivariant backbone model. }
\label{fig:rotation_invariant}
\par\vspace{0pt}
\end{minipage}
\hspace{-10pt}
\begin{minipage}[b]{0.23\linewidth}
\includegraphics[width=0.9\linewidth, height=0.9\linewidth]{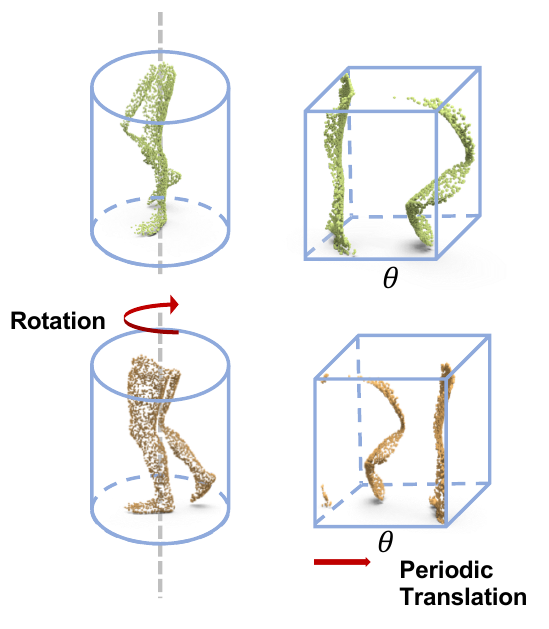}
\caption{\small Equivalence between Cylindrical translation and Cartesian rotation. }
\label{fig:equivalence}
\par\vspace{0pt}
\end{minipage}
\hfill
\vspace{-15pt}
\end{figure*}

\vspace{-5pt}
\subsection{Rotation-Equivariant Backbone}
\label{sec:backbone}
To this end, it is expected that the cross-view pose estimation problem can be addressed by aforementioned occlusion-invariant framework with random rotation augmentations. However, applying aggressive data augmentation is not a sound solution and the vanilla operations on 3D data like MLPs~\cite{you2020pointwise} or convolutions~\cite{worrall2018cubenet} are inherently sensitive to rotational transformations. To address this, we introduce our rotation-equivariant backbone, which not only realizes the rotation equivariance by convolution in the novel coordinate system, but also is more efficient than direct 3D convolution. We firstly have a revisit of group equivariance and then introduce our rotation-euqivariant backbone. 

\subsubsection{Group Equivariance Theorem}
The most straightforward group equivariant operation is the 2D planar convolution. 2D images/filters can be considered as planar signals, $f,h:\mathbb{R}^2\mapsto \mathbb{R}$. The planar convolution can be viewed as the coordinate value addition over a group of translations on these planar signals \cite{esteves2019equivariant}. 

\begin{equation}
\small
   (f*h)(y) = \int_{x\in\mathbb{R}^2}f(x)h(y-x)\,dx
\end{equation}

It can be extended to the generalized group convolution for any transformation group G and its value functions $f,h: G \mapsto \mathbb{R}$
\begin{equation}
    (f*h)(y) = \int_{g\in G} f(g)h(g^{-1}y)\,dg
\end{equation}

It can be easily proved that the group convolution is equivariant to the operations belonging to this group,
\begin{equation}
\small
\begin{aligned}
  (L_\lambda f * h)(y) &= \int_{g\in G} f(\lambda^{-1}g)h(g^{-1}y)\,dg \\
  & = \int_{g'\in G} f(g')h(g'^{-1}\lambda^{-1}y)\, dg' \\
  & = (f * h) (\lambda^{-1} y) \\
  & = L_\lambda(f*h)(y)
\end{aligned}
\end{equation}

\noindent where $L_\lambda f(g)$ = $f(\lambda^{-1}g)$ refers to a group operation of group $\lambda$ performed on signal $f$.

In our case, the rotation $R$ around $z$ axis can be viewed as the SO(2) rotation group~\cite{esteves2018polar}, and our objective is to perform convolution on the SO(2) group with the goal of realizing rotational equivariance\footnote{According to the chirality of lower limbs, we assume the min-max normalization can shift the 3D data to the central rotation axis and normalize the scale, with only small variances left in the translations and scales. Therefore, in this work we only consider the equivariance of SO(2) group.}.  

\subsubsection{Rotation-Equivariant Convolution with Cylindrical Representations}

% \begin{equation}
% \begin{aligned}
%  h(r)&=\int_{s\in SO(2)}f(s)\phi(s^{-1}r)ds \\
%  h(r)&=\int_{s\in SO(2)}
% \end{aligned}
% \end{equation}

Instead of directly applying group convolution on SO(2) in the original Cartesian coordinates, we convert $(x,y,z)$ to Cylindrical coordinates $(\theta,\rho,z)$, from $(x,y,z)$ to $(\rho cos(\theta), \rho sin(\theta), z)$.     
Considering the group convolution over the translation of $\theta$ axis, it can be directly associated with SO(2) equivariance under Cartesian coordinates (explained in Fig.~\ref{fig:equivalence}) as shown below, 

\begin{equation}
\small
\begin{aligned}
    (f*h)(y) &= \int_{g\in SO(2)} f(g)h(g^{-1}y)dg \\ 
             &= \int_{\theta \in \mathbb{R}} \hat{f}(\theta) \hat{h}(\theta_y-\theta)\,d\theta   
\end{aligned}
\end{equation}

% If we denote SO(2)$\times\mathbb{R}^{\tiny{+}}\times\mathbb{R}$.
% \begin{equation}
% \begin{aligned}
%     (f*h)(r) = \int_{s\in SO(2)} f(s)h(s^{-1}r)ds 
% \end{aligned}
% \end{equation}

% The $R\in$SO(2) rotational transformation belongs to a symmetric group, which means $R_m R_n = R_{mn}$. Considering the planar convolution operation between $f,h:\mathbb{R}^2\rightarrow \mathbb{R}$

% We target at the large changes occurring at yaw angle, with slight changes in other axes which cannot be addressed by data augmentation. Therefore, we assume that we cannot be 

% We assume that the original axis is around the middle of the observed lower-limb point clouds. To address the translational invariance, we applied 

\subsubsection{$\theta\mbox{-}r$ $\theta\mbox{-}z$ Heatmap Estimation}
By voxelizing the point sets under Cylindrical coordinates, $\{[-\pi,\pi),[0,\rho_0),[0,z_0)\}\mapsto[0,C)^3$, we can generate the volumetric representation, where $\rho_0$ and $z_0$ are predefined boundary parameters, and $C$ is the converted cube length. Till now, architectures similar to V2V can be applied to estimate per-voxel probability via convolutions. As discussed in Section~\ref{sec:depthpose}, estimating 3D heatmap demonstrates superior performance compared to direct coordinate regression. However, such manner is computationally expensive and not suited for deployment in mobile devices~\cite{vasileiadis2019optimising}.     

% Furthermore, outputting heatmap has been a standardized manner for 2D RGB based pose estimation~\cite{newell2016stacked}.
To compensate the computational inefficiency of 3D heatmap prediction, while preserving the advantage of utilizing heatmaps as output, we converted the original 3D volumetric data to $\theta\mbox{-} r$ and $\theta\mbox{-}z$ via anisotropic 3D convolutions and subsequently applied 2D convolutions by periodic padding on those two planes to estimate corresponding 2D heatmaps. Both heatmaps are fused together to estimate the 3D keypoint positions by (sinusoidal) soft-argmax, as shown in Fig.~\ref{fig:rotation_invariant}. The model details are described as below.

\subsubsection{Model Details}
\paragraph{Anisotropic 3D Convolution}
Anisotropic 3D convolution proposed in \cite{qi2016volumetric} is featured by its elongated anisotropic kernel to aggregate long-range interactions along the elongated axis. The anisotropic probing can ``project'' 3D data into 2D planes while encoding the 3D structure information into the feature channel. As shown in Fig.~\ref{fig:rotation_invariant}, the anisotropic 3D convolution along $z$ and $r$ axis converts original volume data to $\theta\mbox{-}r$ and $\theta\mbox{-}z$ planes separately. ResNet based pose estimation networks are thereafter applied on them to estimate corresponding 2D heatmaps. We applied ResNet-50 as the backbone in practice.  

\paragraph{Periodic Padding}
Normally, for convolution operations, commonly applied padding operations are zero padding. This is not ideal for our case, as there exists periodic property along the $\theta$ axis. Differently, periodic padding allows to pad the boundary values on one side to the other side over $\theta$ axis, which enables continuous cyclic sliding of kernel filters even at the boundary.   

% periodic padding is based on the periodic property of the $\theta$ axis $[0,2\pi)$, where samples distributed at the edges is continuous.  

\paragraph{Sinusoidal Soft-Argmax}
Similarly, according to the periodicity, we applied sinusoidal soft-argmax as proposed in \cite{joung2020cylindrical} to derive the targeted $\theta$ positions formulated as below, 

\begin{equation}
\small
\begin{aligned}
\theta^* = atan2 \left ( \int_{\tiny\mbox{-}\pi}^{\tiny \pi^{\mbox{-}}}e^{p(\theta)}\sin\theta d\theta, \int_{\tiny\mbox{-}\pi}^{\tiny \pi^{\mbox{-}}}e^{p(\theta)}\cos\theta d\theta \right )
\end{aligned}
\end{equation}
where $p(\theta)$ denotes the fused probability of two branches. For $\rho \& z$, we applied the original soft-argmax manner to derive the estimated positions. 

% This sinusoidal soft-argmax could 

% \begin{equation}
% \begin{aligned}
%         x & = \rho \cos{\theta} = \sqrt{2r} \cos{\theta}; \\ 
%         y &= \rho \sin{\theta} = \sqrt{2r} \sin{\theta}; \\
%         z &= z  
% \end{aligned}
% \end{equation}
    
% \begin{equation}
% \begin{aligned}
%         dV  = \rho \, d\rho \, d\theta \, dz = \sqrt{2r} \, d\sqrt{2r} \, d\theta \, dz = dr \, d\theta \, dz 
% \end{aligned}
% \end{equation}

\vspace{-10pt}
\section{Results}
\subsection{Experimental Settings}
Our experiments were conducted with Pytorch + GeForce RTX 2080Ti GPUs. The architecture details can be found in our Supplementary Material. 
For our proposed method, the 3D data was firstly applied with min-max normalization and the cube size was set as 128. Subsequently, slight rotational augmentation around x/y axes [-5$^\circ$, 5$^\circ$], slight 3D translational augmentation [-5, 5] cube units, and Chirality augmentation \cite{yeh2019chirality} were applied. We did not exert additional rotational augmentation around $z$ axis for our method to highlight the rotational equivariance of our backbone. Further details can be found in our Supplementary Material.%Besides, we add another effective augmentation, Chirality augmentation, introduced in \cite{yeh2019chirality}. 

\begin{table*}[]
    \begin{minipage}[b]{0.75\linewidth}
   \footnotesize
    \caption{Quantitative results of cross-subject (CS) validation}
\begin{tabular}{@{}ccccccccc@{}}
\toprule
\multirow{2}{*}{Methods} & \multicolumn{4}{c}{Dist (cm) $\downarrow$}   & \multicolumn{4}{c}{mAP (5cm) $\uparrow$ }   \\ \cmidrule(l){2-9} 
                         & Hip          & Knee           & Ankle         & Toe          & Hip & Knee & Ankle & Toe    \\ \midrule
PN++~\cite{ge2018hand}   & 4.57$\pm$2.65 &  4.26$\pm$2.39   &  4.93$\pm$2.97 &  5.45$\pm$3.39       &   0.660   &    0.711  &    0.606   &  0.535         \\
SO-Net~\cite{chen2019so} &  3.65$\pm$2.06    &   3.43$\pm$1.93  &   3.82$\pm$2.32     &   4.40$\pm$2.68   &   0.808  &  0.848   &  0.789&   0.689         \\
A2J~\cite{xiong2019a2j}  &  4.25$\pm$6.51      &   2.89$\pm$2.08  &    3.52$\pm$4.53      &   4.19$\pm$5.31       &   0.837  & 0.921     & 0.882      &  0.800            \\
V2V~\cite{moon2018v2v}   &  3.75$\pm$2.11  &     3.37$\pm$1.30  &    3.37$\pm$1.74      &   3.84$\pm$2.21       &  0.823  &    0.895  &     0.887  &    0.812         \\ \midrule
Proposed-88 &  \underline{3.01}$\pm$1.46 & \textbf{2.53}$\pm$1.19 & \underline{2.81}$\pm$1.56 & \underline{3.25}$\pm$1.79 & \textbf{0.914} & \textbf{0.970} & \underline{0.927} & \underline{0.867}     \\ 
Proposed     & \textbf{2.97}$\pm$1.56 & \underline{2.56}$\pm$1.23 & \textbf{2.74}$\pm$1.59 & \textbf{3.18}$\pm$1.86 & \underline{0.912} & \underline{0.964} & \textbf{0.934}  & \textbf{0.874}        \\ \bottomrule
\end{tabular}
    \label{tab:CS}
    \par\vspace{0pt}
    \end{minipage}
    \hspace{-25pt}
    \begin{minipage}[b]{0.23\linewidth}
    \includegraphics[width=1.3\linewidth]{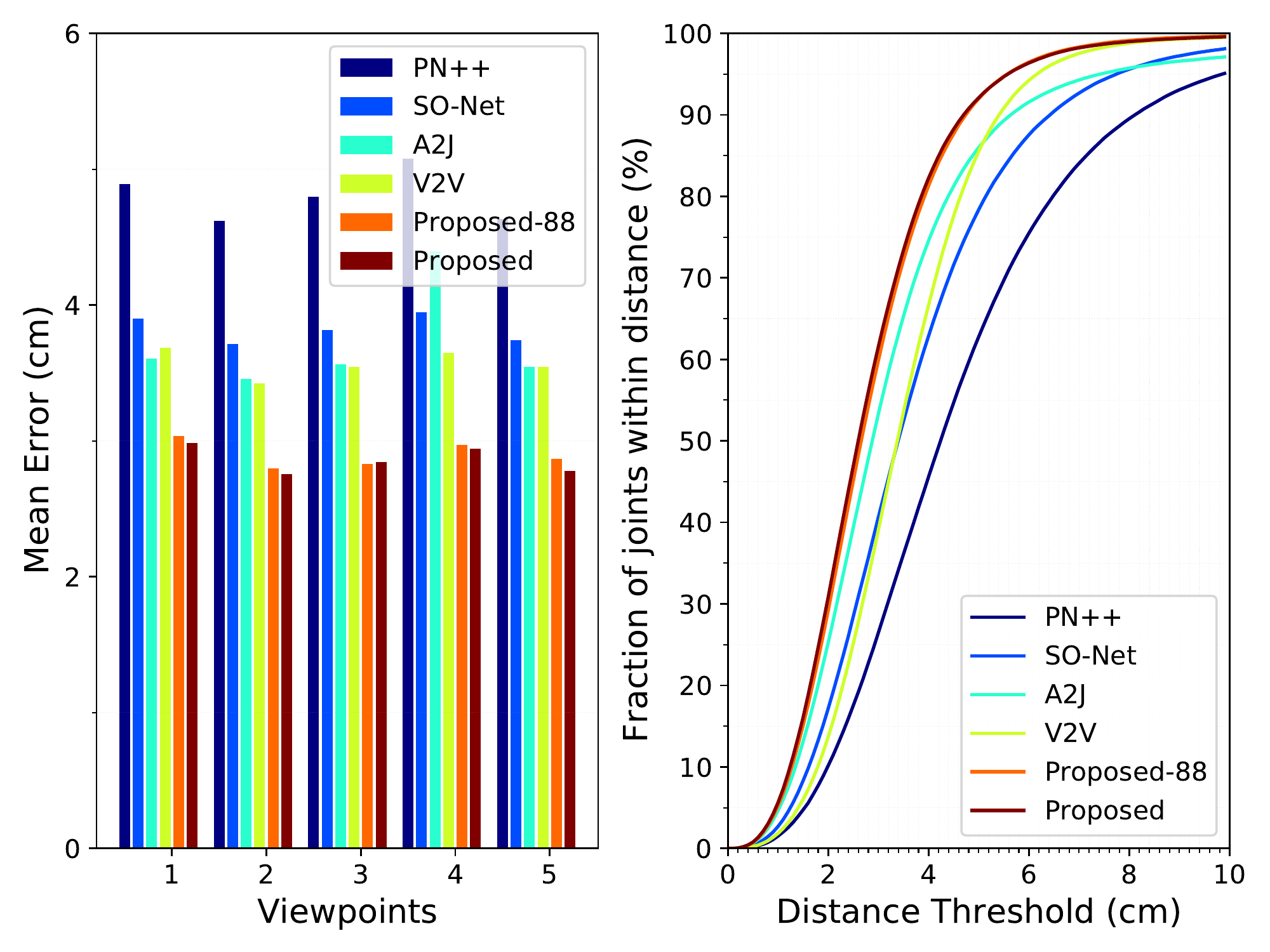}
    \par\vspace{-10pt}
    \end{minipage}
    \vspace{-10pt}
\end{table*}

\subsubsection{Evaluation Protocols}
To highlight the cross-view cross-subject robustness of our framework and the novel backbone, we introduce three evaluation protocols as mentioned below. The training subset is further divided to training (85\%) and validation (15\%). 

\begin{itemize}
    \item \textbf{Cross-Subject (CS)}
    The cross-subject validation protocol was used to evaluate the performance of our backbone model compared to other methods under intra-view settings. We split data into two folds, with four subjects each, and then apply training on one-fold and testing on the other. 
    \item \textbf{Cross-View (CV): Train-A-Test-X}
    To match the practical scenarios where the model trained on one view is expected to generalize to other arbitrary views, our cross-view validation is conducted under the extreme condition by leave-one-view-out for training rather than the commonly used leave-one-view-out for testing. We denote this protocol as cross-view (CV). We use mix the data of all the subjects in the CV only settings.
    
    % Here A is denoted as a viewpoint randomly selected from all the five available views, and X is denoted as the held-out four other views. 
    
    % In practice, in each loop session, we utilized real-world data from the training view and synthetic data from all the views for training. 
    \item \textbf{Cross-View Cross-Subject (CV-CS): Train-A-Test-X -Cross-Subject}
    On the basis of CV, CV-CS further split data into two groups by the cross-subject setting of CS protocol. 
    
    %\item Train-X-Test-A-Cross-Subject (XA-CS)
    
\end{itemize}

% \usepackage{booktabs}
% \usepackage{multirow}
\subsubsection{Compared Methods}
\begin{itemize}
    \item \textbf{PN++} \cite{ge2018hand} This approach is based on PointNet++~\cite{qi2017pointnet++} extracting hierarchical-level information from input point clouds. It is highlighted by its normalization procedure to minimize difference across viewpoints. Our implementation is based on its official codes\footnote{\url{https://github.com/3huo/Hand-Pointnet}}.  

    \item \textbf{SO-Net} \cite{chen2019so} SO-Net is a multi-task framework with additional point cloud self-reconstruction branch for hand pose estimation. We implemented its fully training version\footnote{\url{https://github.com/TerenceCYJ/SO-HandNet}} by simultaneously minimizing the self-reconstruction and pose estimation losses with all the labelled data.

    \item \textbf{A2J} \cite{xiong2019a2j} A2J stands for anchor-to-joint, which firstly estimates 2D heatmaps from the depth maps to derive anchors, and subsequently predicts the joint depth offsets based on surrounding depth values. It can be viewed as a hybrid detection-regression approach\footnote{\url{https://github.com/zhangboshen/A2J}}.

    \item \textbf{V2V} \cite{moon2018v2v} V2V stands for voxel-to-voxel, which directly estimates 3D heatmaps by an hourglass-like architecture composed of 3D convolutions. Our implementation is based on its Pytorch version\footnote{\url{https://github.com/dragonbook/V2V-PoseNet-pytorch}}.
   % \item Proposed
\end{itemize}

In the CS session, only the real-world data was used for training. In the CV and CV-CS session, our semi-supervised training was based on labelled real data from view A $[R_A\circ\mathcal{P}_A](x^r)$ and unlabelled data from all the views $[R_A\circ\mathcal{P}_{V_i}](x^s)$. For compared methods, 
largely degraded performance is inevitable if directly training on real data from A and testing on X. We applied a mixed supervised training\footnote{We noticed both low performance for such mixed training manner and train-syn + fine-tune-real, therefore only the results of mixed training are presented.} with real data from A $[R_A\circ\mathcal{P}_A](x^r)$, and labelled synthetic data from all views $[R_{V_i}\circ\mathcal{P}_{V_i}](x^s)$; they were evenly sampled in each batch. Our semi-supervised framework was also applied on V2V with random rotation augmentations for comparison, denoted as V2V-Semi. A similar mixed supervised training based on our backbone model was also implemented, called Proposed-Mix. 

% We observed the inability of achieving convergence when simply mixing real and synthetic data together, as their relative ground truth positions are different. Therefore, we compared two 
\subsubsection{Metrics}
We applied two commonly used 3D pose estimation metrics for evaluation. The first is the 3D mean Euclidean distance error (\textbf{Dist}), which calculates the average 3D Euclidean distance per joint. The second is the mean average precision (\textbf{mAP}), which measures the detection ratio within the precision threshold. Different from the 10cm utilized in \cite{haque2016towards}, we applied 5cm to set a stricter threshold as required for biomechanical gait analysis applications.  

%GDI-Diff (Gait Deviation Index Difference) 

\vspace{-5pt}
\subsection{Quantitative Results}
The quantitative results of pose estimation under different protocols are presented in Table~\ref{tab:CS}$\&$\ref{tab:CV_Full}$\&$\ref{tab:CV_CS} respectively, with both the error plot of each view and error curve plot attached alongside the corresponding table. 

% The cross-subject validation is conducted to firstly compare the model performance across subjects of similar views, which to some degree compare the generalization performance of the backbone itself.
\paragraph{CS Validation} It is observed in Table~\ref{tab:CS} that V2V and our proposed backbone perform better than the others. This is mainly due to the heatmap output used in V2V and ours, rather than the  coordinate regression used in the regression (PN++, SO-Net) or semi-regression (A2J) methods. It is also found that our method performs better than V2V, which demonstrates the effectiveness of our backbone by directly regressing on two projected planes. As the viewpoints and shapes might slightly differ across subjects, our method on Cylindrical representations are more robust against such variations. 

Meanwhile, the input resolution of our Cylindrical representations is Cube-128, whereas V2V under Cartesian coordinates applies Cube-88 by its default settings. Direct comparison between these two partitioning resolution is not reasonably fair. Therefore, we also compared the variant version of our proposed model with a cube size of 88, referred to as Proposed-88. Proposed-88 presents a slight difference compared to the original version, yet still outperforms V2V significantly. 

It should be noted that V2V applied 3D Residual as its backbone whereas our method applied 2D ResNet-50 as the backbone of each planar branch. Although they share similar residual architectures, the 3D backbone of V2V is rather shallower than ResNet-50. 
The performance discrepancy might be caused by this; however, the V2V is memory-expensive due to 3D convolutions, and therefore is not feasible for a much deeper design in practice. 

\paragraph{CV \& CV-CS Validation} The cross-view validation is applied to evaluate the cross-view generalization performance. As shown in Table~\ref{tab:CV_Full} \& Table~\ref{tab:CV_CS}, our method outperforms all other supervised methods for both metrics. 
V2V-semi and our proposed perform better than V2V and Proposed-Mix respectively, which shows our semi-supervised framework can overcome the domain heterogeneity between real and synthetic datasets without any domain adaptation techniques.

It is observed that for those compared methods, i.e. PN++, SO-Net, A2J, V2V, Proposed-Mix, they all demonstrate inferior results. This validates the inferior performance of directly mixing real and synthetic data for training. Although synthesizing data from novel viewpoints could provide supplementary information of these unseen views, there exists substantial domain discrepancy between real and synthetic data even under the same viewpoint. One one hand, noises, shape variances, cloth/shoe distortions contribute to the difference of data distribution between two domains. On the other hand, the joint data-label distribution is also biased, as the keypoint positions in SMPL are not exactly the same with those defined in clinical gait models. Please refer to our Supplementary Materials.

It is noteworthy that in both the CV and CV-CS settings, the proposed method have better qualitative and quantitative results for Hip, Knee, and Ankle. However, compared to V2V-semi, our method shows slight inferior performance regarding mAP (5cm) of Toe. It is probably caused by the coordinate conversion from Cartesian to Cylindrical, which leads to sparser grid sampling in peripheral areas (with larger radius) of Cartesian space. Toes tend to locate in these peripheral areas during walking, and such sparse sampling would decrease the precision of keypoint estimation in Cartesian space. 

\begin{figure*}[]
    \centering
    \begin{tabular}{P{.1\linewidth}P{.13\linewidth}P{.13\linewidth}P{.13\linewidth}P{.13\linewidth}P{.13\linewidth}P{.05\linewidth}} 
          \footnotesize{PN++} & \IncG[trim={0 0 4.2cm 0},clip,width=\linewidth]{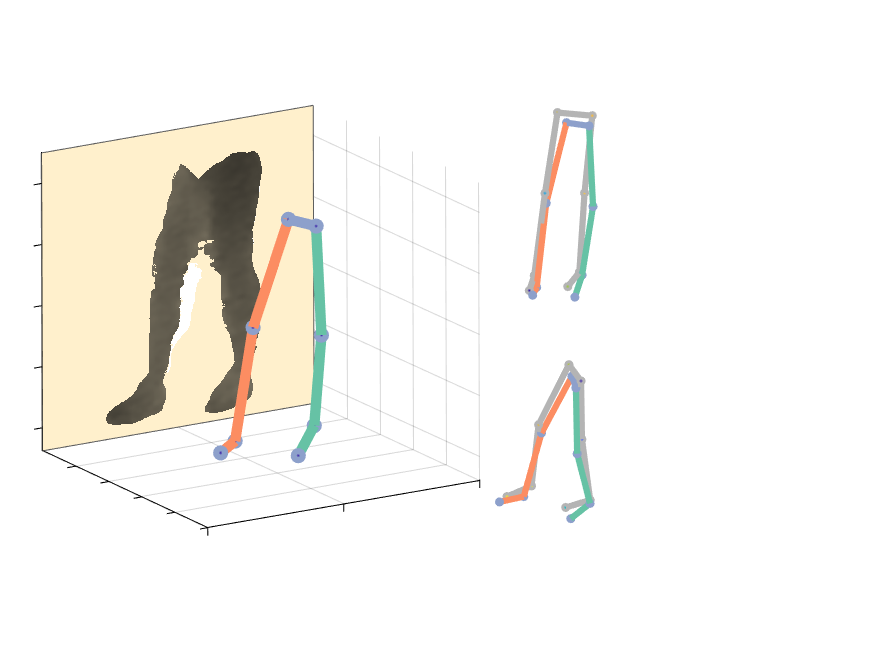} & \IncG[trim={0 0 4.2cm 0},clip,width=\linewidth]{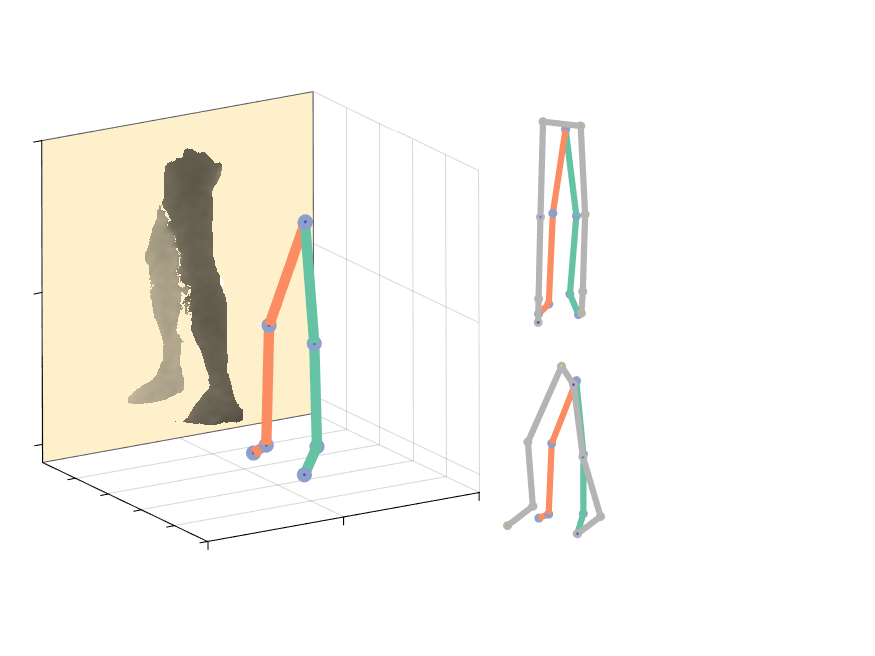} & \IncG[trim={0 0 4.2cm 0},clip,width=\linewidth]{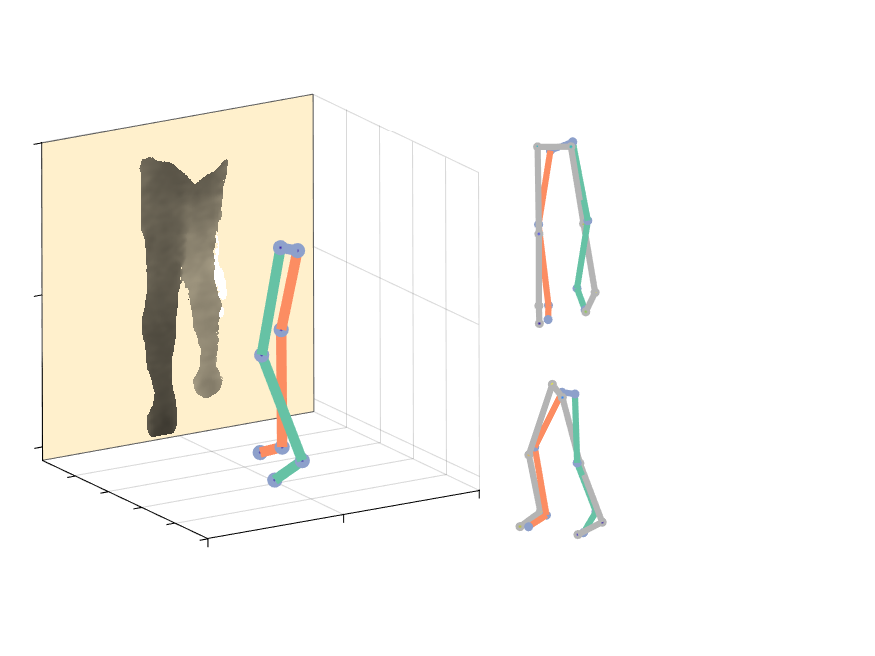} & \IncG[trim={0 0 4.2cm 0},clip,width=\linewidth]{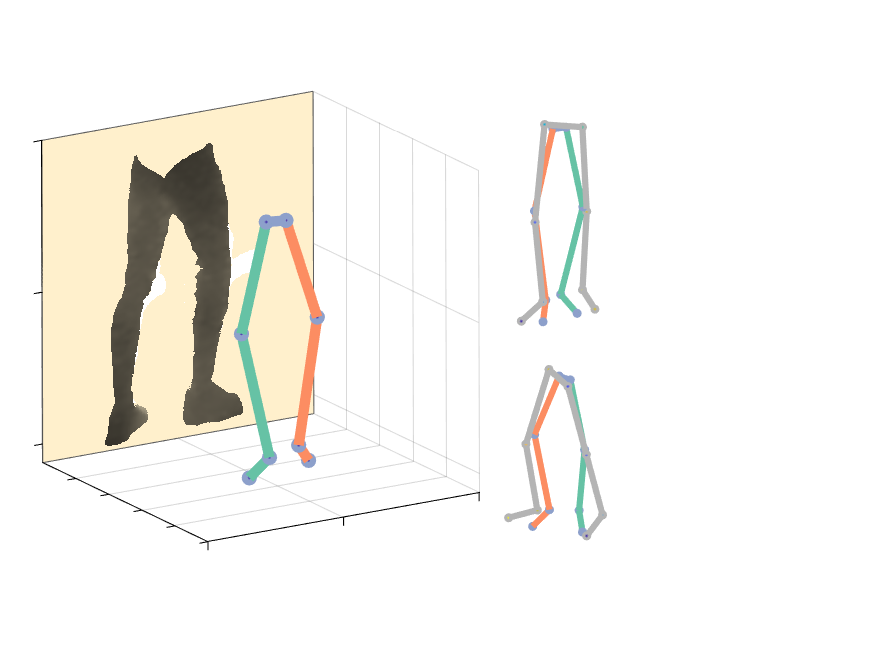} & \IncG[trim={0 0 4.2cm 0},clip,width=\linewidth]{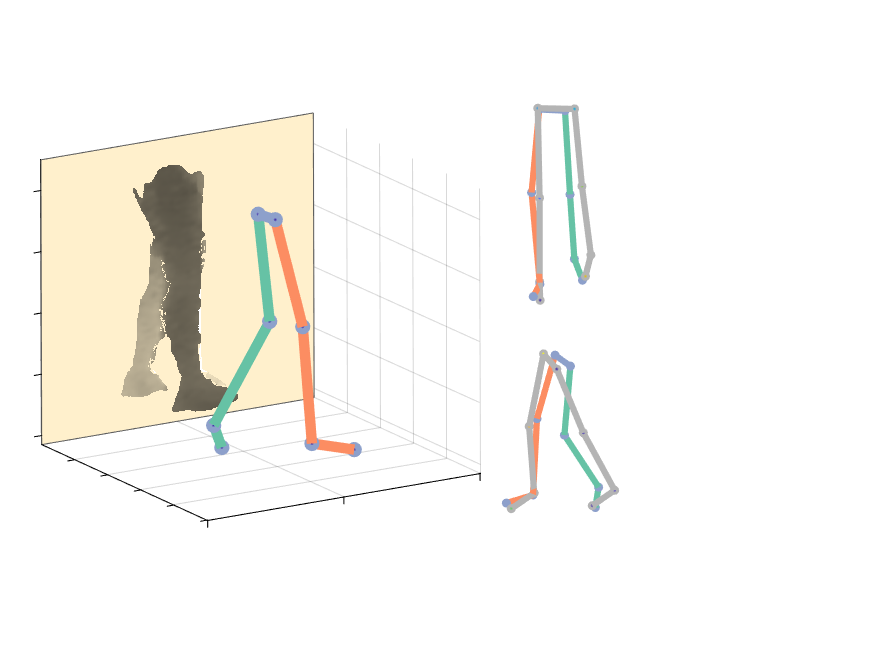} &
     \\ [-1.8em]
          \footnotesize{SO-Net} &  \IncG[trim={0 0 4.2cm 0},clip,width=\linewidth]{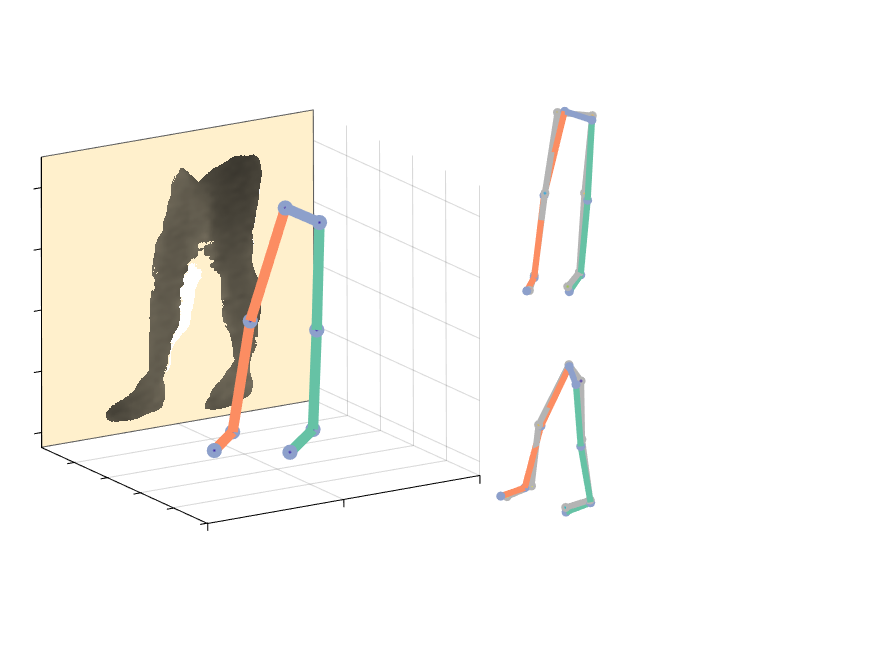} & \IncG[trim={0 0 4.2cm 0},clip,width=\linewidth]{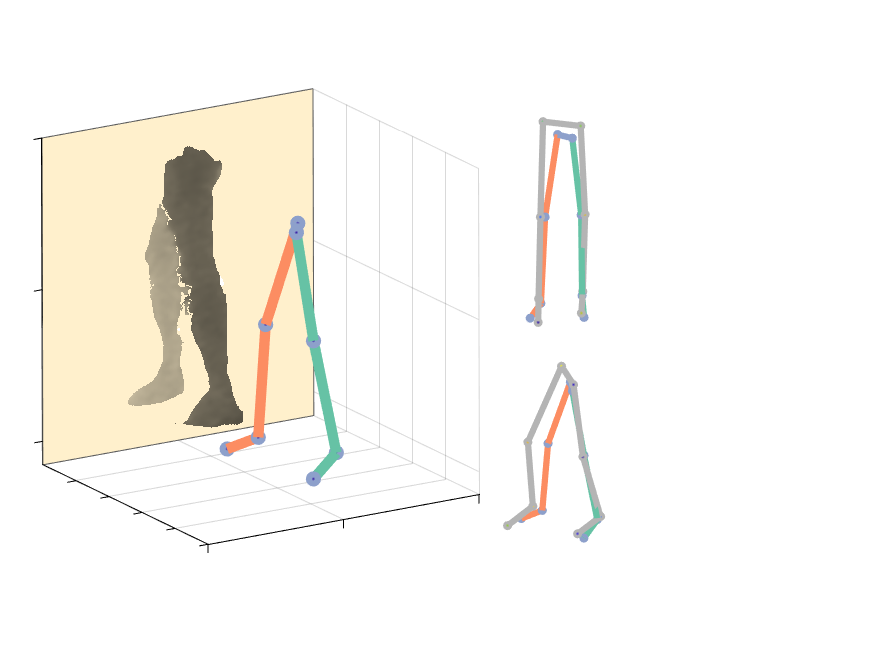} & \IncG[trim={0 0 4.2cm 0},clip,width=\linewidth]{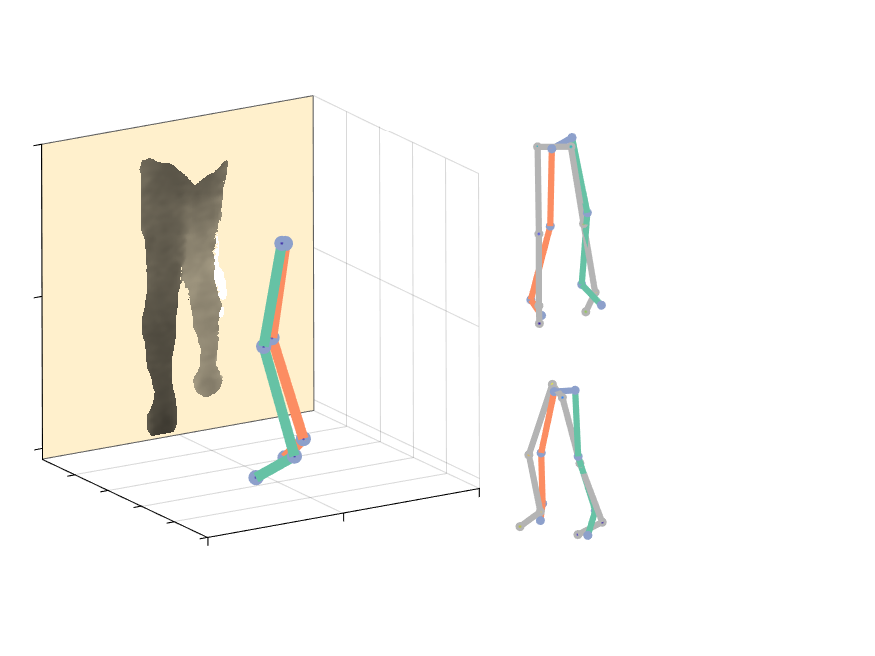} & \IncG[trim={0 0 4.2cm 0},clip,width=\linewidth]{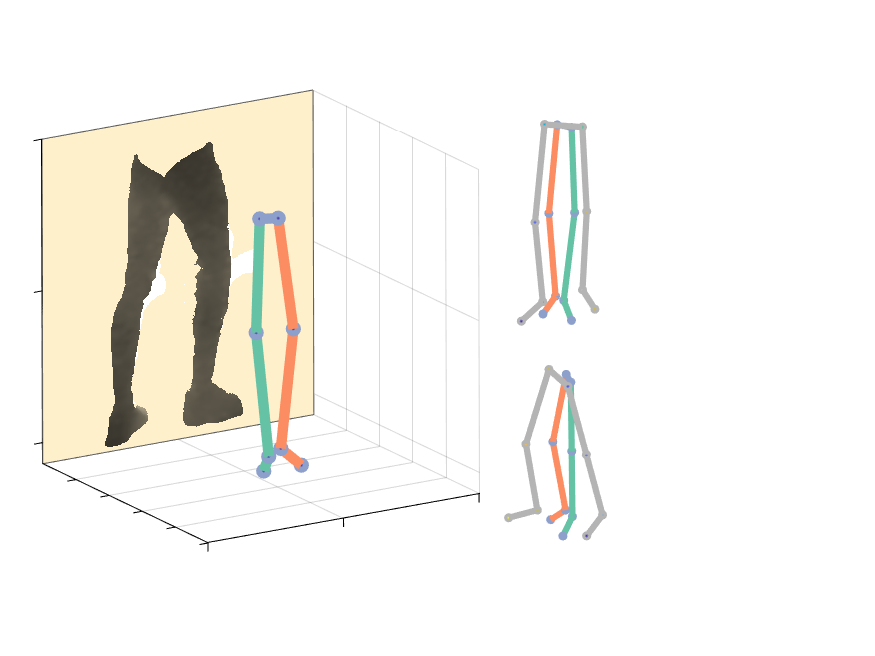} & \IncG[trim={0 0 4.2cm 0},clip,width=\linewidth]{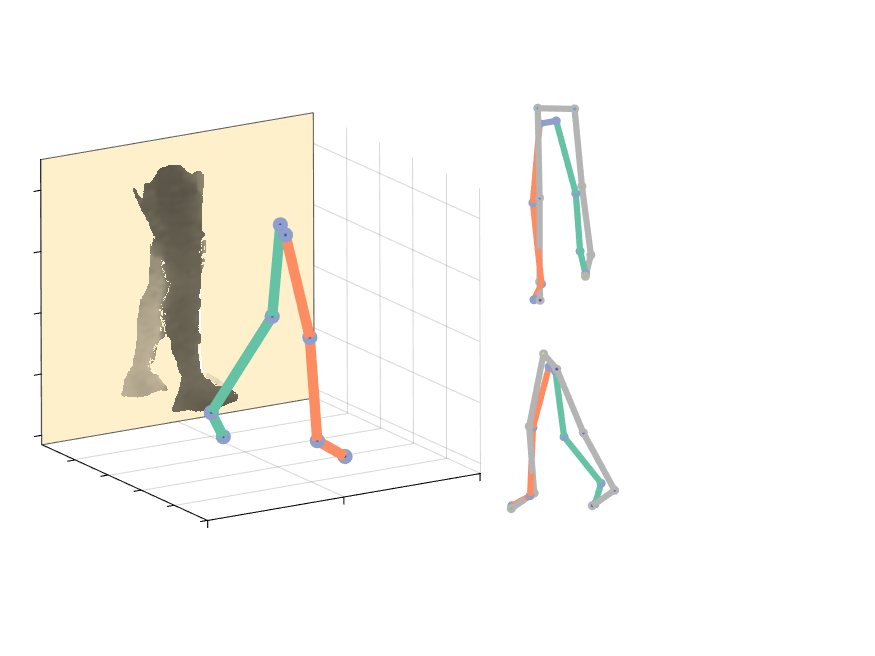} &
     \\ [-1.8em]
      \footnotesize{V2V} & \IncG[trim={0 0 4.2cm 0},clip,width=\linewidth]{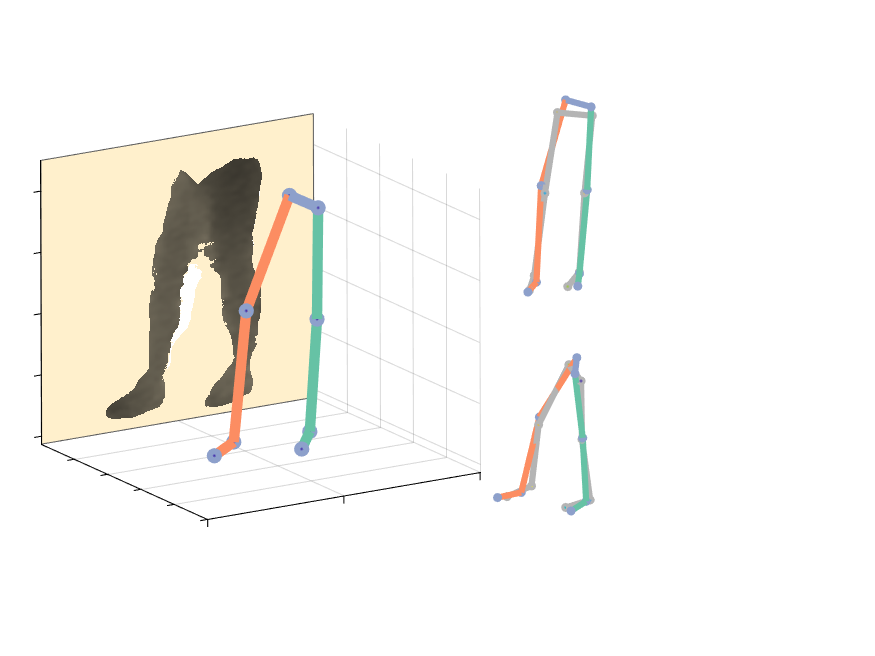} & \IncG[trim={0 0 4.2cm 0},clip,width=\linewidth]{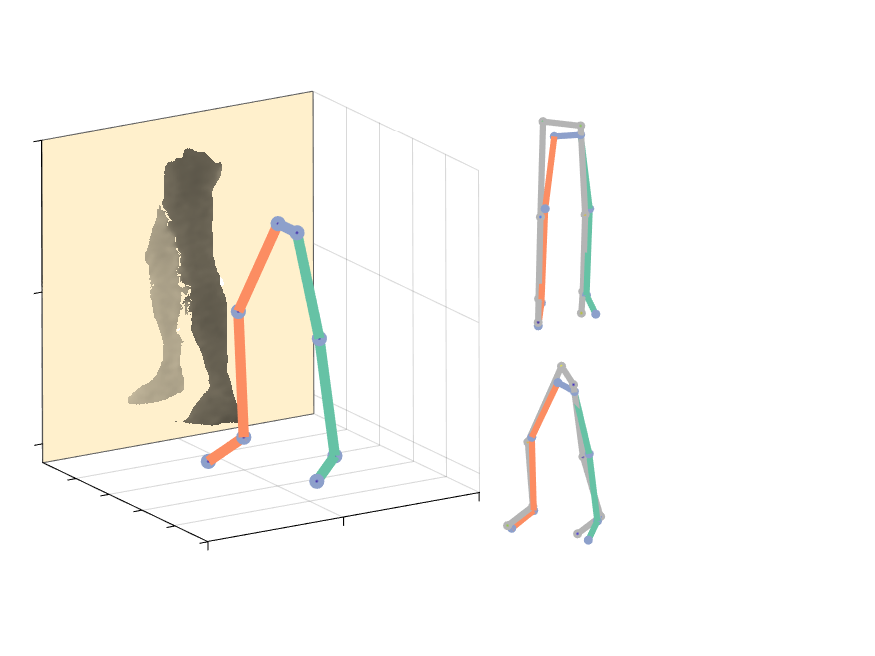} & \IncG[trim={0 0 4.2cm 0},clip,width=\linewidth]{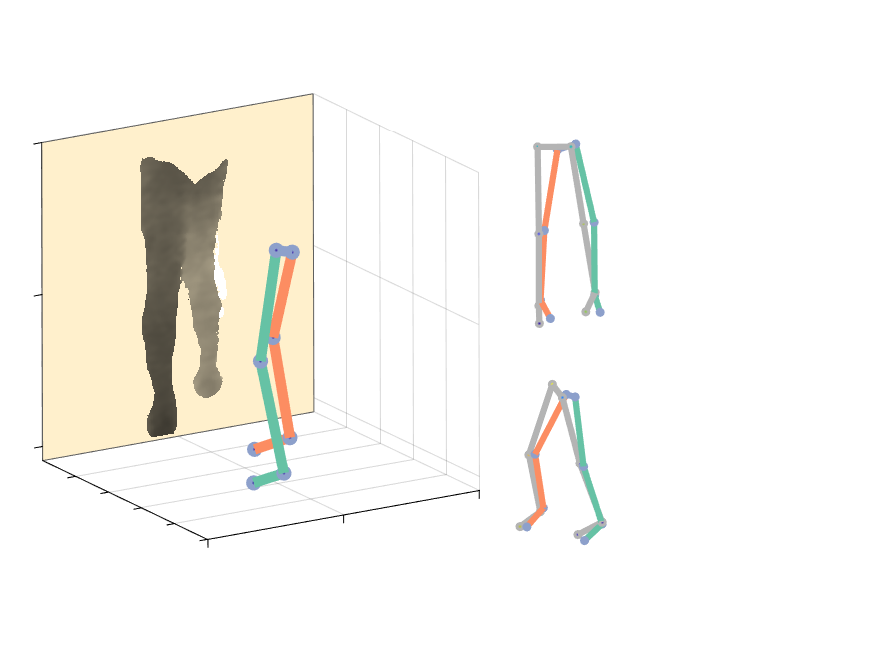} & \IncG[trim={0 0 4.2cm 0},clip,width=\linewidth]{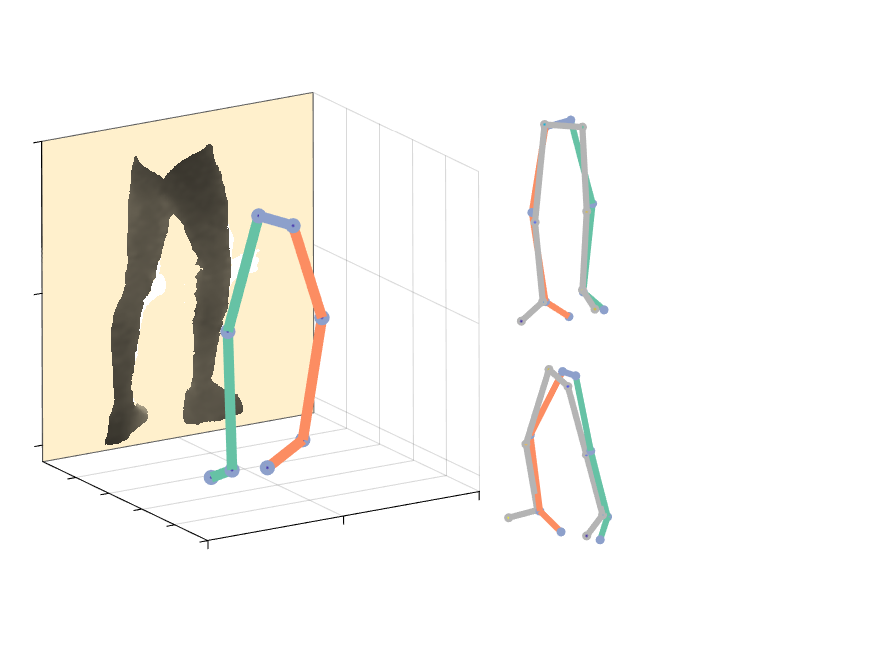} & \IncG[trim={0 0 4.2cm 0},clip,width=\linewidth]{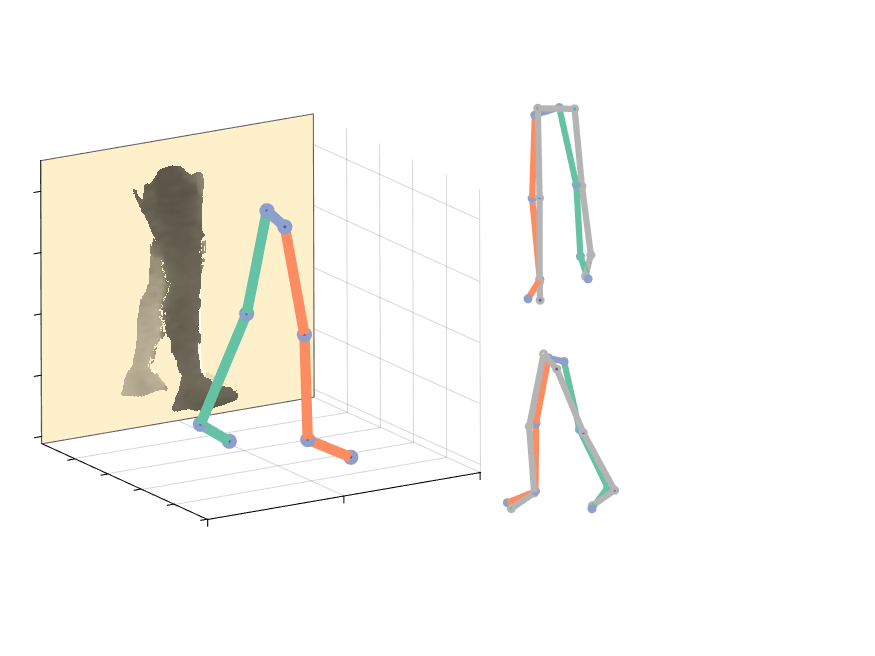} &
    \vspace{-30pt}\multirow{3}{*}{\IncG[width=0.9\linewidth, height=4.5\linewidth]{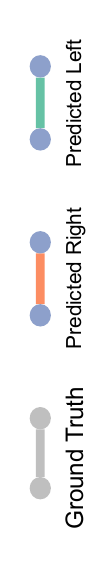}} \\ [-1.8em]
    \footnotesize{Proposed-Mix} & \IncG[trim={0 0 4.2cm 0},clip,width=\linewidth]{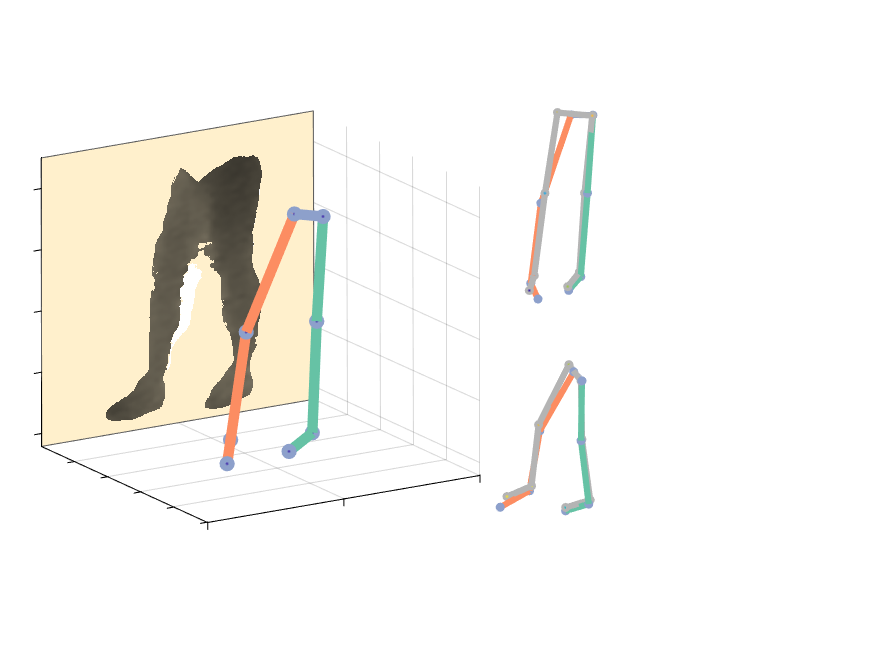} & \IncG[trim={0 0 4.2cm 0},clip,width=\linewidth]{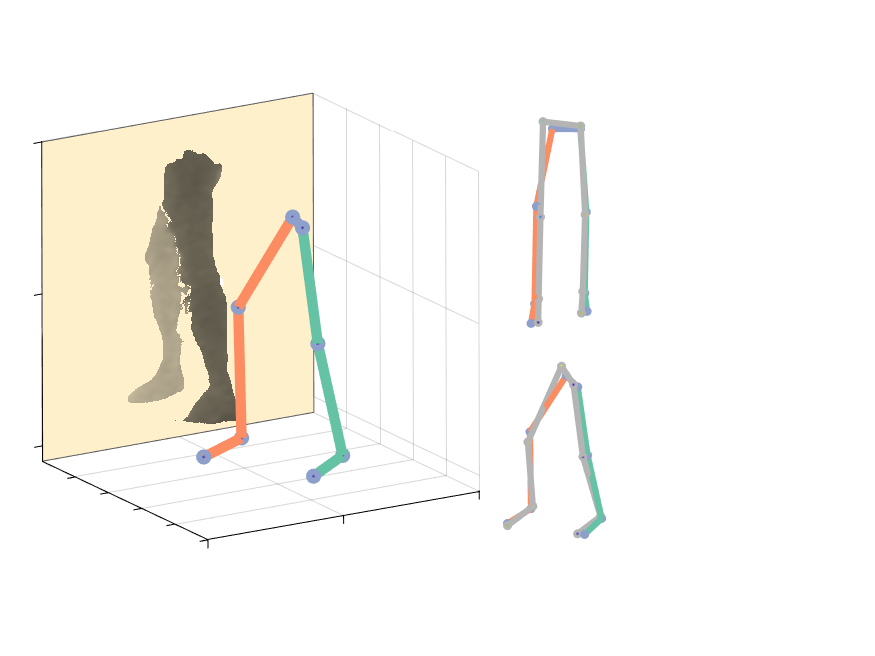} & \IncG[trim={0 0 4.2cm 0},clip,width=\linewidth]{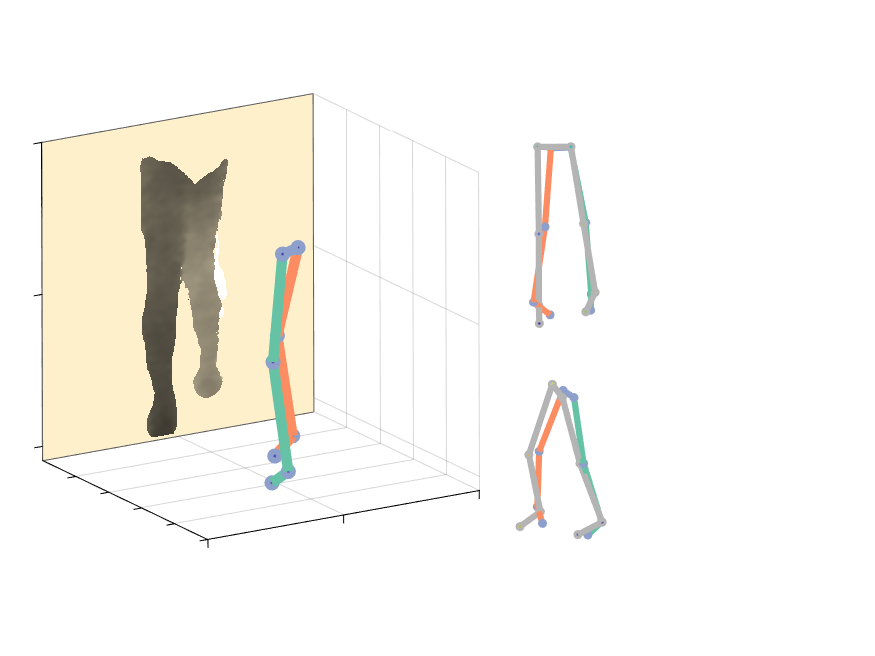} & \IncG[trim={0 0 4.2cm 0},clip,width=\linewidth]{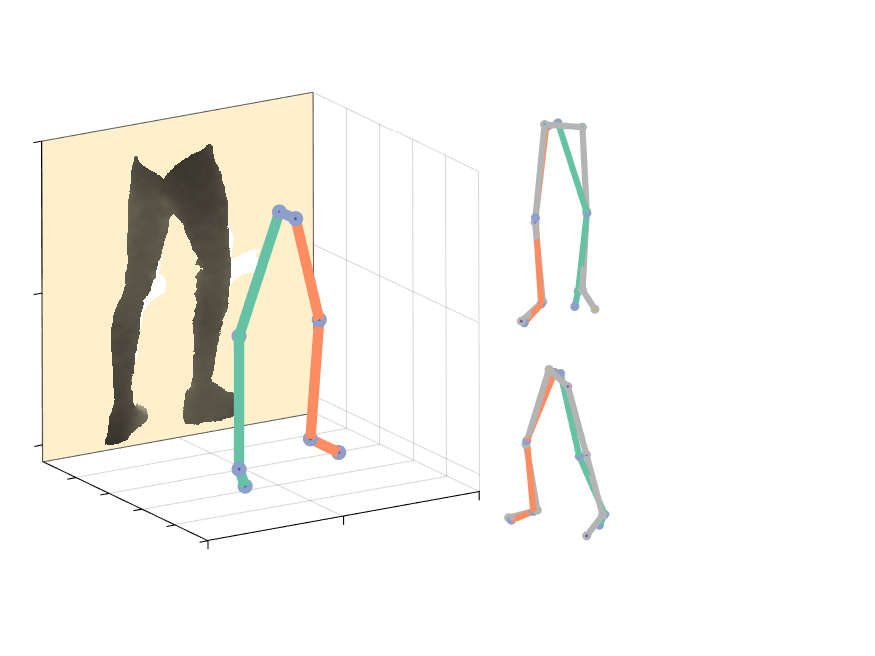} & \IncG[trim={0 0 4.2cm 0},clip,width=\linewidth]{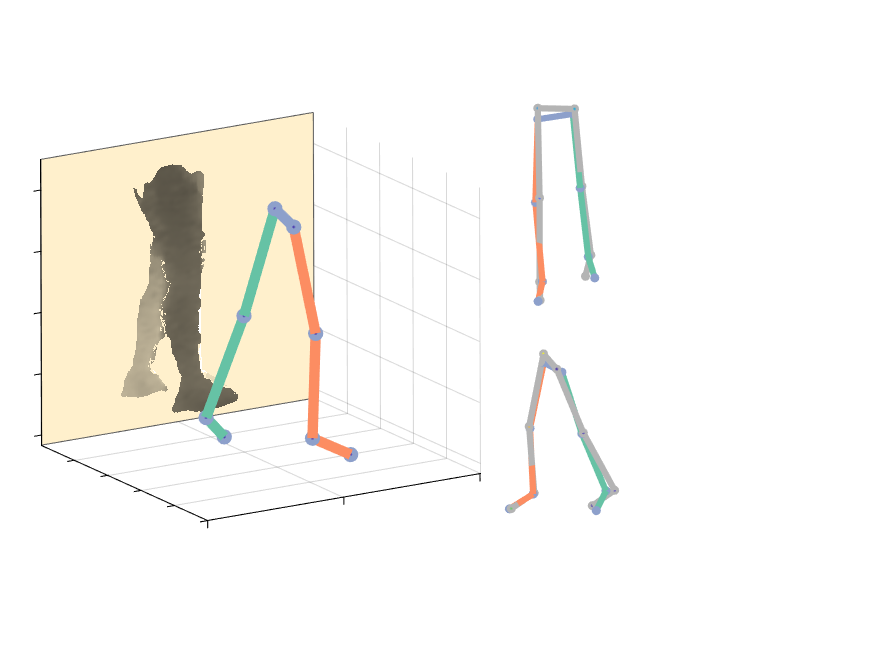} &
   \\ [-1.8em]
    \footnotesize{V2V-Semi} & \IncG[trim={0 0 4.2cm 0},clip,width=\linewidth]{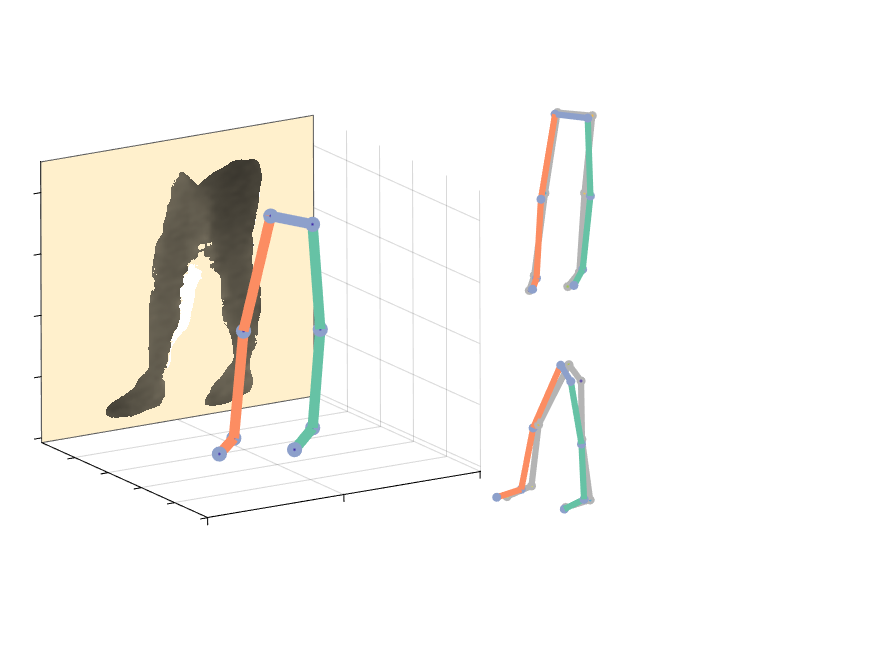} & \IncG[trim={0 0 4.2cm 0},clip,width=\linewidth]{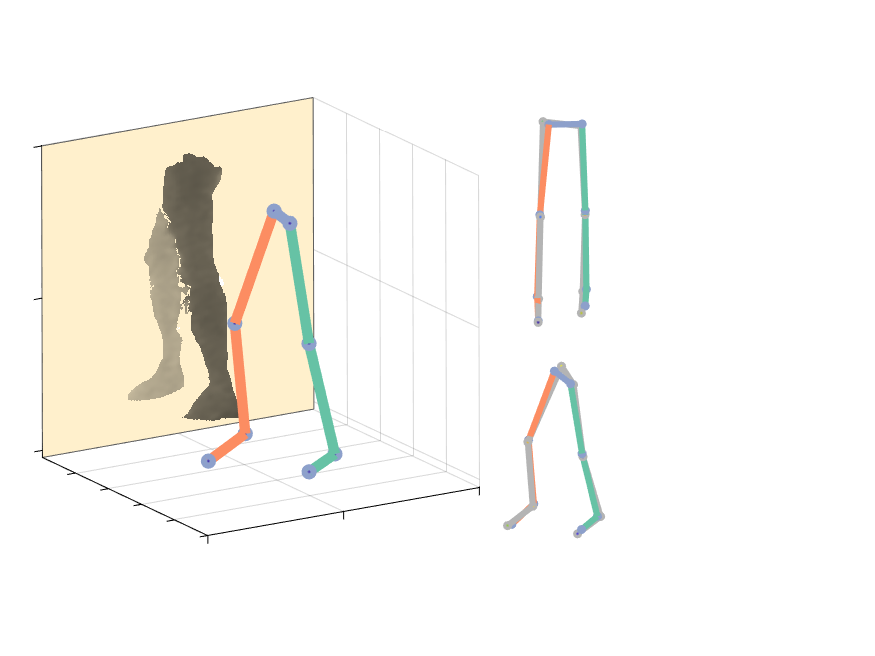} & \IncG[trim={0 0 4.2cm 0},clip,width=\linewidth]{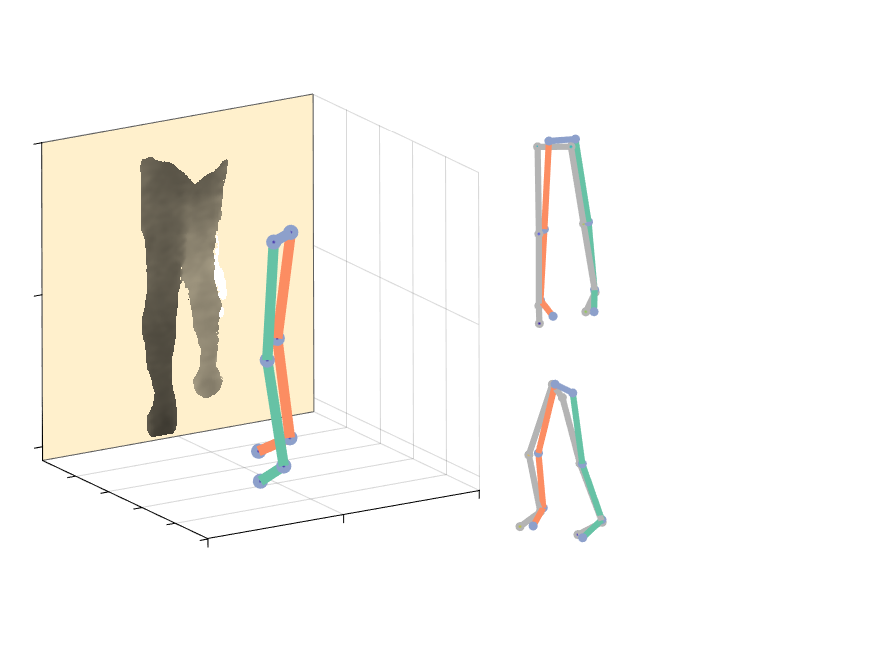} & \IncG[trim={0 0 4.2cm 0},clip,width=\linewidth]{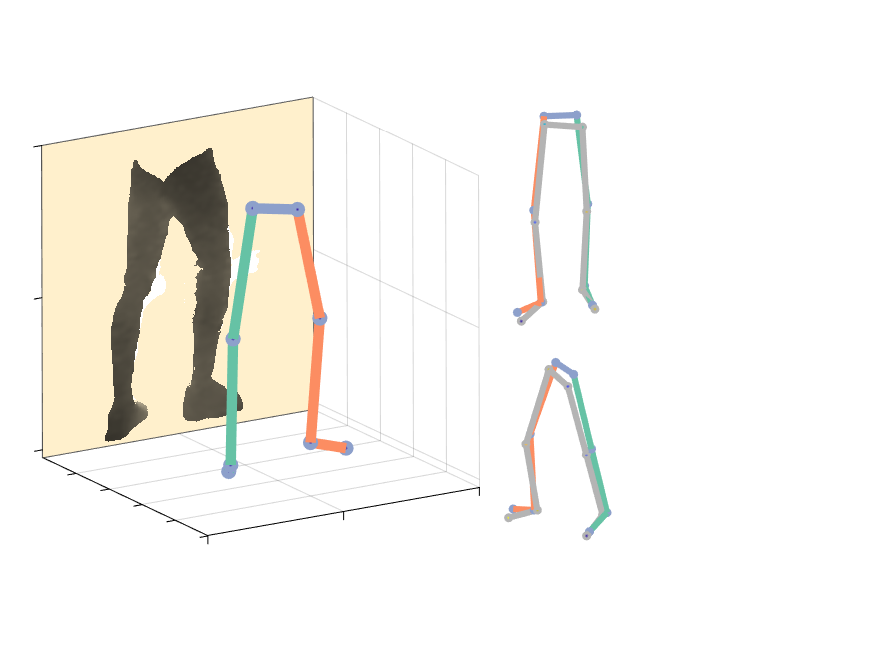} & \IncG[trim={0 0 4.2cm 0},clip,width=\linewidth]{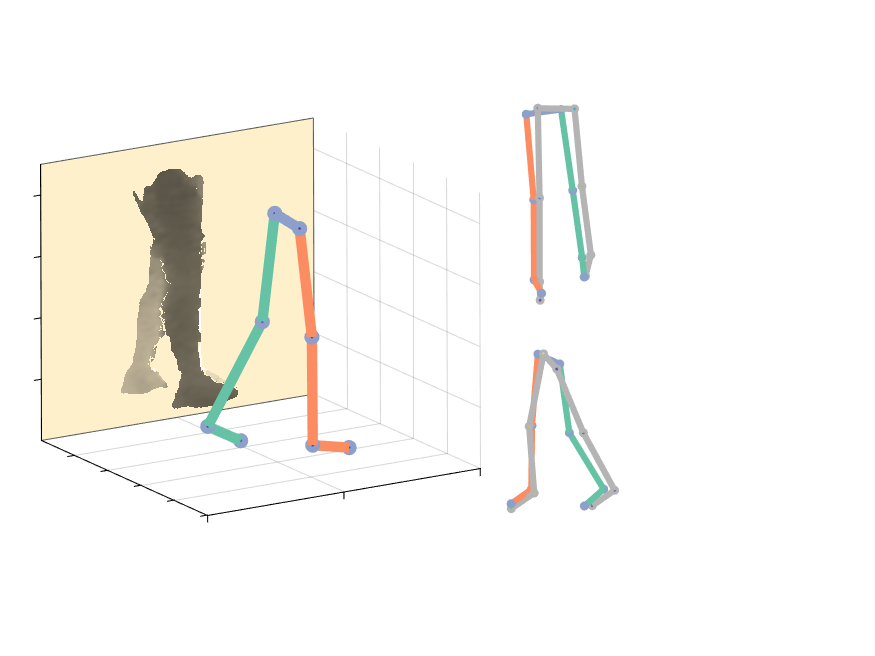} &
   \\ [-1.8em]
    \footnotesize{Proposed} &\IncG[trim={0 0 4.2cm 0},clip,width=\linewidth]{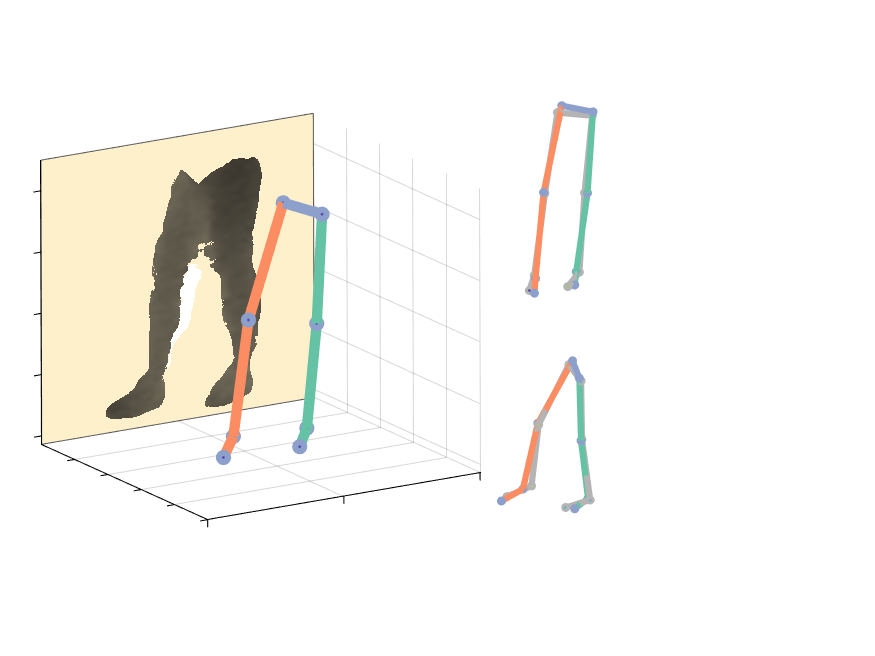} & \IncG[trim={0 0 4.2cm 0},clip,width=\linewidth]{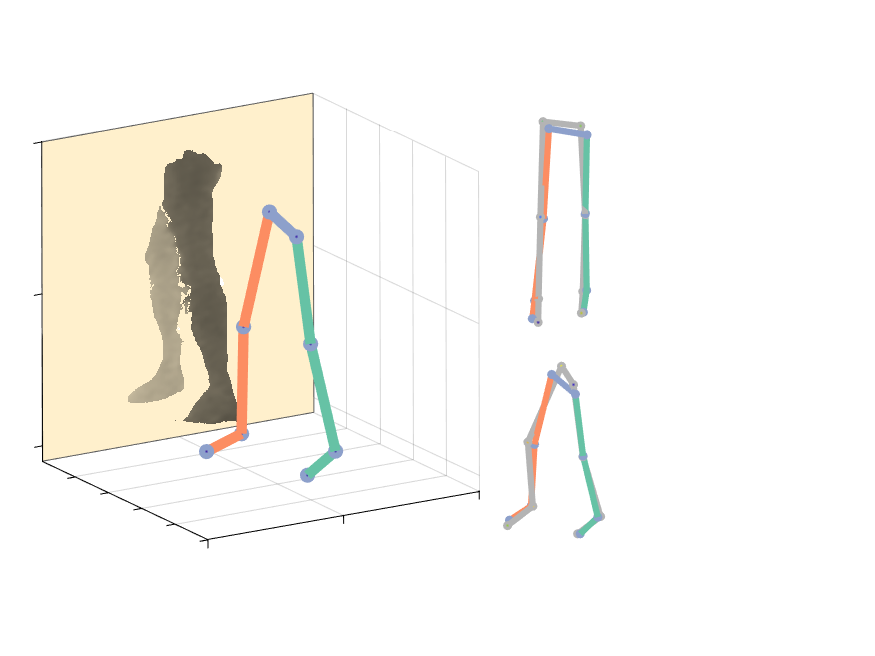} & \IncG[trim={0 0 4.2cm 0},clip,width=\linewidth]{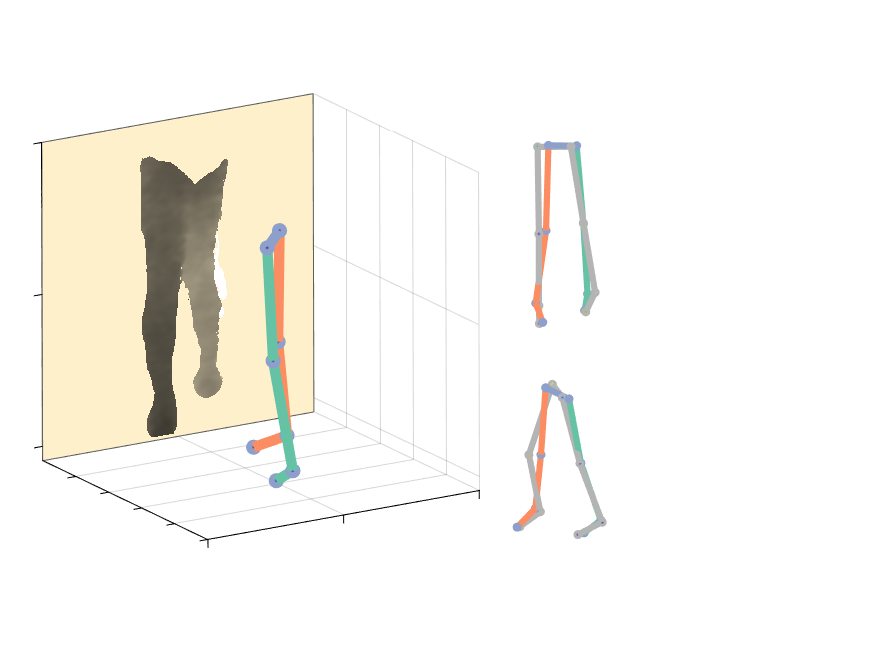} & \IncG[trim={0 0 4.2cm 0},clip,width=\linewidth]{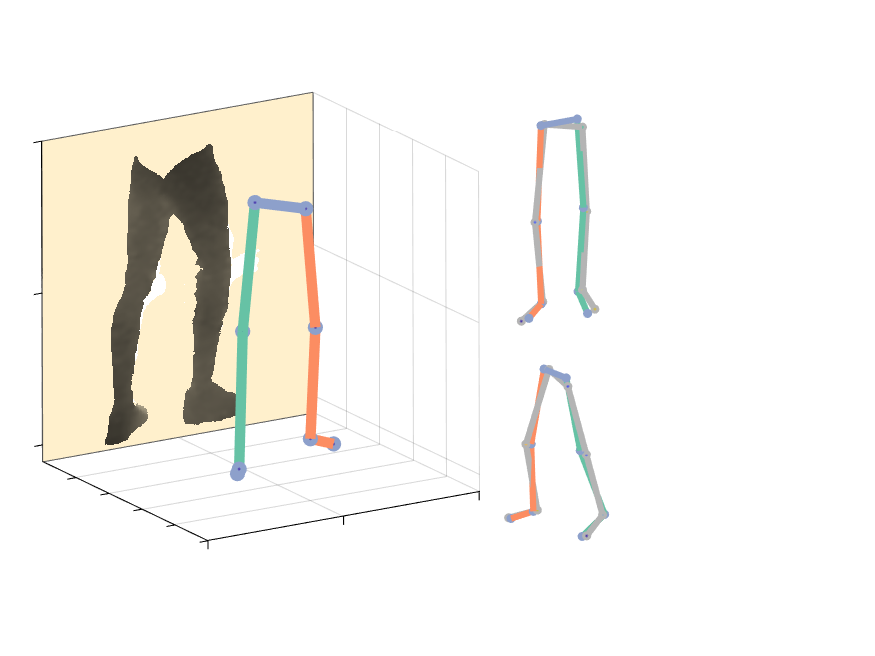} & \IncG[trim={0 0 4.2cm 0},clip,width=\linewidth]{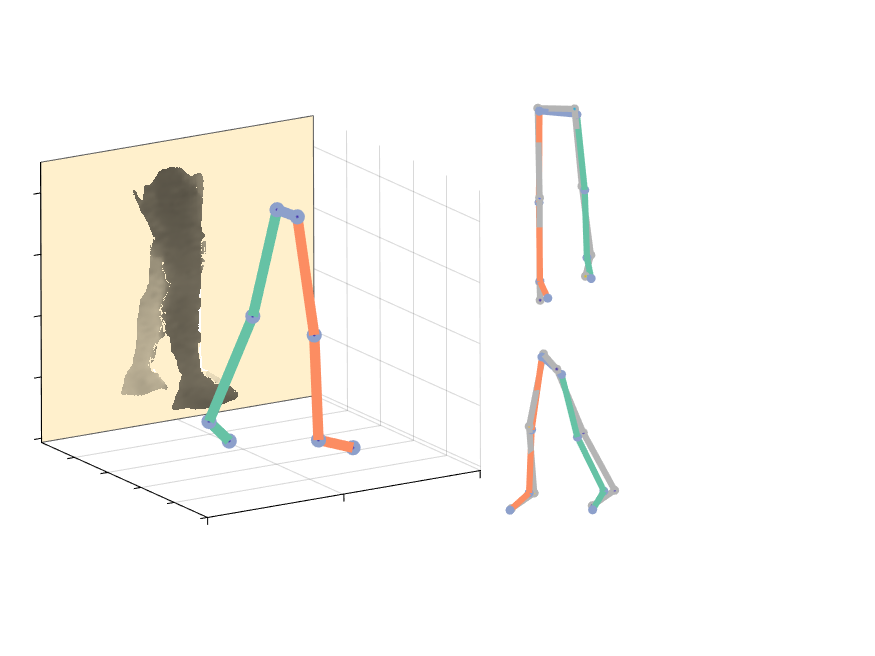} & \\ [-1em]
    & \footnotesize{View1 Supination} & \footnotesize{View2 Pronation} & \footnotesize{View3 Toe-in} & \footnotesize{View4 Toe-out} & \footnotesize{View5 Normal} \\
   % & \multicolumn{6}{c}{\IncG[width=0.4\linewidth]{Figure/legend.pdf}}
    \end{tabular}
    % \includegraphics[trim={0 0 3cm 0},clip,width=0.19\linewidth]{Figure/test.png}
    % \includegraphics[trim={0 0 3cm 0},clip,width=0.19\linewidth,]{Figure/test.png}
    % \includegraphics[trim={0 0 3cm 0},clip,width=0.19\linewidth]{Figure/test.png}
    % \includegraphics[trim={0 0 3cm 0},clip,width=0.19\linewidth]{Figure/test.png}
    % \includegraphics[trim={0 0 3cm 0},clip,width=0.19\linewidth]{Figure/test.png}   
    
    % \includegraphics[trim={0 0 3cm 0},clip,width=0.19\linewidth]{Figure/test.png}
    % \includegraphics[trim={0 0 3cm 0},clip,width=0.19\linewidth,]{Figure/test.png}
    % \includegraphics[trim={0 0 3cm 0},clip,width=0.19\linewidth]{Figure/test.png}
    % \includegraphics[trim={0 0 3cm 0},clip,width=0.19\linewidth]{Figure/test.png}
    % \includegraphics[trim={0 0 3cm 0},clip,width=0.19\linewidth]{Figure/test.png}   
    %\vspace{-5pt}
    \caption{Representative qualitative pose estimation results of different methods (CV-CS validation).}
    \label{fig:qualitative}
    \centering
    \includegraphics[width=0.68\linewidth]{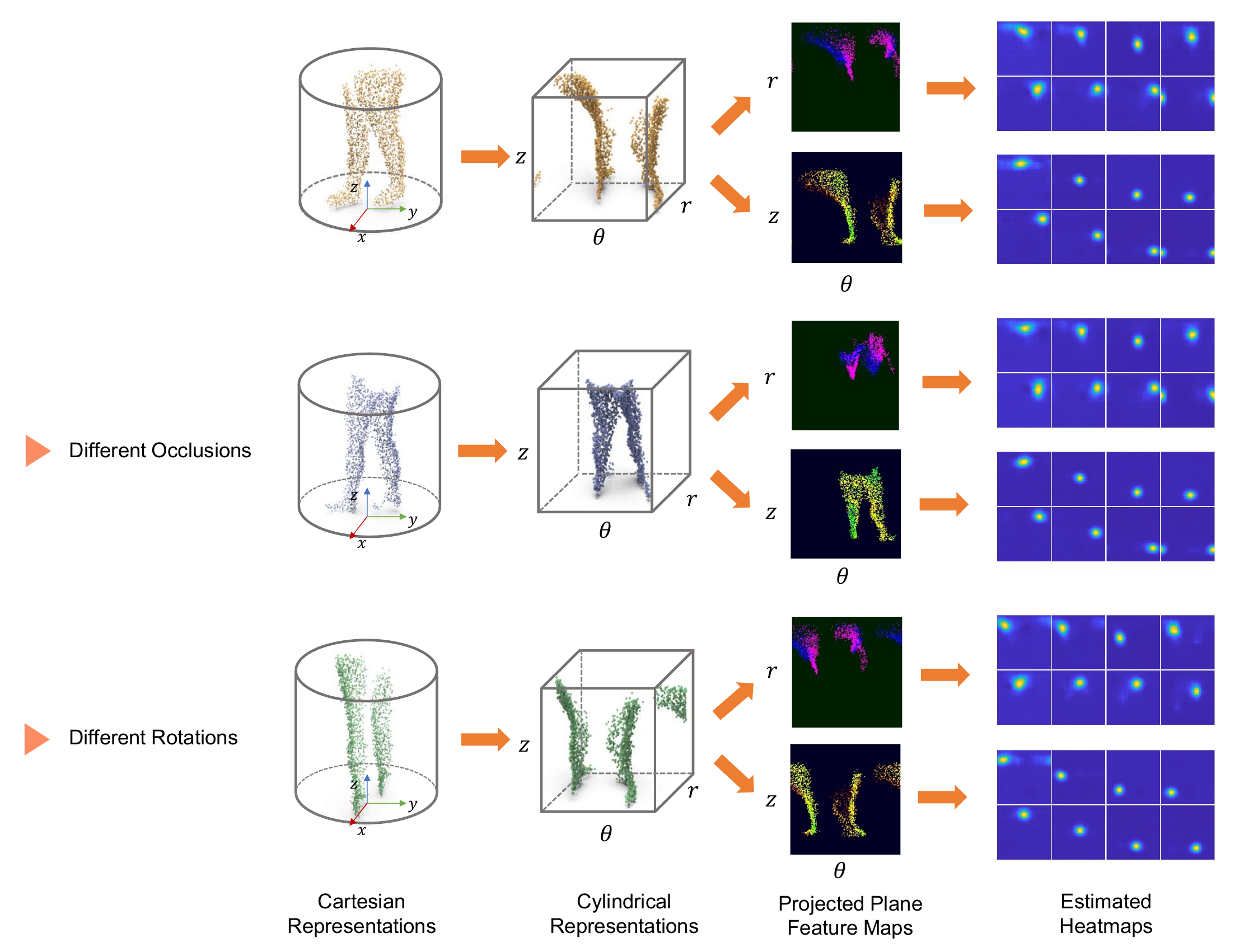}
    \caption{Visualization of the intermediate feature maps and estimated heatmaps with different occlusions or rotations. For each heatmap grid, the top row contains the heatmaps of left hip, knee, ankle, and toe (from leftmost to rightmost) respectively, whereas the bottom row includes the corresponding right keypoint heatmaps. Occlusion-invariance is reflected by the comparison between the $1{st}$ and $2{nd}$ row, whereas rotation-equivariance is reflected by the comparison between the $1{st}$ and $3{rd}$ row.}
    \label{fig:feature}
\end{figure*}

\begin{table*}[h!]
    \centering
\begin{minipage}[b]{0.75\linewidth}
   \footnotesize
   \caption{Quantitative results of cross-view full-subject (CV) validation.}
\begin{tabular}{@{}ccccccccc@{}}
\toprule
\multirow{2}{*}{Methods} & \multicolumn{4}{c}{Dist (cm) $\downarrow$}  & \multicolumn{4}{c}{mAP (5cm) $\uparrow$ }   \\ \cmidrule(l){2-9} 
                         & Hip          & Knee           & Ankle         & Toe         & Hip & Knee & Ankle & Toe   \\ \midrule
PN++~\cite{ge2018hand}     &   8.18$\pm$3.88   &  7.42$\pm$4.38 & 9.69$\pm$6.04 & 10.49$\pm$6.97 & 0.197 & 0.335 & 0.212 & 0.194   \\
SO-Net~\cite{chen2019so} &   8.04$\pm$3.41  &    5.94$\pm$ 4.10      &      7.18$\pm$5.60     &   8.44$\pm$6.57     & 0.166 & 0.524 & 0.440 & 0.308    \\
%A2J~\cite{xiong2019a2j}  &   $\pm$        &     $\pm$      &     $\pm$      &    $\pm$       &     &      &       &     &         &          &         \\
V2V~\cite{moon2018v2v}   &   8.20$\pm$4.33        &     4.56$\pm$2.87      &    4.42$\pm$5.30      &    7.09$\pm$7.92 &  0.219 & 0.688 & 0.787 & 0.487              \\ \midrule
V2V-Semi~\cite{moon2018v2v}   & \underline{6.24}$\pm$4.08 &  \underline{3.94}$\pm$3.01      &     \underline{3.83}$\pm$3.71  &   \underline{4.83}$\pm$5.06     & \underline{0.465} & \underline{0.814} & \underline{0.827} & \textbf{0.699}        \\ \midrule
Proposed-Mix    &  7.76$\pm$4.91  &  5.41$\pm$6.30  &  5.96$\pm$8.35   & 8.73$\pm$10.14  &  0.292 & 0.718 & 0.691 & 0.441 \\
%Proposed-88             &                &               &               &\\
Proposed    &  \textbf{5.00}$\pm$3.32 & \textbf{3.36}$\pm$2.33 & \textbf{3.49}$\pm$2.61 & \textbf{4.67}$\pm$3.31 &  \textbf{0.628}  & \textbf{0.873}  & \textbf{0.842}  & \underline{0.656}       \\ \bottomrule  
    \end{tabular}
    \label{tab:CV_Full}
       \par\vspace{0pt}
    \end{minipage}
    \hspace{-10pt}
\begin{minipage}[b]{0.23\linewidth}
    \includegraphics[width=1.2\linewidth]{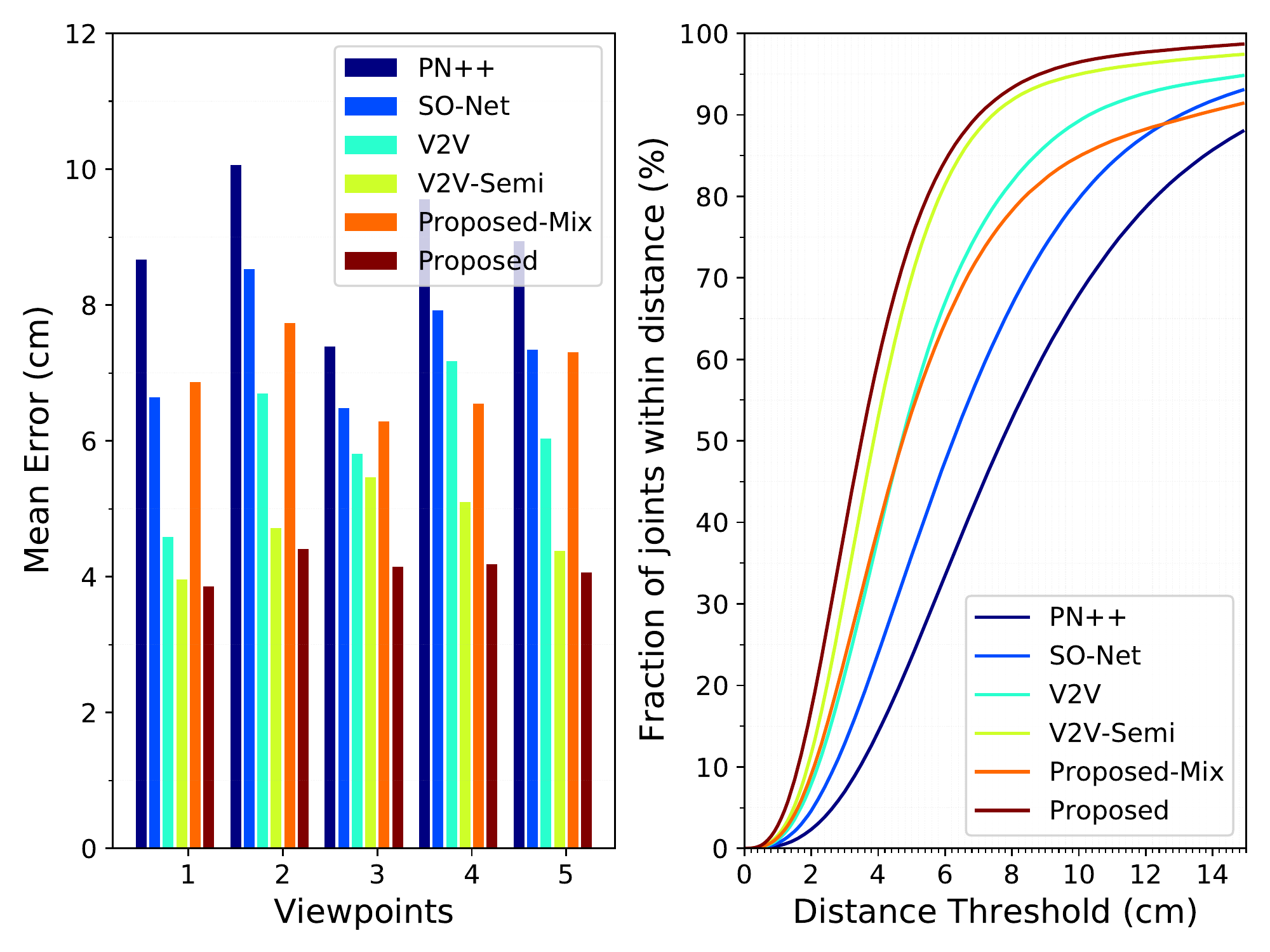}
       \par\vspace{-5pt}
\end{minipage}
\hfill
\vspace{-10pt}
\end{table*}

\begin{table*}[h!]
    \centering
    \begin{minipage}[b]{0.75\linewidth}
   \footnotesize
       \caption{Quantitative results of cross-view cross-subject (CV-CS) validation.}
\begin{tabular}{@{}ccccccccc@{}}
\toprule
\multirow{2}{*}{Methods} & \multicolumn{4}{c}{Dist (cm) $\downarrow$}    & \multicolumn{4}{c}{mAP (5cm) $\uparrow$ }   \\ \cmidrule(l){2-9} 
                         & Hip          & Knee           & Ankle         & Toe          & Hip & Knee & Ankle & Toe    \\ \midrule
PN++~\cite{ge2018hand}   & 8.50$\pm$3.96 &  7.81$\pm$4.45  &  10.30$\pm$6.03      & 11.06$\pm$7.01      &  0.174  &  0.295   &  0.175    & 0.169      \\
SO-Net~\cite{chen2019so} &  8.31$\pm$3.56  &    6.52$\pm$4.26       &      8.43$\pm$6.07     &   9.50$\pm$7.03   & 0.154 &  0.443  & 0.319  & 0.246        \\
%A2J~\cite{xiong2019a2j}  &   $\pm$        &     $\pm$      &     $\pm$      &    $\pm$       &     &      &       &     &         &          &         \\
V2V~\cite{moon2018v2v}    & 8.13$\pm$4.20 & $ 4.61\pm$2.84 & 4.47$\pm$5.30 & 7.09$\pm$7.50 & 0.211 & 0.669 & 0.777 & 0.468 \\ \midrule
V2V-Semi~\cite{moon2018v2v}   &  \underline{6.36}$\pm$3.79  &   \underline{4.07}$\pm$2.83      &     \underline{4.16}$\pm$3.92  &   \underline{5.23}$\pm$4.64       &  \underline{0.408}    &  \underline{0.780}    &  \underline{0.783}     &   \textbf{0.627}    \\ \midrule
Proposed-Mix & 7.99$\pm$7.49 & 5.04$\pm$4.99 & 5.46$\pm$7.93 & 8.82$\pm$1.08 & 0.316 & 0.695 & 0.708 & 0.423 \\
%Proposed-88 &&&&& \\
Proposed                 & \textbf{5.51}$\pm$3.70 & \textbf{3.69}$\pm$2.91 & \textbf{3.68}$\pm$2.99 & \textbf{5.09}$\pm$3.81 &  \textbf{0.563}  &  \textbf{0.832}  & \textbf{0.824}  & \underline{0.609}      \\ \bottomrule
    \end{tabular}
    \label{tab:CV_CS}
     \par\vspace{0pt}
    \end{minipage}
    \hspace{-10pt}
    \begin{minipage}[b]{0.23\linewidth}
        \includegraphics[width=1.2\linewidth]{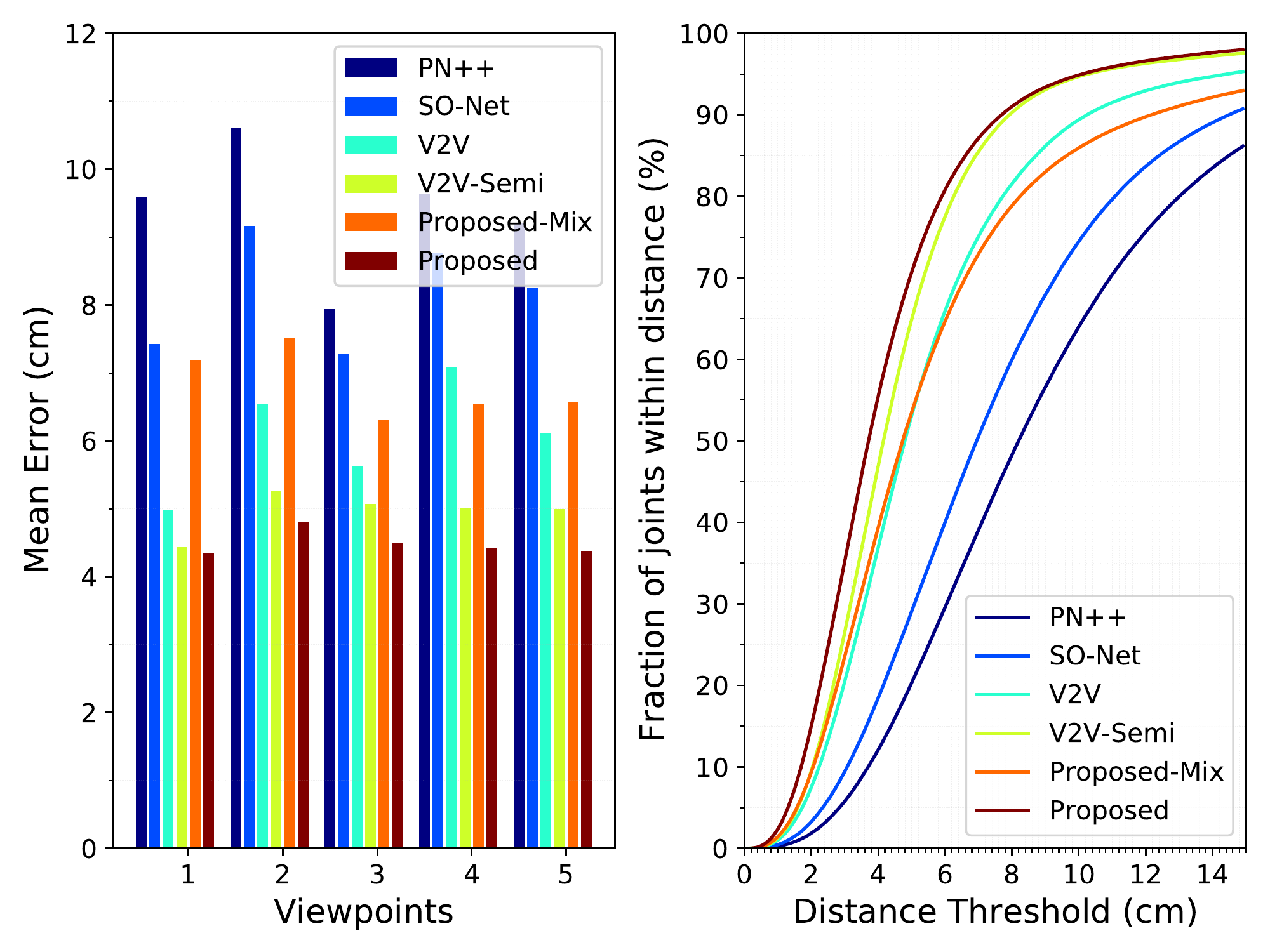}
     \par\vspace{-5pt}
    \end{minipage}
    \hfill
\vspace{-10pt}
\end{table*}

\begin{figure}
\centering
\begin{minipage}[]{0.75\linewidth}
    \centering
    \subfigure[V2V-Semi]{
   \includegraphics[width=0.45\linewidth]{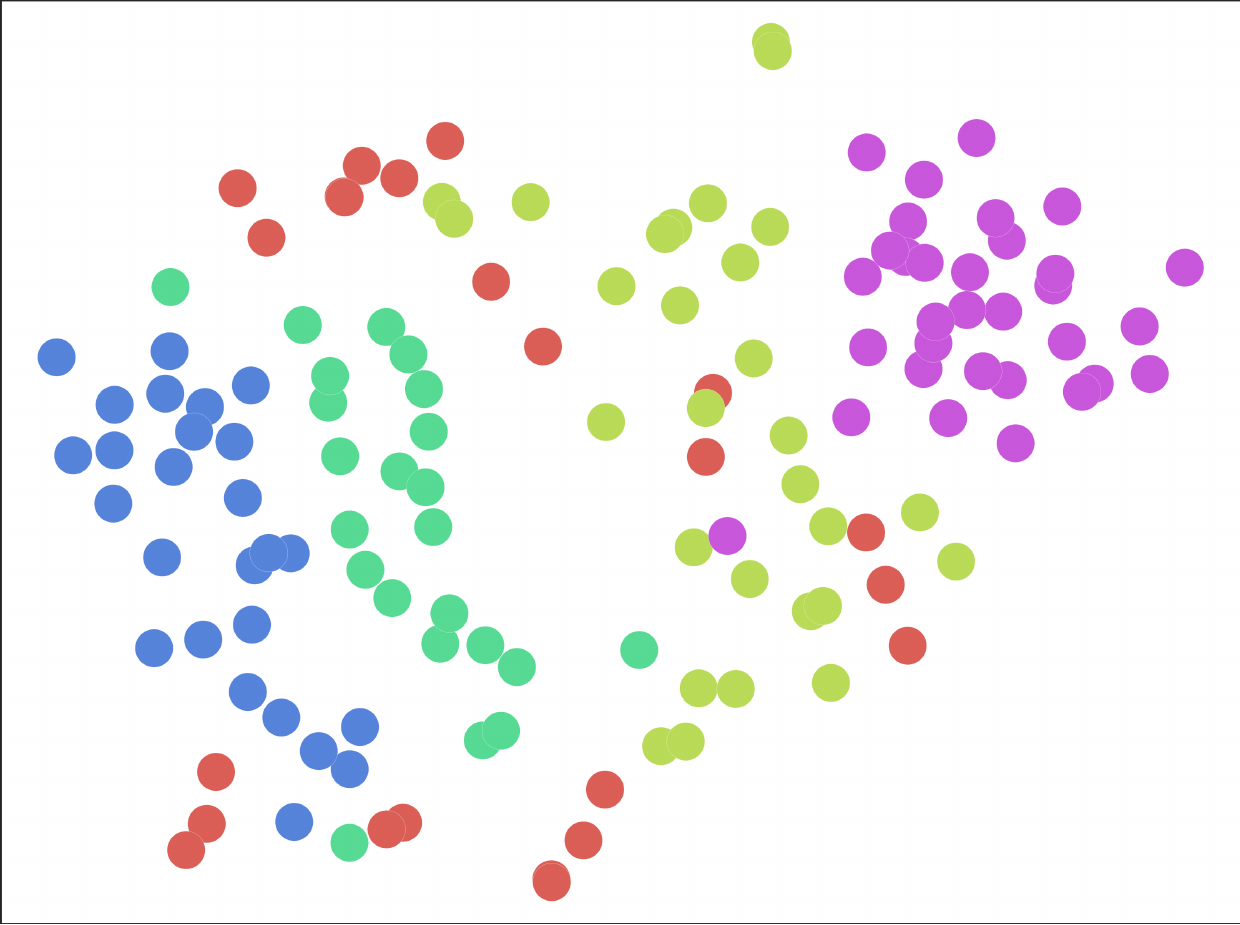}}
   \subfigure[Proposed]{
   \includegraphics[width=0.45\linewidth]{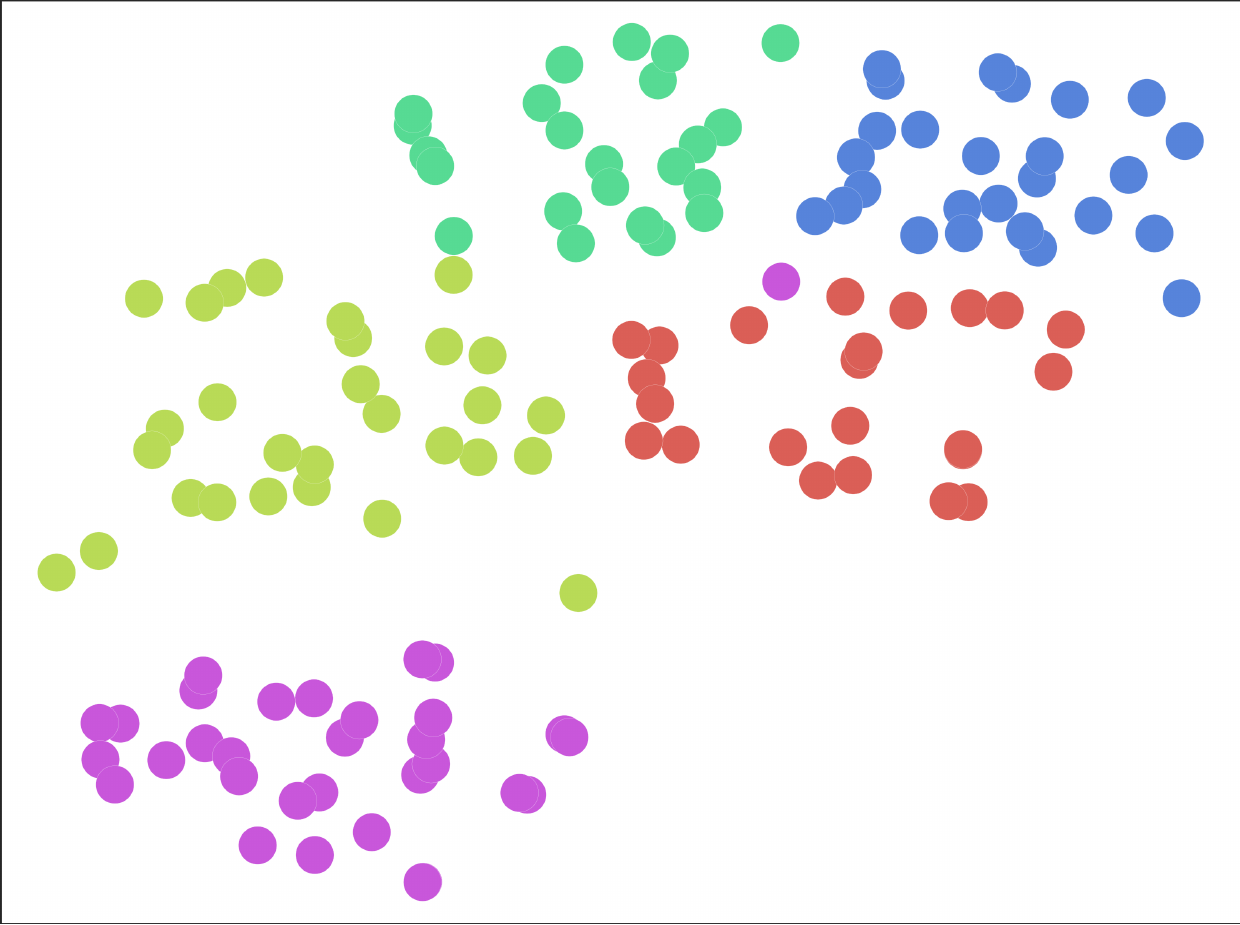}}
\end{minipage}
\begin{minipage}[]{0.22\linewidth}
\includegraphics[trim={0 0 0 0.5cm},clip,width=\linewidth]{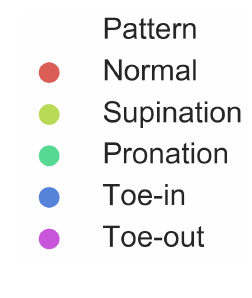}
\end{minipage}
\caption{t-SNE plot of gait cycle samples of one representative subject across different walking conditions (CV-CS validation). } 
\label{fig:tsne}
\vspace{-12pt}
\end{figure}

\begin{table}[]
\centering
\begin{minipage}[b]{0.45\linewidth}
\centering
\footnotesize
\tabcaption{Efficiency comparison of backbone models.}
\begin{tabular}{@{}ccc@{}}
\toprule
Methods                  & Time (ms)         \\ \midrule
PN++ \cite{ge2018hand}    &   18.86          \\
SO-Net \cite{chen2019so}  &   13.27          \\
A2J \cite{xiong2019a2j}   &   33.72          \\
V2V \cite{moon2018v2v}    &   43.41          \\ \midrule
Proposed                 &    29.40           \\ \bottomrule
\end{tabular}
\label{tab:efficiency}
\par\vspace{0pt}
\end{minipage}
\begin{minipage}[b]{0.53\linewidth}
\includegraphics[width=\linewidth]{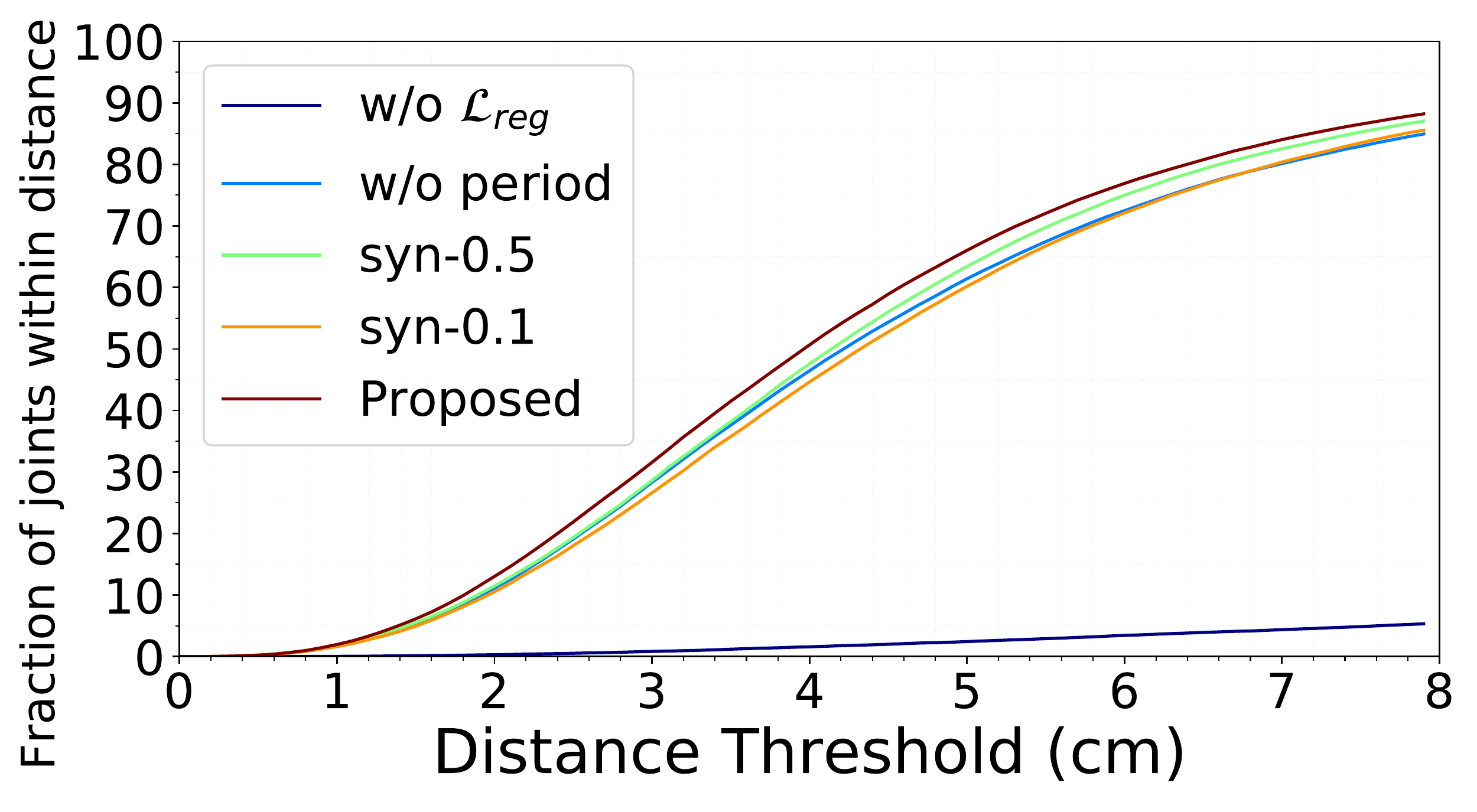}
\figcaption{Comparison results of our ablation study.}
\label{fig:ablation}
\par\vspace{0pt}
\end{minipage}
\vfill
\vspace{-18pt}
\end{table}

\vspace{-10pt}
\subsection{Qualitative Results}
\paragraph{Pose Estimation Results}
In Fig.~\ref{fig:qualitative}, qualitative pose estimation results from CV-CS validation are displayed. We selected results from different conditions \textbf{C} and viewpoints \textbf{V} to make a comprehensive comparison. It is shown that our method can generate visually better results in most cases, and is more robust against occlusions. %\textcolor{red}{For some cases like View 2 Supination, V2V-Semi could even generate some strange hip positions, due to the rotational biases introduced by the cross-view validation settings.}

\paragraph{Visualizations of occlusion-invariance and rotation -equivariance}
The features from intermediate layers between the Anisotropic Conv3D and PoseRes2D, as well as the final estimated heatmap are displayed in Fig.~\ref{fig:feature}. 
It presents the results of {same occlusions \& different rotations} and {different occlusions \& same rotations}. 
As shown in Fig.~\ref{fig:feature}, for different occlusions, the generated heatmaps keep consistent, whereas for different rotations, the generated heatmap is periodically shifted across the $\theta$ axis as compared to the original one. 

\paragraph{t-SNE Plot across Walking Conditions \textbf{C}}
To validate that the features unique to each $C$ are preserved during pose estimation and to avoid the heterogeneity caused by individual differences~\cite{gu2020cross}, we applied our model from CV-CS validation on the complete data of one representative subject. The estimated skeletons were rotated to canonical coordinates, segmented into each gait cycle (from right heel-strike to next heel-strike), and resampled to the same length, similar to~\cite{gu2020cross}. We directly performed t-distributed stochastic neighbor embedding (t-SNE) on the extracted gait cycle samples, and the results are shown in Fig.~\ref{fig:tsne}. It can be observed that our model can generate slightly better results compared to V2V-Semi, in terms of the discernment of estimated skeletons from different \textbf{C}. Although as observed in Fig.~\ref{fig:qualitative} the proposed method demonstrates only slight visual performance gains compared to V2V-Semi, these subtle differences would influence the discernment of different gait types as shown in Fig.~\ref{fig:tsne}. This emphasizes the necessity of ensuring high precision in biomechanical gait analysis. 

\vspace{-8pt}
\subsection{Other Results}
\begin{figure}
\begin{minipage}[b]{0.6\linewidth}
   \centering
     \begin{tabular}{P{.33\linewidth}P{.33\linewidth}P{.2\linewidth}}
     \IncG[trim={4cm 1.3cm 3.5cm 1cm},clip, width=\linewidth, height=\linewidth]{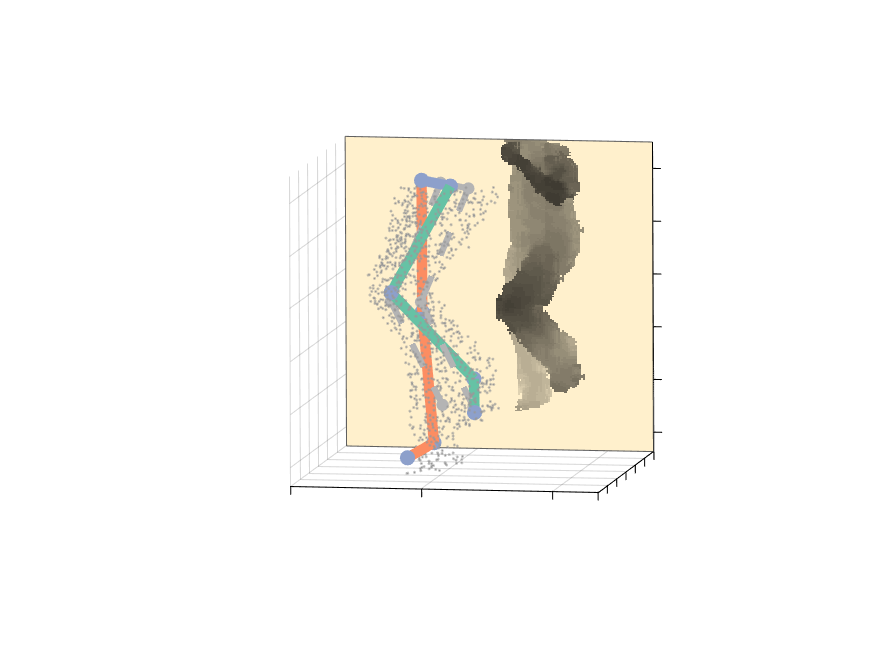} &
    \IncG[trim={4cm 1cm 3.5cm 1cm},clip, width=\linewidth, height=\linewidth]{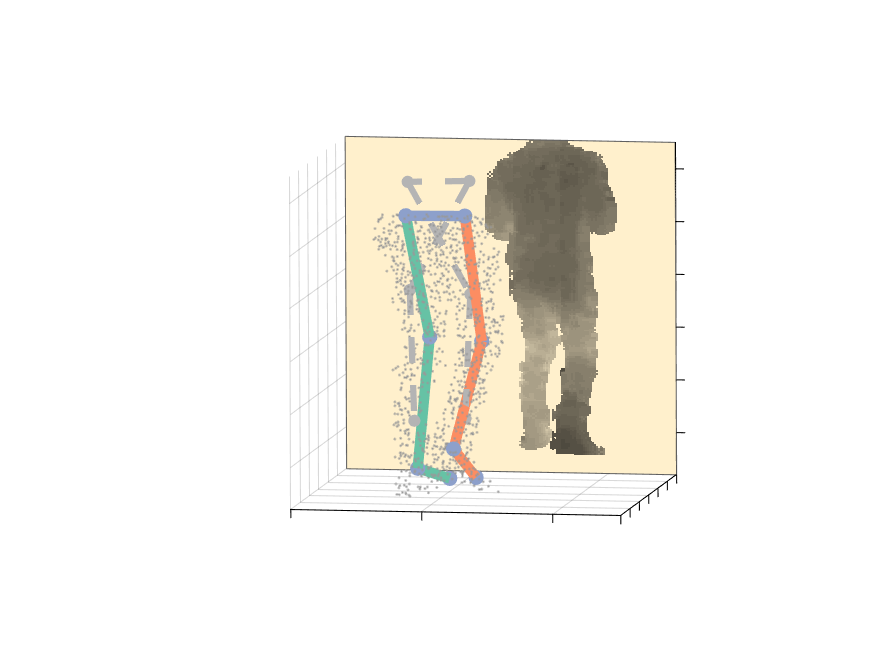} & \multirow{2}{*}{\IncG[trim={0.3cm 0cm 0 0.3cm}, clip, width=0.5\linewidth,
    height=2.5\linewidth]{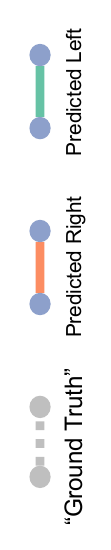}}    \\
    \IncG[trim={4cm 1.3cm 3.5cm 1cm},clip, width=\linewidth,
    height=\linewidth]{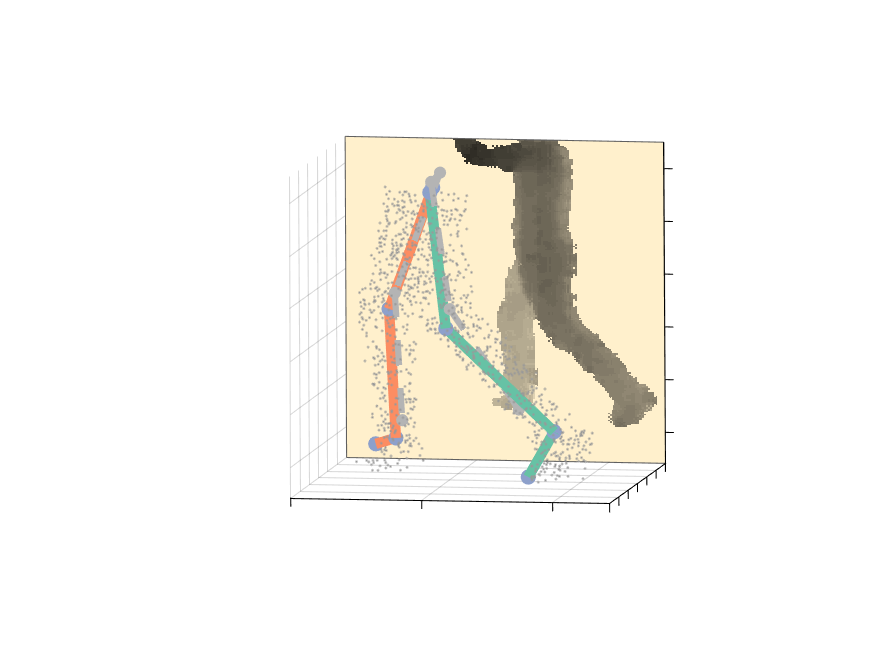} &
    \IncG[trim={4cm 1cm 3.5cm 1cm},clip, width=\linewidth,
    height=\linewidth]{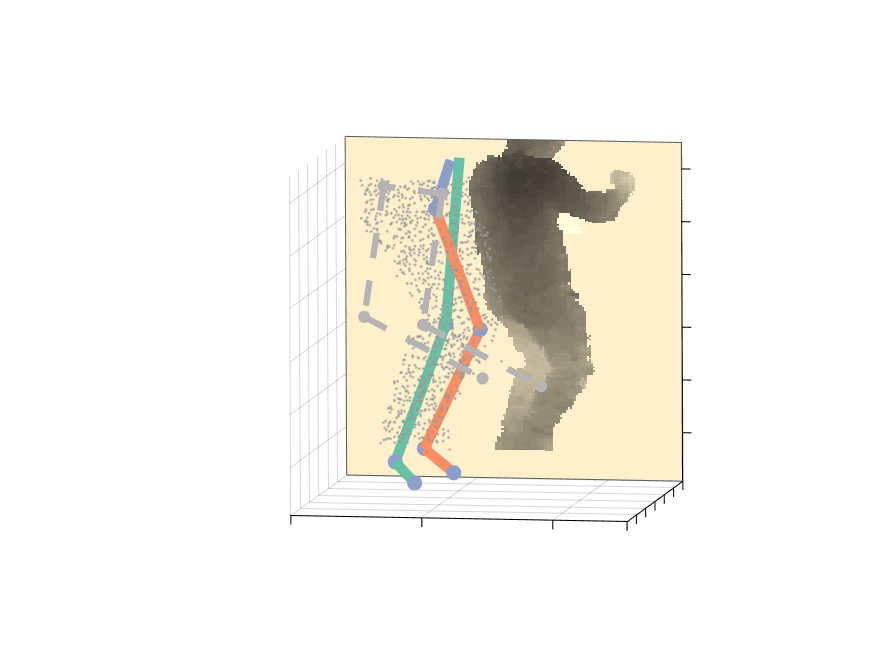}
    \hfill
    \end{tabular}
   \caption{Qualitative results of our model on ITOP dataset.}    
    \label{fig:itop}
    \par\vspace{0pt}
\end{minipage}
\hfill
\begin{minipage}[b]{0.35\linewidth}
  \centering
    \footnotesize
    \tabcaption{Quantitative Results of ITOP Side-Test subset based on its provided `ground truth' skeleton data.}
    \begin{tabular}{@{}cccc@{}}
       \toprule
       Joint & mAP-15cm \\
       \midrule
       Hip   & 0.854    \\
       Knee  & 0.907    \\
       Ankle & 0.894    \\
       \bottomrule
    \end{tabular}
    \label{tab:itop}
    \par\vspace{0pt}
\end{minipage}
\hfill
\vspace{-15pt}
\end{figure}

\paragraph{Computation Efficiency Comparison}
The comparison of the testing time for each sample is presented in Table~\ref{tab:efficiency}, which highlights the efficiency of our model compared to V2V. 
% \subsubsection{Cylindrical Representations}
% \subsubsection{Fraction Number of Synthetic Data}
\paragraph{Results on ITOP Dataset}
We reported the results from Side-Test set of ITOP dataset to demonstrate the generalization capability of our method. The ``ground truth'' in ITOP dataset is semi-manually annotated and does not include toe points. Some qualitative results are displayed in Fig.~\ref{fig:itop}. We also superimpose the `ground truth' (gray dashed line) for comparison. It can be observed that our method is able to actually derive the actual joint positions compared to the original labels provided by the dataset. 

Nonetheless, the qualitative results based on its given ``ground truth'' are reported in terms of the metric mAP, which is a common metric associated with this dataset \cite{haque2016towards}. Given the low quality and bias of ITOP GT (see Fig.~\ref{fig:itop}), we apply mAP(15cm) as the metric. They are all around 90\% for the mAP(15cm) metric. This is consistent with the qualitative results, demonstrating better quality of our estimated lower limb keypoints even without training on this specific dataset. 

\paragraph{Ablation Study}
Here we show the results of without applying $\mathcal{L}_{reg}$ (w/o $\mathcal{L}_{reg}$) and of applying default zero padding instead of periodic padding (i.e., w/o period). We also compared the results of different ratios (0.1 \& 0.5) of synthetic data for training. They are compared under one session of CV-CS validation. The better performance of our model compared to these ablated ones can be observed in Fig.~\ref{fig:ablation}. The extremely low performance of w/o $\mathcal{L}_{reg}$ demonstrates that the model will degenerate to trivial solutions without the constraint enforced by $\mathcal{L}_{reg}$. Periodic padding also plays an important role in our method, as shown by the result of w/o period.

\section{Discussion and Conclusion}
To achieve cross-view generalization for depth based pose estimation, we proposed a novel approach based on a multi-view occlusion-invariant framework built on a novel rotation-equivariant backbone model. The occlusion-invariant framework leverages 3D data with different occlusions to achieve occlusion-invariance learning. The rotation-equivariant backbone converts the rotation group equivariance to translational group equivariance by coordinate transformation. Furthermore, it performs 3D pose estimation with a dual-orthogonal-2D-heatmap estimation method, which is more computationally efficient than conventional 3D convolutions as in the state-of-the-art method. We validated our proposed method in a gait dataset and it is demonstrated that the lower-limb skeleton data could be extracted more precisely with discriminative gait features preserved. 

Although this work focuses on lower limbs, it opens up the possibility of integrating equivariant geometric learning to 3D human pose estimation to enable the generalization in novel viewpoints. Further work should be aimed at further leveraging synthetic data by sim2real adaptation and generalizing our method to 3D rotations, as well as a more general framework for full-body pose estimation.

{\small
\bibliographystyle{IEEEtran}
\bibliography{ref}

\begin{thebibliography}{10}
\providecommand{\url}[1]{#1}
\csname url@samestyle\endcsname
\providecommand{\newblock}{\relax}
\providecommand{\bibinfo}[2]{#2}
\providecommand{\BIBentrySTDinterwordspacing}{\spaceskip=0pt\relax}
\providecommand{\BIBentryALTinterwordstretchfactor}{4}
\providecommand{\BIBentryALTinterwordspacing}{\spaceskip=\fontdimen2\font plus
\BIBentryALTinterwordstretchfactor\fontdimen3\font minus
  \fontdimen4\font\relax}
\providecommand{\BIBforeignlanguage}[2]{{%
\expandafter\ifx\csname l@#1\endcsname\relax
\typeout{** WARNING: IEEEtran.bst: No hyphenation pattern has been}%
\typeout{** loaded for the language `#1'. Using the pattern for}%
\typeout{** the default language instead.}%
\else
\language=\csname l@#1\endcsname
\fi
#2}}
\providecommand{\BIBdecl}{\relax}
\BIBdecl

\bibitem{haque2020illuminating}
A.~Haque, A.~Milstein, and L.~Fei-Fei, ``Illuminating the dark spaces of
  healthcare with ambient intelligence,'' \emph{Nature}, vol. 585, no. 7824,
  pp. 193--202, 2020.

\bibitem{gu2020cross}
X.~Gu, Y.~Guo, F.~Deligianni, B.~Lo, and G.-Z. Yang, ``Cross-subject and
  cross-modal transfer for generalized abnormal gait pattern recognition,''
  \emph{IEEE Transactions on Neural Networks and Learning Systems}, 2020.

\bibitem{deligianni2019emotions}
F.~Deligianni, Y.~Guo, and G.-Z. Yang, ``From emotions to mood disorders: A
  survey on gait analysis methodology,'' \emph{IEEE journal of biomedical and
  health informatics}, vol.~23, no.~6, pp. 2302--2316, 2019.

\bibitem{guo2021mcdcd}
Y.~Guo, X.~Gu, and G.-Z. Yang, ``Mcdcd: Multi-source unsupervised domain
  adaptation for abnormal human gait detection,'' \emph{IEEE Journal of
  Biomedical and Health Informatics}, 2021.

\bibitem{cao2019openpose}
Z.~Cao, G.~Hidalgo, T.~Simon, S.-E. Wei, and Y.~Sheikh, ``Openpose: realtime
  multi-person 2d pose estimation using part affinity fields,'' \emph{IEEE
  transactions on pattern analysis and machine intelligence}, vol.~43, no.~1,
  pp. 172--186, 2019.

\bibitem{li2019crowdpose}
J.~Li, C.~Wang, H.~Zhu, Y.~Mao, H.-S. Fang, and C.~Lu, ``Crowdpose: Efficient
  crowded scenes pose estimation and a new benchmark,'' in \emph{Proceedings of
  the IEEE/CVF Conference on Computer Vision and Pattern Recognition}, 2019,
  pp. 10\,863--10\,872.

\bibitem{iqbal2020weakly}
U.~Iqbal, P.~Molchanov, and J.~Kautz, ``Weakly-supervised 3d human pose
  learning via multi-view images in the wild,'' in \emph{Proceedings of the
  IEEE/CVF Conference on Computer Vision and Pattern Recognition}, 2020, pp.
  5243--5252.

\bibitem{pavllo20193d}
D.~Pavllo, C.~Feichtenhofer, D.~Grangier, and M.~Auli, ``3d human pose
  estimation in video with temporal convolutions and semi-supervised
  training,'' in \emph{Proceedings of the IEEE/CVF Conference on Computer
  Vision and Pattern Recognition}, 2019, pp. 7753--7762.

\bibitem{haque2016towards}
A.~Haque, B.~Peng, Z.~Luo, A.~Alahi, S.~Yeung, and L.~Fei-Fei, ``Towards
  viewpoint invariant 3d human pose estimation,'' in \emph{European Conference
  on Computer Vision}.\hskip 1em plus 0.5em minus 0.4em\relax Springer, 2016,
  pp. 160--177.

\bibitem{armagan2020measuring}
A.~Armagan, G.~Garcia-Hernando, S.~Baek, and T.~K. Kim, ``Measuring
  generalisation to unseen viewpoints, articulations, shapes and objects for 3d
  hand pose estimation under hand-object interaction,'' in \emph{16th European
  Conference on Computer Vision, ECCV 2020}.\hskip 1em plus 0.5em minus
  0.4em\relax CVF, 2020.

\bibitem{varol2017learning}
G.~Varol, J.~Romero, X.~Martin, N.~Mahmood, M.~J. Black, I.~Laptev, and
  C.~Schmid, ``Learning from synthetic humans,'' in \emph{Proceedings of the
  IEEE Conference on Computer Vision and Pattern Recognition}, 2017, pp.
  109--117.

\bibitem{colyer2018review}
S.~L. Colyer, M.~Evans, D.~P. Cosker, and A.~I. Salo, ``A review of the
  evolution of vision-based motion analysis and the integration of advanced
  computer vision methods towards developing a markerless system,''
  \emph{Sports medicine-open}, vol.~4, no.~1, p.~24, 2018.

\bibitem{kainz2017reliability}
H.~Kainz, D.~Graham, J.~Edwards, H.~P. Walsh, S.~Maine, R.~N. Boyd, D.~G.
  Lloyd, L.~Modenese, and C.~P. Carty, ``Reliability of four models for
  clinical gait analysis,'' \emph{Gait \& posture}, vol.~54, pp. 325--331,
  2017.

\bibitem{ionescu2013human3}
C.~Ionescu, D.~Papava, V.~Olaru, and C.~Sminchisescu, ``Human3. 6m: Large scale
  datasets and predictive methods for 3d human sensing in natural
  environments,'' \emph{IEEE transactions on pattern analysis and machine
  intelligence}, vol.~36, no.~7, pp. 1325--1339, 2013.

\bibitem{trumble2017total}
M.~Trumble, A.~Gilbert, C.~Malleson, A.~Hilton, and J.~P. Collomosse, ``Total
  capture: 3d human pose estimation fusing video and inertial sensors.'' in
  \emph{BMVC}, vol.~2, no.~5, 2017, pp. 1--13.

\bibitem{mehta2017monocular}
D.~Mehta, H.~Rhodin, D.~Casas, P.~Fua, O.~Sotnychenko, W.~Xu, and C.~Theobalt,
  ``Monocular 3d human pose estimation in the wild using improved cnn
  supervision,'' in \emph{2017 international conference on 3D vision
  (3DV)}.\hskip 1em plus 0.5em minus 0.4em\relax IEEE, 2017, pp. 506--516.

\bibitem{moon2018v2v}
G.~Moon, J.~Y. Chang, and K.~M. Lee, ``V2v-posenet: Voxel-to-voxel prediction
  network for accurate 3d hand and human pose estimation from a single depth
  map,'' in \emph{Proceedings of the IEEE conference on computer vision and
  pattern Recognition}, 2018, pp. 5079--5088.

\bibitem{qiu2019cross}
H.~Qiu, C.~Wang, J.~Wang, N.~Wang, and W.~Zeng, ``Cross view fusion for 3d
  human pose estimation,'' in \emph{Proceedings of the IEEE/CVF International
  Conference on Computer Vision}, 2019, pp. 4342--4351.

\bibitem{rhodin2018learning}
H.~Rhodin, J.~Sp{\"o}rri, I.~Katircioglu, V.~Constantin, F.~Meyer,
  E.~M{\"u}ller, M.~Salzmann, and P.~Fua, ``Learning monocular 3d human pose
  estimation from multi-view images,'' in \emph{Proceedings of the IEEE
  Conference on Computer Vision and Pattern Recognition}, 2018, pp. 8437--8446.

\bibitem{xiong2019a2j}
F.~Xiong, B.~Zhang, Y.~Xiao, Z.~Cao, T.~Yu, J.~T. Zhou, and J.~Yuan, ``A2j:
  Anchor-to-joint regression network for 3d articulated pose estimation from a
  single depth image,'' in \emph{Proceedings of the IEEE/CVF International
  Conference on Computer Vision}, 2019, pp. 793--802.

\bibitem{fang2020jgr}
L.~Fang, X.~Liu, L.~Liu, H.~Xu, and W.~Kang, ``Jgr-p2o: Joint graph reasoning
  based pixel-to-offset prediction network for 3d hand pose estimation from a
  single depth image,'' in \emph{European Conference on Computer Vision}.\hskip
  1em plus 0.5em minus 0.4em\relax Springer, 2020, pp. 120--137.

\bibitem{ge20173d}
L.~Ge, H.~Liang, J.~Yuan, and D.~Thalmann, ``3d convolutional neural networks
  for efficient and robust hand pose estimation from single depth images,'' in
  \emph{Proceedings of the IEEE Conference on Computer Vision and Pattern
  Recognition}, 2017, pp. 1991--2000.

\bibitem{ge2018hand}
L.~Ge, Y.~Cai, J.~Weng, and J.~Yuan, ``Hand pointnet: 3d hand pose estimation
  using point sets,'' in \emph{Proceedings of the IEEE Conference on Computer
  Vision and Pattern Recognition}, 2018, pp. 8417--8426.

\bibitem{chen2019so}
Y.~Chen, Z.~Tu, L.~Ge, D.~Zhang, R.~Chen, and J.~Yuan, ``So-handnet:
  Self-organizing network for 3d hand pose estimation with semi-supervised
  learning,'' in \emph{Proceedings of the IEEE/CVF International Conference on
  Computer Vision}, 2019, pp. 6961--6970.

\bibitem{huang2020hand}
L.~Huang, J.~Tan, J.~Liu, and J.~Yuan, ``Hand-transformer: Non-autoregressive
  structured modeling for 3d hand pose estimation,'' in \emph{European
  Conference on Computer Vision}.\hskip 1em plus 0.5em minus 0.4em\relax
  Springer, 2020, pp. 17--33.

\bibitem{lin2020cross}
K.~Lin, L.~Wang, K.~Luo, Y.~Chen, Z.~Liu, and M.-T. Sun, ``Cross-domain
  complementary learning using pose for multi-person part segmentation,''
  \emph{IEEE Transactions on Circuits and Systems for Video Technology}, 2020.

\bibitem{tran2017disentangled}
L.~Tran, X.~Yin, and X.~Liu, ``Disentangled representation learning gan for
  pose-invariant face recognition,'' in \emph{Proceedings of the IEEE
  conference on computer vision and pattern recognition}, 2017, pp. 1415--1424.

\bibitem{chao2021gaitset}
H.~Chao, K.~Wang, Y.~He, J.~Zhang, and J.~Feng, ``Gaitset: Cross-view gait
  recognition through utilizing gait as a deep set,'' \emph{IEEE Transactions
  on Pattern Analysis and Machine Intelligence}, 2021.

\bibitem{liang2019cross}
G.~Liang, X.~Lan, X.~Chen, K.~Zheng, S.~Wang, and N.~Zheng, ``Cross-view person
  identification based on confidence-weighted human pose matching,'' \emph{IEEE
  Transactions on Image Processing}, vol.~28, no.~8, pp. 3821--3835, 2019.

\bibitem{wang2019learning}
Y.~Wang, C.~Song, Y.~Huang, Z.~Wang, and L.~Wang, ``Learning view invariant
  gait features with two-stream gan,'' \emph{Neurocomputing}, vol. 339, pp.
  245--254, 2019.

\bibitem{worrall2017harmonic}
D.~E. Worrall, S.~J. Garbin, D.~Turmukhambetov, and G.~J. Brostow, ``Harmonic
  networks: Deep translation and rotation equivariance,'' in \emph{Proceedings
  of the IEEE Conference on Computer Vision and Pattern Recognition}, 2017, pp.
  5028--5037.

\bibitem{lafarge2021roto}
M.~W. Lafarge, E.~J. Bekkers, J.~P. Pluim, R.~Duits, and M.~Veta,
  ``Roto-translation equivariant convolutional networks: Application to
  histopathology image analysis,'' \emph{Medical Image Analysis}, vol.~68, p.
  101849, 2021.

\bibitem{cohen2016group}
T.~Cohen and M.~Welling, ``Group equivariant convolutional networks,'' in
  \emph{International conference on machine learning}.\hskip 1em plus 0.5em
  minus 0.4em\relax PMLR, 2016, pp. 2990--2999.

\bibitem{esteves2019equivariant}
C.~Esteves, Y.~Xu, C.~Allen-Blanchette, and K.~Daniilidis, ``Equivariant
  multi-view networks,'' in \emph{Proceedings of the IEEE/CVF International
  Conference on Computer Vision}, 2019, pp. 1568--1577.

\bibitem{esteves2018polar}
C.~Esteves, C.~Allen-Blanchette, X.~Zhou, and K.~Daniilidis, ``Polar
  transformer networks,'' in \emph{International Conference on Learning
  Representations}, 2018.

\bibitem{gu2021cross}
X.~Gu, Y.~Guo, G.-Z. Yang, and B.~Lo, ``Cross-domain self-supervised complete
  geometric representation learning for real-scanned point cloud based
  pathological gait analysis,'' \emph{IEEE Journal of Biomedical and Health
  Informatics}, pp. 1--1, 2021.

\bibitem{cui2019effects}
W.~Cui, C.~Wang, W.~Chen, Y.~Guo, Y.~Jia, W.~Du, and C.~Wang, ``Effects of
  toe-out and toe-in gaits on lower-extremity kinematics, dynamics, and
  electromyography,'' \emph{Applied Sciences}, vol.~9, no.~23, p. 5245, 2019.

\bibitem{mahmood2020evaluation}
I.~Mahmood, U.~Martinez-Hernandez, and A.~A. Dehghani-Sanij, ``Evaluation of
  gait transitional phases using neuromechanical outputs and somatosensory
  inputs in an overground walk,'' \emph{Human Movement Science}, vol.~69, p.
  102558, 2020.

\bibitem{Joo_2017_TPAMI}
H.~Joo, T.~Simon, X.~Li, H.~Liu, L.~Tan, L.~Gui, S.~Banerjee, T.~S. Godisart,
  B.~Nabbe, I.~Matthews, T.~Kanade, S.~Nobuhara, and Y.~Sheikh, ``Panoptic
  studio: A massively multiview system for social interaction capture,''
  \emph{IEEE Transactions on Pattern Analysis and Machine Intelligence}, 2017.

\bibitem{loper2015smpl}
M.~Loper, N.~Mahmood, J.~Romero, G.~Pons-Moll, and M.~J. Black, ``Smpl: A
  skinned multi-person linear model,'' \emph{ACM transactions on graphics
  (TOG)}, vol.~34, no.~6, pp. 1--16, 2015.

\bibitem{kocabas2019self}
M.~Kocabas, S.~Karagoz, and E.~Akbas, ``Self-supervised learning of 3d human
  pose using multi-view geometry,'' in \emph{Proceedings of the IEEE/CVF
  Conference on Computer Vision and Pattern Recognition}, 2019, pp. 1077--1086.

\bibitem{you2020pointwise}
Y.~You, Y.~Lou, Q.~Liu, Y.-W. Tai, L.~Ma, C.~Lu, and W.~Wang, ``Pointwise
  rotation-invariant network with adaptive sampling and 3d spherical voxel
  convolution,'' in \emph{Proceedings of the AAAI Conference on Artificial
  Intelligence}, vol.~34, no.~07, 2020, pp. 12\,717--12\,724.

\bibitem{worrall2018cubenet}
D.~Worrall and G.~Brostow, ``Cubenet: Equivariance to 3d rotation and
  translation,'' in \emph{Proceedings of the European Conference on Computer
  Vision (ECCV)}, 2018, pp. 567--584.

\bibitem{vasileiadis2019optimising}
M.~Vasileiadis, C.-S. Bouganis, G.~Stavropoulos, and D.~Tzovaras, ``Optimising
  3d-cnn design towards human pose estimation on low power devices.'' in
  \emph{BMVC}, 2019, p.~42.

\bibitem{qi2016volumetric}
C.~R. Qi, H.~Su, M.~Nie{\ss}ner, A.~Dai, M.~Yan, and L.~J. Guibas, ``Volumetric
  and multi-view cnns for object classification on 3d data,'' in
  \emph{Proceedings of the IEEE conference on computer vision and pattern
  recognition}, 2016, pp. 5648--5656.

\bibitem{joung2020cylindrical}
S.~Joung, S.~Kim, H.~Kim, M.~Kim, I.-J. Kim, J.~Cho, and K.~Sohn, ``Cylindrical
  convolutional networks for joint object detection and viewpoint estimation,''
  in \emph{Proceedings of the IEEE/CVF Conference on Computer Vision and
  Pattern Recognition}, 2020, pp. 14\,163--14\,172.

\bibitem{yeh2019chirality}
R.~A. Yeh, Y.~T. Hu, and A.~G. Schwing, ``Chirality nets for human pose
  regression,'' \emph{Advances in Neural Information Processing Systems},
  vol.~32, 2019.

\bibitem{qi2017pointnet++}
C.~R. Qi, L.~Yi, H.~Su, and L.~J. Guibas, ``Pointnet++ deep hierarchical
  feature learning on point sets in a metric space,'' in \emph{Proceedings of
  the 31st International Conference on Neural Information Processing Systems},
  2017, pp. 5105--5114.

\bibitem{robinette2002civilian}
K.~M. Robinette, S.~Blackwell, H.~Daanen, M.~Boehmer, S.~Fleming, T.~Brill,
  D.~Hoeferlin, and D.~Burnsides, ``Civilian american and european surface
  anthropometry resource (caesar) final report,'' \emph{Volume I: Summary
  (United States Air Force Research Laboratory, Wright-Patterson Air Force
  Base, OH, 2002)}, 2002.

\bibitem{xiao2018simple}
B.~Xiao, H.~Wu, and Y.~Wei, ``Simple baselines for human pose estimation and
  tracking,'' in \emph{Proceedings of the European conference on computer
  vision (ECCV)}, 2018, pp. 466--481.

\end{thebibliography}
}

\newpage

\cleardoublepage
%\appendix
\noindent \textbf{\Large Supplemental Materials}

\section{Dataset Details}
\subsection{Experiment Settings}

\begin{figure}
\begin{minipage}[]{\linewidth}
\centering
\vspace{-6pt}
\includegraphics[width=0.85\linewidth]{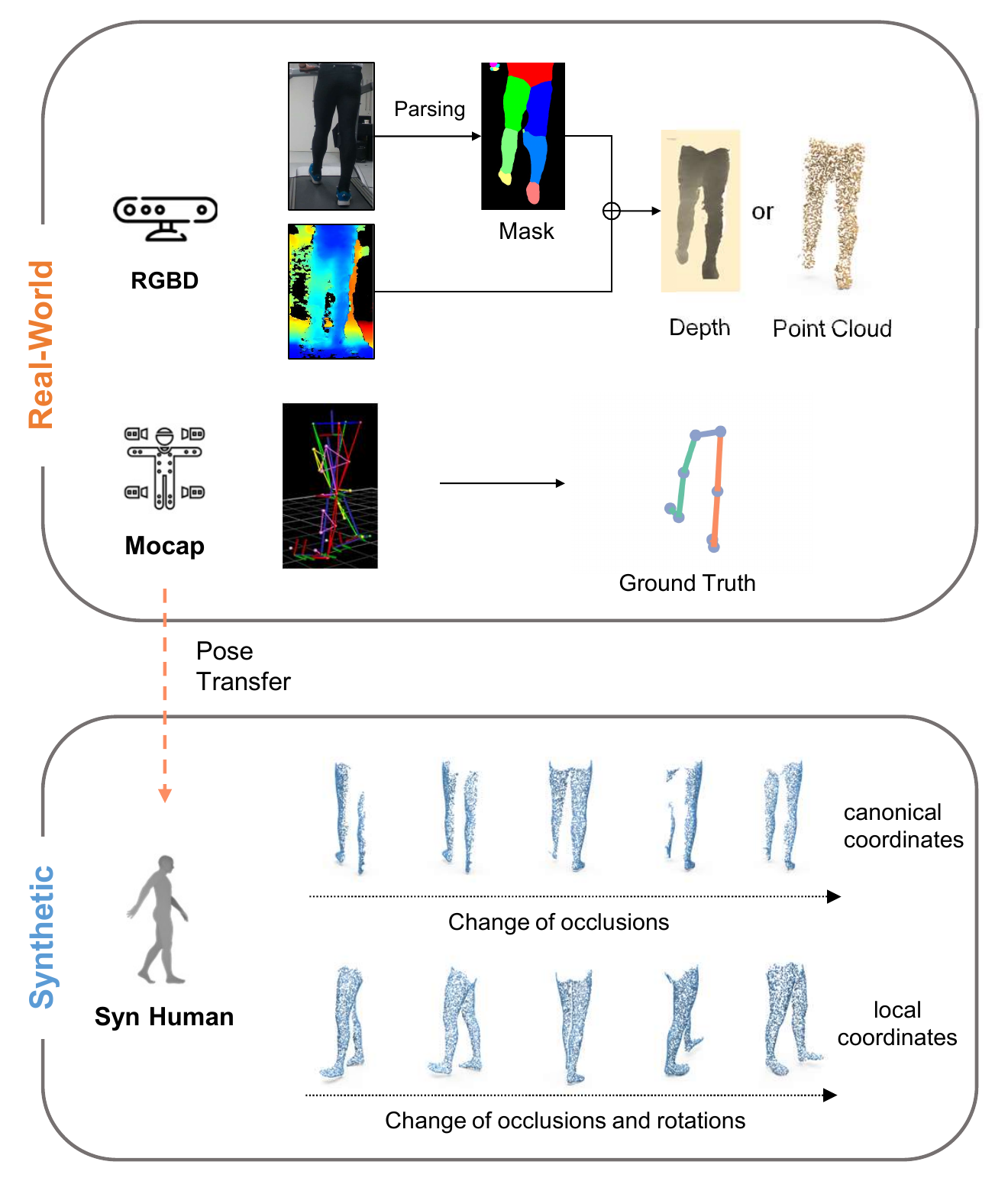}
\figcaption{Illustration of the data recording and processing procedures of our dataset. The depth camera and Mocap system simultaneously collected RGBD and ground truth skeleton/kinematics. The RGB image was processed by \cite{lin2020cross} to generate semantic labels for segmenting the lower limbs. The kinematics derived from Mocap was used to generate synthetic point clouds of different occlusions and rotations.}
\label{fig:datacollection}
\vspace{8pt}
\footnotesize
\centering
\tabcaption{Details of our gait pose dataset. (\textbf{C}\&\textbf{V} denote Condition and View separately.)}
\begin{tabular}{@{}P{0.5cm}P{0.7cm}P{0.5cm}P{1.2cm}c@{}}
\toprule
Subj & Gender & Age & Sample \# & Missing Trials \\ \midrule
1        & F      & 25  &  2200 & \scriptsize{\textbf{C}1-\textbf{V}1, \textbf{C}2-\textbf{V}2, \textbf{C}4-\textbf{V}1} \\
2        & M      & 24  &  2500  & -  \\
3        & M      & 32  &  2400 & \scriptsize{\textbf{C}3-\textbf{V}3} \\
4        & M      & 22  &  2500  & -  \\
5        & M      & 27  &  2400 & \scriptsize{\textbf{C}4-\textbf{V}4} \\
6        & M      & 30  &  2500 & -  \\
7        & M      & 24  &  2500 & -  \\
8        & F      & 22  &  2100 & \scriptsize{\textbf{C}2-\textbf{V}3, \textbf{C}2-\textbf{V}4, \textbf{C}4-\textbf{V}3, \textbf{C}5-\textbf{V}3} \\ \midrule
Total    & -      & -    & 19100  & - \\ \bottomrule
\end{tabular}
\label{tab:dataset}
\vspace{-3pt}
\end{minipage}
\end{figure}

\begin{figure}
    \begin{minipage}[b]{\linewidth}
      \centering
    \includegraphics[width=\linewidth]{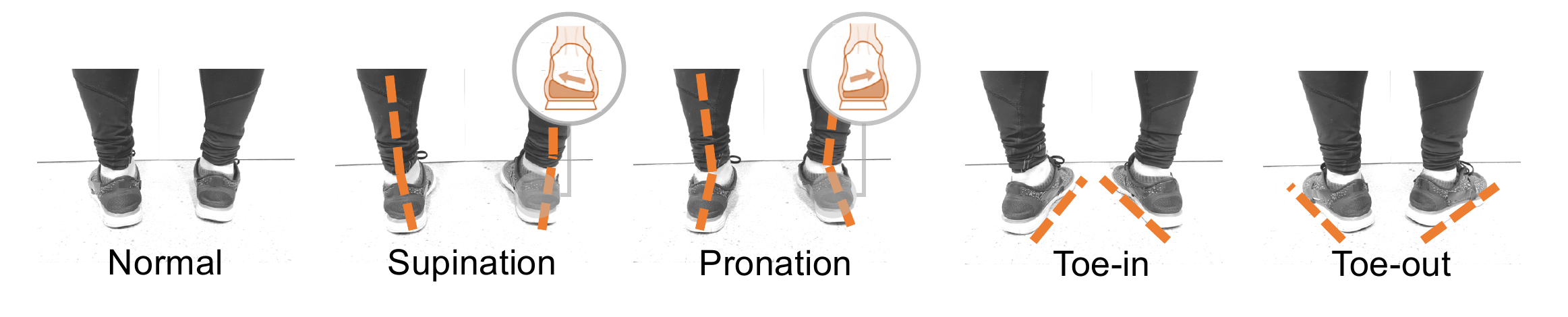}
    \caption{Illustration of different walking conditions. Supination and Pronation represent outward and inward roll of ankle joints respectively, which are induced by additional correction insoles in the participant's shoes. Toe-in and Toe-out represent the inward and outward foot orientation during walking, which are imitated by the participant. Images are adapted from \cite{gu2020cross}.}
    \label{fig:gaitcondition}
    \end{minipage}
    \begin{minipage}[b]{\linewidth}
        \centering
    \subfigure[Hips]{
    \includegraphics[width=0.15\linewidth]{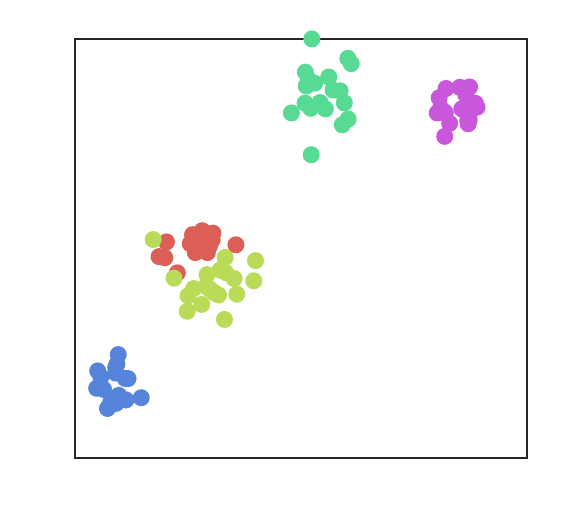}}
    \subfigure[Knees]{
    \includegraphics[width=0.15\linewidth]{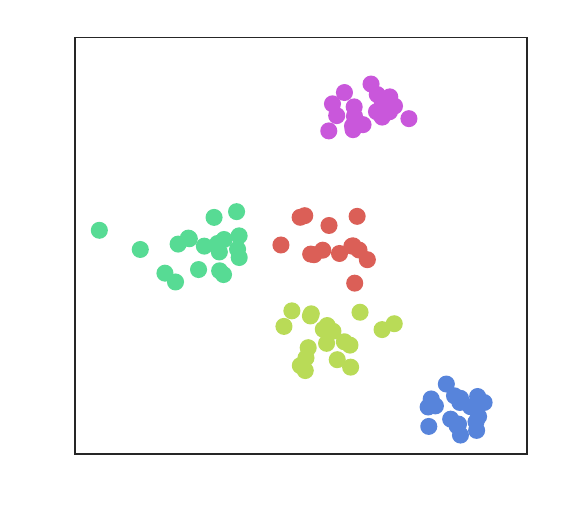}}
    \subfigure[Ankles]{
    \includegraphics[width=0.15\linewidth]{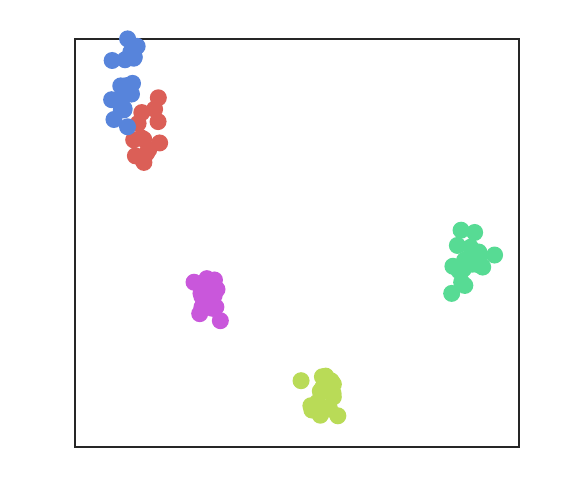}}
    \subfigure[All]{
    \includegraphics[width=0.15\linewidth]{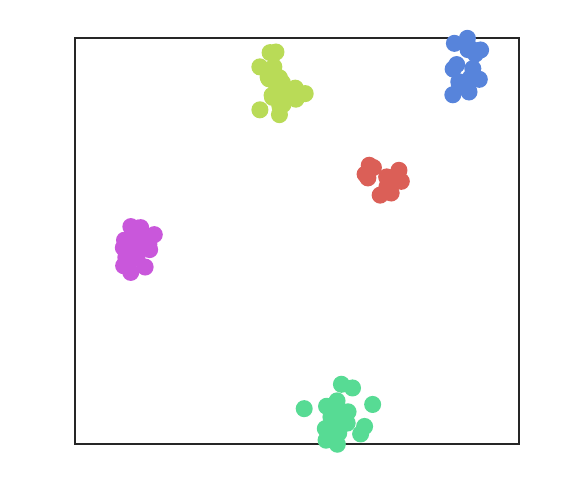}}
    \subfigure[Subjs]{
    \includegraphics[width=0.14\linewidth]{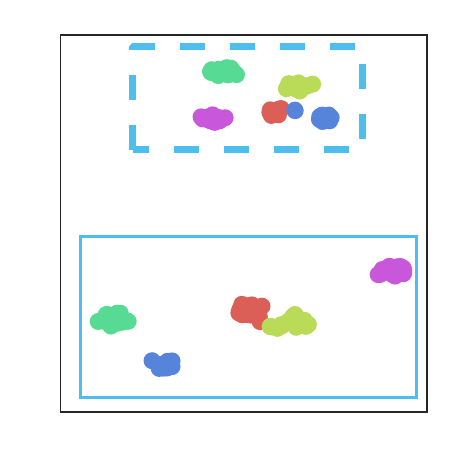}\label{fig:tsne_subj}}
    \subfigure{
    \includegraphics[trim=0 0 0 15pt, clip, width=0.08\linewidth]{Figure/legend_pattern.pdf}} \hfill
    \caption{\small t-SNE plots of (a)hip, (b)knee, (c)ankle rotational angles, or (d)all the joints of one subject. (e) is an extension of (d) to two subjects (dashed/solid rectangle). }
    \label{fig:angle}
    \vspace{-10pt}
    \end{minipage}
\end{figure}

The experiment settings have been briefly introduced in the main paper. Here, a more detailed illustration of the experiment is provided as shown in Fig.~\ref{fig:datacollection}. For each sample, RGB images, depth images, kinematics, and ground truth skeletons were recorded. The RGB image was processed by the human parsing algorithm cross-domain-complementary-learning (CDCL~\cite{lin2020cross}) to generate masks of lower limbs for background removal and lower limb segmentation. The segmented depth maps can be applied to generate 3D point clouds by the camera intrinsic parameters. Each trial lasts about 30s, and the first 100 frames of each trial are included in the public dataset. The illustration of different walking conditions is shown in Fig.~\ref{fig:gaitcondition}. The 3D data directly converted from the depth maps is denoted as $[R_{V_i}\circ\mathcal{P}_{V_i}](x^r)$.

For each subject, we randomly selected 1000 kinematics samples from the recorded kinematics sequence for synthetic data generation. The skinned multi-person linear model (SMPL~\cite{loper2015smpl}) is a realistic articulated human body model and has been applied in existing pose estimation research~\cite{varol2017learning}. It is controlled by two groups of parameters, the shape parameters and the pose parameters. The pose parameters are based on the kinematics derived from our real-world dataset, whereas the shape parameters are sampled from~\cite{robinette2002civilian}. We applied the hidden point removal provided by Open3D\footnote{\url{http://www.open3d.org/}} to simulate the occlusions contributed from different viewpoints. They were sampled from a canonical coordinate system for calculating multi-view consistency loss, as shown in the top layer of the synthetic block of Fig.~\ref{fig:datacollection}. For the methods without applying the multi-view consistency constraints, their inputs were sampled from their local camera coordinate system, as shown in the bottom layer of the synthetic block of Fig.~\ref{fig:datacollection}. 

It should be noted that for the viewpoint ${V_i}$ of synthetic and real data, although they share the same subscript, are roughly but not exactly the same, as we did not set any strict requirement for the extrinsic parameters of the camera. 

\vspace{-5pt}
\subsection{Subject Details}
The details of the data from each subject recruited in our experiments are presented in Table~\ref{tab:dataset}. In total eight subjects (2 females and 6 males) participated in our experiments and the sample size of each subject is given in Table~\ref{tab:dataset}.

\subsection{Discriminative Gait Patterns}
Fig.~\ref{fig:angle} demonstrates the discriminativeness between different gait  patterns for each joint category. It is plotted based on the t-SNE visualizations of the rotational angles derived from the Mocap system. It can be found that, functioning as a kinematic chain, the whole lower-limb kinematics would be influenced by simulating different walking patterns. Nevertheless, the whole lower limb kinematics show more distinct distributions. This emphasizes the necessity of paying attention to the whole lower limb joints, although the stimulation was raised from feet. We also provide the distributions of two selected subjects in Fig.~\ref{fig:tsne_subj}. It is observed that although intra-subject features present distinct distributions between different classes, inter-subject biases dominant the feature distribution. Such biases pose the challenge associated with overcoming individual difference when performance cross-subject validation.  

\section{Model Details}
As introduced in the main paper, based on voxelized cylindrical representations, each branch of our pose estimation model is composed of two parts, the 3D anisotropic convolution part and the 2D heatmap estimation part. There are two branches to handle $\theta\mbox{-}r$ and $\theta\mbox{-}z$ planes separately. 

\begin{figure}
    \centering
    \includegraphics[width=\linewidth]{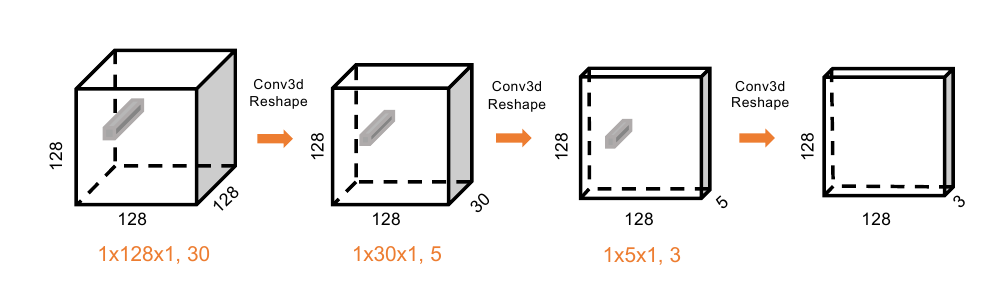}
    \caption{Details of the anisotropic 3D convolution network architectures. In each step, elongated anisotropic kernels (1$\times$k$\times$1) were applied to capture the long-distance spatial relationships along the axis to be ``projected'' through. Each convolution operation is followed by a Batch-Norm and ReLU layer. The input is a volumetric cube with a length of 128, while the output can be viewed as a 128$\times$128 planar image with a channel number of 3.}
    \label{fig:3d}
    \centering
    \includegraphics[width=\linewidth]{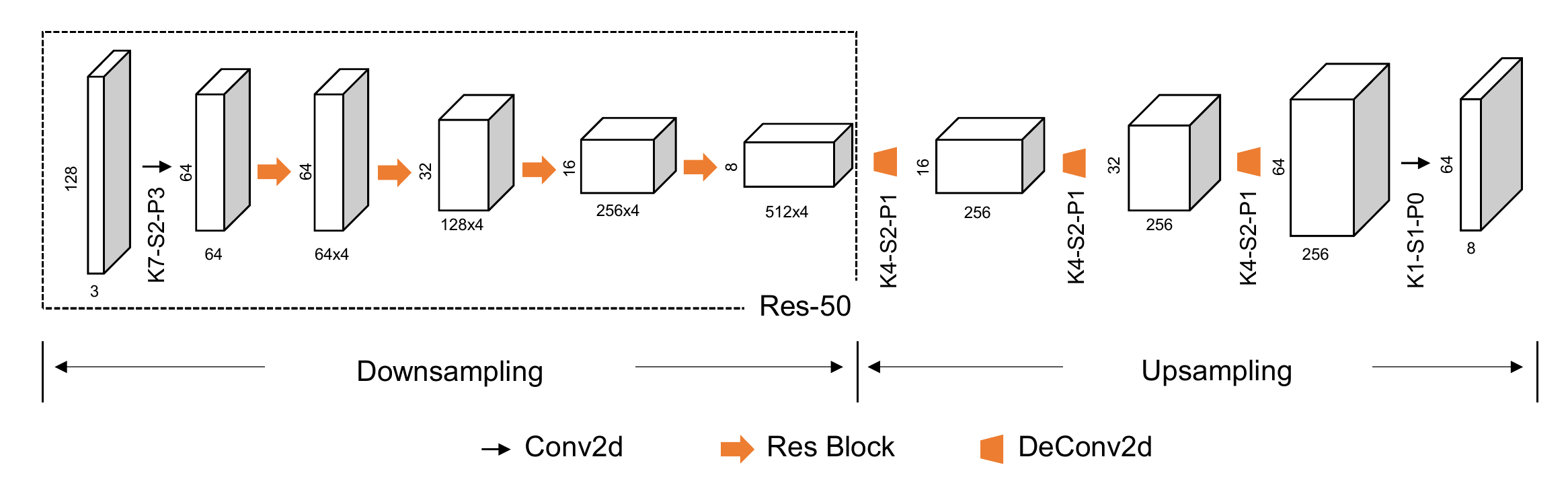}
    \caption{Details of our heatmap estimation network based on ResNet-50, modified from \cite{xiao2018simple}, where K, S, P denote the kernel size, stride number, and padding number respectively. The downsampling part is based on the default settings of the ResNet-50 without the max-pooling in the front layer and with our customized padding operations. The upsampling part is based on deconvolution operations, each followed by a Batch-Norm and a ReLU layer. The output of the upsampling branch is further filtered by a 1$\times$1 convolution layer to generate the final heatmap. The padding operation is based on periodic padding of $\theta$ axis and zero padding of the other axis. }
    \label{fig:poseres}
\end{figure}

The details of our network architectures are shown in Fig.~\ref{fig:3d} (Anisotropic Conv3D) and Fig.~\ref{fig:poseres} (PoseRes2D). The 3D anisotropic convolution aims to project the original 3D data into planar signals. Take the upper branch from 3D $\theta\mbox{-}r\mbox{-}z$ to 2D $\theta\mbox{-}r$ as an example. Although the point cloud is sparse, directly encoding the $z$ value of each occupied voxel into the planar grid might cause conflict problems. Therefore, we applied the 3D anisotropic convolution to capture and encode the long-range spatial features across $z$ axis. This 3D anisotropic convolution can be viewed as 2D convolutions by considering the $z$ as the feature channel dimension. The details of this part are described in Fig.~\ref{fig:3d}.  

After the ``projection'', a heatmap prediction model modified from \cite{xiao2018simple} was applied to generate the heatmap for each plane, as shown in Fig.~\ref{fig:poseres}. The downsampling part is based on ResNet-50 (w/o maxpooling after the first convolution layer) whereas the upsampling part is built on the deconvolution operations. All the involved padding operations for $\theta$ axis are based on the periodic padding according to the periodicity across $\theta$ axis.

The ResNet-50 part of PoseRes2D was initialized by its pre-trained weights. The remaining convolution/deconvolution weights were initialized by normal distribution of std 0.001 and bias by 0. The weights from Batch-Norm layer were initialized by constant values 1 and bias by 0. 

\section{Training Details}
\subsection{Cross-View Validation}
The training details of cross-view (CV) \& cross-view cross-subject (CV-CS) validation protocols are given below.  
\subsubsection{Proposed Method}
For our proposed method, we applied the Adam optimizer with the learning rate initialized as 1e-4. The learning rate was decayed by 0.5 per 20 steps. We started the semi-supervised training from Epoch 25 ($epoch_S$) and the epoch number of the whole training is 45. Our whole training procedure is described in Algorithm~\ref{algorithm}. It should be noted that one epoch represents a complete iterative sampling of our real-world dataset $x^r$. Empirically, this group of hyper-parameters was stable for all the validation protocols. 

The input for the proposed was $\{[R_{A}\circ\mathcal{P}_{A}](x^r)\}$ and $\{[R_{A}\footnote{Actually, for the synthetic input, the rotation does not need to be $R_A$ due to the rotation-equivariance of our backbone model. The synthetic input with different occlusions just need to be in a canonical coordinate system. Here setting it as $R_A$ is to make this framework also generalize to V2V-Semi. In practice for our proposed, we set it as $R_1$ for convenience.}\circ\mathcal{P}_{V_i}](x^s)\}$. 
We applied min-max normalization for all the input to normalize the center and scale of the input, and added Gaussian noises to the synthetic input to simply augment the realism.   
Only [-5$^\circ$, 5$^\circ$] rotation around x/y axis and slight 3D translation augmentation were then applied.

\begin{table*}[h]
    \centering
   \footnotesize
   \caption{Comparison of our proposed method w/ and w/o rotational augmentations around z axis under cross-view cross-subject (CV-CS) validation. No significant difference (p$<$0.05) can be observed, validating the rotation equivariance of our backbone.}
\begin{tabular}{@{}ccccccccc@{}}
\toprule
\multirow{2}{*}{Methods} & \multicolumn{4}{c}{Dist (cm) $\downarrow$}    & \multicolumn{4}{c}{mAP (5cm) $\uparrow$ }   \\ \cmidrule(l){2-9} 
                         & Hip          & Knee           & Ankle         & Toe          & Hip & Knee & Ankle & Toe    \\ \midrule
w/ rot & 5.44$\pm$3.80 & 3.70$\pm$2.76 & 3.68$\pm$2.70 & 5.08$\pm$3.82 & 0.567 & 0.834 & 0.817 & 0.608 \\
%Proposed-88 &&&&& \\
w/o rot & 5.51$\pm$3.70 & 3.69 $\pm$2.91 & 3.68$\pm$2.99 & 5.09$\pm$3.81 &  0.563  &  0.832  & 0.824  & 0.609      \\ \bottomrule
    \end{tabular}
    \label{tab:rot_compare}
\end{table*}

\begin{algorithm}[tp!]
\small
\caption{Semi-Supervised Framework}\label{algorithm}
\KwInput{training data \{$[R_A\circ\mathcal{P}_A](x^r), R_A y^r$\}, $[R_A\circ\mathcal{P}_{V_i}](x^s)$; scheduler, optimizer}
\KwOutput{Optimized pose estimation model $\mathcal{F}$}
\For{$epoch<epoch_{S}$}{
\For{iterations}{
 Sample $[R_A\circ\mathcal{P}_A](x^r)$, $R_A y^r$  \\
 Update $\mathcal{F}$ with $\mathcal{L}_S$  
}}

$\mathcal{F}_\gamma \leftarrow \mathcal{F}$

\For{$epoch>=epoch_{S}$}{
\For{iterations}{
  Sample $[R_A\circ\mathcal{P}_A](x^r)$, $R_A y^r$  \\
  Sample $\{[R_A\circ\mathcal{P}_{V_i}](x^s)\}_{i=1}^5$ \\ 
  Update $\mathcal{F}$ with $\mathcal{L}_S+\mathcal{L}_M+\mathcal{L}_{reg}$   
}
}
\end{algorithm}

\begin{figure}
    \centering
    \includegraphics[width=\linewidth]{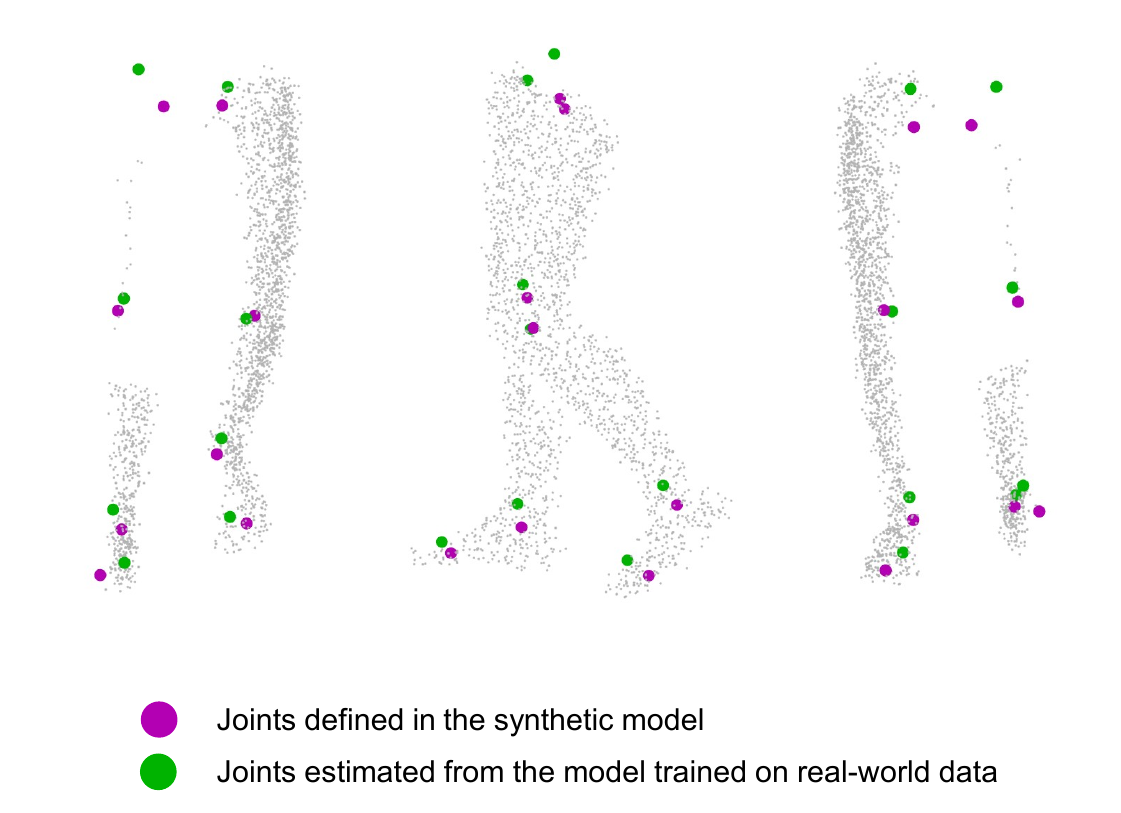}
    \caption{Comparison of the original keypoints defined in SMPL human model and the joints estimated from the model trained on the real-world data.}
    \label{fig:joint_offset}
\end{figure}

\subsubsection{Compared Methods}
For compared methods, we followed the default settings based on their online available codes. 

The input for PN++, SO-Net, V2V, and Proposed-Mix was $\{[R_{A}\circ\mathcal{P}_{A}](x^r)\}$ and $\{[R_{V_i}\circ\mathcal{P}_{V_i}](x^s)\}$. Only the synthetic data from the training subjects was utilized for training. All the synthetic input was added with Gaussian noises for augmentation. For SO-Net and PN++, we adopted their original preprocessing methods. For V2V and Proposed-Mix, we firstly performed the min-max normalization, and then applied [-5$^\circ$, 5$^\circ$] rotation around x/y axis, [-20$^\circ$, 20$^\circ$] rotation around the z axis, and slight 3D translation augmentation.

For V2V-Semi, the input was $\{[R_{A}\circ\mathcal{P}_{A}](x^r)\}$ and $\{[R_{A}\circ\mathcal{P}_{V_i}](x^s)\}$. We applied the min-max normalization, [-5$^\circ$, 5$^\circ$] around x/y axis, random rotational augmentation [-180$^\circ$, 180$^\circ$) around $z$ axis, together with slight translation augmentation.  

\subsection{Cross-Subject Validation}
For cross-subject validation, data from all the viewpoints of the training subjects was used for training. 
The input of all the methods was $\{[R_{V_i}\circ\mathcal{P}_{V_i}](x^r)\}$. 

For our proposed method, we applied the same hyper-parameters for training without using $\mathcal{L}_{M}$ and $\mathcal{L}_{reg}$, with the same preprocessing and augmentations as in the CV \& CV-CS session.

For A2J compared in the CS protocol, we adopted its default settings including the data augmentations. The center utilized in its data was based on the lower-limb masks generated from \cite{lin2020cross}. We did not compare A2J \cite{xiong2019a2j} in CV \& CV-CS protocols, since we directly generated synthetic point clouds. 

For PN++, SO-Net and V2V, their training was the same as that in the CV \& CV-CS protocols, except for the absence of synthetic data for training.

\section{Supplementary Results}
\subsection{Joint Difference Visualization}
To illustrate the domain shift between real-world data and synthetic data, we derived the pose estimation results of synthetic data based on the model directly trained with the real-world data. The predefined keypoint positions and estimated positions are visualized in Fig.~\ref{fig:joint_offset}. The differences of these two groups of keypoints are resulted from not only the original position deviations but also the heterogeneity between the real and synthetic input data caused by shapes, cloth/shoe distortions, noises, etc.

\subsection{Quantitative Results}
\paragraph{Comparison between w/ \& w/o rotation augmentations}
To highlight the rotation-equivariance of our proposed backbone, we did not exert additional z-axis rotational augmentations during the training. Here we also compare the version with additional aggressive rotation augmentations and w/o rotation augmentations. \textcolor{black}{As shown in Table~\ref{tab:rot_compare}}, no significant difference (paired t-test, p$<$0.05) can be seen between the results w/ and w/o rotational augmentations.

On the other hand, whilst additional aggressive rotational augmentations can achieve reasonably good performance of V2V (i.e. V2V-Semi), the performance contributed by such augmentations is still inferior to that of realizing the equivariance by the network itself (i.e. Our proposed).

\end{document}